\documentclass[acmtog,review=false,nonacm=true]{acmart}

\usepackage{booktabs} 

\setcitestyle{square}

\usepackage[ruled]{algorithm2e} 

\SetAlFnt{\small}
\SetAlCapFnt{\small}
\SetAlCapNameFnt{\small}
\SetAlCapHSkip{0pt}
\IncMargin{-\parindent}

\acmJournal{TOG}
\acmVolume{38}
\acmNumber{5}
\acmArticle{145}
\acmYear{2019}
\acmMonth{7}

\setcopyright{authorversion}

\acmDOI{10.1145/3341156}

\usepackage{booktabs} 
\usepackage{tabularx}
\usepackage{xargs}
\usepackage{mathtools}
\usepackage{amsmath,amssymb}
\usepackage{microtype}
\usepackage{units}
\usepackage[binary-units]{siunitx}
\usepackage{subcaption}
\usepackage{overpic}
\usepackage{cancel}
\usepackage{wrapfig}
\usepackage{placeins}

\usepackage{multirow}

\usepackage{url} 
\usepackage[utf8]{inputenc}

\def\commentType{0}

\ifnum\commentType=0
    \newcommandx{\customComment}[3]{}
    \newcommandx{\customTODO}[3]{}
\fi
\ifnum\commentType=1
    \usepackage[
    type=upperleft,
    unit=in
    ]{fgruler}
    \newcommandx{\customComment}[3]{\textcolor{#2}{\textsl{#1: #3}}}
    \newcommandx{\customTODO}[3]{\textcolor{#2}{\textsl{#1: #3}}}
\fi
\ifnum\commentType=2
    \usepackage[
    type=upperleft,
    unit=in
    ]{fgruler}
    \usepackage{pdfcomment}
    \newcommandx{\customComment}[3]{\pdfcomment[icon=Comment,opacity=0.5,color=#2,author=#1]{#3}}
    \newcommandx{\customTODO}[3]{\pdfcomment[icon=Note,opacity=0.5,color=#2,author=#1]{#3}}
\fi
\ifnum\commentType=3
    \usepackage[
    type=upperleft,
    unit=in
    ]{fgruler}
    \usepackage{pdfcomment} 
    \usepackage{todonotes} 
    \usepackage{silence}
    \newcommandx{\customComment}[3]{\todo[color=#2!40,size=\small]{\textbf{#1:} #3}}
    \newcommandx{\customTODO}[3]{\todo[color=#2!40,size=\small]{\textbf{#1:} #3}}
\fi

\definecolor{amber}{rgb}{1.0, 0.49, 0.0}
\definecolor{darkgreen}{rgb}{0.1, 0.7, 0.1}
\newcommandx{\All}[1]{\customComment{All}{red}{#1}}
\newcommandx{\Jan}[1]{\customComment{Jan}{magenta}{#1}}
\newcommandx{\Brian}[1]{\customComment{Brian}{blue}{#1}}
\newcommandx{\Fabrice}[1]{\customComment{Fabrice}{amber}{#1}}
\newcommandx{\Thomas}[1]{\customComment{Thomas}{darkgreen}{#1}}

\newcommandx{\TODO}[1]{\customTODO{TODO}{red}{#1}}
\newcommandx{\JanTODO}[1]{\customTODO{Jan}{magenta}{#1}}
\newcommandx{\FabriceTODO}[1]{\customTODO{Fabrice}{amber}{#1}}
\newcommandx{\ThomasTODO}[1]{\customTODO{Thomas}{darkgreen}{#1}}

\newcommand{\IGNORE}[1]{}

\newcommand{\ADD}[1]{#1} 
\newcommand{\ADDTWO}[1]{#1} 

\def\equationautorefname~#1\null{%
  Equation~(#1)\null{}
}

\newcommand{\Domain}{\mathcal{D}}

\newcommand{\Paths}{{\mathcal{P}}}

\newcommand{\PrimarySS}{{\mathcal{U}}}

\newcommand{\Diff}[1]{\,\mathrm{d}#1}

\newcommand{\aSpace}{\mathcal{X}}
\newcommand{\bSpace}{\mathcal{Y}}
\newcommand{\map}{h}

\newcommand{\compoundMap}{\widehat{\map}}
\newcommand{\cmap}{C}
\newcommand{\nnet}{m}

\newcommand{\partitionA}{A}
\newcommand{\partitionB}{B}

\newcommand{\MC}[1]{{\langle {#1} \rangle}}

\newcommand{\ldb}{\mathopen{\lbrack{} \! \lbrack{}}}
\newcommand{\rdb}{\mathclose{\rbrack{} \! \rbrack{}}}

\newcommand{\nlayers}{L}
\newcommand{\layerIn}{x}
\newcommand{\layerOut}{y}
\newcommand{\softmax}{\sigma}
\newcommand{\dimId}{i}
\newcommand{\binId}{j}
\newcommand{\containingBinId}{b}

\newcommand{\matQ}{Q}
\newcommand{\matW}{W}

\newcommand{\matV}{V}
\newcommand{\raw}[1]{{\widehat{#1}}}
\newcommand{\expFunc}[1]{{\exp\left(#1\right)}}
\newcommand{\lerpFunc}[3]{{\mathrm{lerp}{({#1},{#2},{#3})}}}

\newcommand{\nbins}{K}

\newcommand{\binWidth}{w}

\newcommand{\MakePath}[1]{\overline{#1}}
\newcommand{\Path}{\MakePath{x}}

\newcommand{\PSSVector}{z}
\newcommand{\PathConstructionMap}{\rho}

\newcommand{\dir}{{\omega}}
\newcommand{\diro}{\dir_\mathrm{o}}
\newcommand{\diri}{\omega}

\newcommand{\pos}{\mathbf{x}}

\newcommand{\radiance}{L}
\newcommand{\inRadiance}{\radiance}
\newcommand{\outRadiance}{\radiance_{\mathrm{o}}}
\newcommand{\emittedRadiance}{\radiance_{\mathrm{e}}}
\newcommand{\reflectedRadiance}{\radiance_{\mathrm{r}}}

\newcommand{\bsdf}{f_{\mathrm{s}}}
\newcommand{\selectProb}{c}

\newcommand{\fsangle}{\gamma}

\newcommand{\Measurement}{I}
\newcommand{\Importance}{W}

\newcommand{\PathThroughput}{T}

\newcommand{\Sample}{X}
\newcommand{\Params}{\theta}

\newcommand{\PdfGt}{p}

\newcommand{\PdfMC}{q}
\newcommand{\PdfPoly}{q}
\newcommand{\PdfOptimized}{q}
\newcommand{\PdfBSDF}{p_{\bsdf}}

\newcommand{\Div}{D}
\newcommand{\KlDiv}{\Div_{\text{KL}}}
\newcommand{\ChiDiv}{\Div_{\chi^2}}
\newcommand{\Expectation}{\mathbb{E}}
\newcommand{\Variance}{\mathbb{V}}

\newcommand{\R}{\mathbb{R}}

\newcommand{\AluminumSphere}{\textsc{Aluminum Sphere}}
\newcommand{\CornellBox}{\textsc{Cornell Box}}
\newcommand{\GlossyCornellBox}{\textsc{Yet Another Box}}

\newcommand{\SalleDeBain}{\textsc{Salle de Bain}}
\newcommand{\Bathroom}{\textsc{Bathroom}}
\newcommand{\Bedroom}{\textsc{Bedroom}}

\newcommand{\Spaceship}{\textsc{Spaceship}}
\newcommand{\WoodenStaircase}{\textsc{Staircase}}
\newcommand{\CopperHairball}{\textsc{Copper Hairball}}

\newcommand{\CountryKitchen}{\textsc{Country Kitchen}}
\newcommand{\WhiteRoom}{\textsc{White Room}}
\newcommand{\SwimmingPool}{\textsc{Swimming Pool}}
\newcommand{\Necklace}{\textsc{Necklace}}
\newcommand{\GlossyKitchen}{\textsc{Glossy Kitchen}}
\newcommand{\Bookshelf}{\textsc{Bookshelf}}
\newcommand{\Sponza}{\textsc{Sponza Atrium}}
\newcommand{\VeachDoor}{\textsc{Veach Door}}

\gdef\useCroppedImages{1}

\gdef\cropInsets{0}

\usepackage[export]{adjustbox}
\usepackage{calc}
\usepackage{tabularx}
\usepackage{tikz}
\usepackage{xstring}

\newlength{\beautyHeight}
\newlength{\beautyPixWidth}
\newlength{\beautyPixHeight}
\newlength{\insetvsep}
\gdef\useInsetA{0}
\gdef\useInsetB{0}
\gdef\useInsetC{0}

\newcommand{\setInset}[6]{%
    \expandafter\gdef\csname useInset#1\endcsname{1}%
    \expandafter\gdef\csname inset#1Color\endcsname{#2}%
    \expandafter\gdef\csname crop#1X\endcsname{#3}%
    \expandafter\gdef\csname crop#1Y\endcsname{#4}%
    \expandafter\gdef\csname crop#1W\endcsname{#5}%
    \expandafter\gdef\csname crop#1H\endcsname{#6}%
}

\newcommand{\unsetInset}[1]{%
    \expandafter\gdef\csname useInset#1\endcsname{0}%
}

\newcommand{\addBeautyCrop}[8]{%
    \pdfpxdimen=\dimexpr 1 in/72\relax
    \def\beauty{%
        \let\cropR\relax%
        \let\cropB\relax%
        \newlength\cropR%
        \newlength\cropB%
        \setlength\cropR{{#3 px}-{#5 px}-{#7 px}}%
        \setlength\cropB{{#4 px}-{#6 px}-{#8 px}}%
        \sbox0{\includegraphics[width=#2\textwidth,trim={#5px {\cropB} {\cropR} #6px},clip]{#1}}%
        \begin{tikzpicture}
            \node[anchor=north west,inner sep=0] at (0,0) {\usebox0};
            \begin{scope}[x=\wd0/#7, y=\ht0/#8]
            \if\useInsetA1{
                \draw[\insetAColor,thick] (\cropAX-#5,-\cropAY+#6) rectangle + (\cropAW,-\cropAH);
            }\fi
            \if\useInsetB1{
                \draw[\insetBColor,thick] (\cropBX-#5,-\cropBY+#6) rectangle + (\cropBW,-\cropBH);
            }\fi
            \if\useInsetC1{
                \draw[\insetCColor,thick] (\cropCX-#5,-\cropCY+#6) rectangle + (\cropCW,-\cropCH);
            }\fi
            \end{scope}
        \end{tikzpicture}
    }%
    \setlength\beautyHeight{\heightof{\beauty}}%
    \setlength\beautyPixWidth{#3px}%
    \setlength\beautyPixHeight{#4px}%
    \global\beautyHeight=\beautyHeight%
    \global\beautyPixWidth=\beautyPixWidth%
    \global\beautyPixHeight=\beautyPixHeight%
    \begin{adjustbox}{valign=t}
        \beauty{}
    \end{adjustbox}
}

\newcommand{\trimInset}[6]{%
    \let\cropR\relax%
    \let\cropB\relax%
    \newlength\cropR%
    \newlength\cropB%
    \setlength\cropR{{\beautyPixWidth}-{#3 px}-{#5 px}}%
    \setlength\cropB{{\beautyPixHeight}-{#4 px}-{#6 px}}%
    \color{#2}%
    \fbox{\includegraphics[width=1\linewidth,trim={{#3 px} {\cropB} {\cropR} {#4 px}},clip]{#1}}%
}

\newcommand{\addInset}[2]{%
    \color{#2}%
    \fbox{\includegraphics[width=1\linewidth]{#1}}%
}

\newcommand{\auxtimes}{x}
\newcommand{\auxplus}{+}
\newcommand{\auxspace}{ }

\newcommand{\addInsets}[1]{%
    \begin{adjustbox}{valign=t}
        \StrSubstitute{#1}{.}{-}[\baseFileName]
        \begin{adjustbox}{totalheight=1\beautyHeight,tabular={c}}
            \if\useInsetA1%
                \def\cropfile{\baseFileName-\cropAW\auxtimes\cropAH\auxplus\cropAX\auxplus\cropAY}
                \if\cropInsets1
                    \immediate\write18{convert #1 -crop \cropAW\auxtimes\cropAH\auxplus\cropAX\auxplus\cropAY\auxspace  -filter point -resize 800\% \cropfile.jpg}
                \fi
                \if\useCroppedImages1
                    \addInset{\cropfile.jpg}{\insetAColor}
                \else
                    \trimInset{#1}{\insetAColor}{\cropAX}{\cropAY}{\cropAW}{\cropAH}%
                \fi%
            \fi%
            \if\useInsetB1%
                \if\useInsetA1\\[\insetvsep]\fi%
                \def\cropfile{\baseFileName-\cropBW\auxtimes\cropBH\auxplus\cropBX\auxplus\cropBY}
                \if\cropInsets1
                    \immediate\write18{convert #1 -crop \cropBW\auxtimes\cropBH\auxplus\cropBX\auxplus\cropBY\auxspace -filter point -resize 800\% \cropfile.jpg}
                \fi
                \if\useCroppedImages1
                    \addInset{\cropfile.jpg}{\insetBColor}
                \else
                    \trimInset{#1}{\insetBColor}{\cropBX}{\cropBY}{\cropBW}{\cropBH}%
                \fi%
            \fi%
            \if\useInsetC1%
                \if\useInsetB1\\[\insetvsep]\fi%
                \def\cropfile{\baseFileName-\cropCW\auxtimes\cropCH\auxplus\cropCX\auxplus\cropCY}
                \if\cropInsets1
                    \immediate\write18{convert #1 -crop \cropCW\auxtimes\cropCH\auxplus\cropCX\auxplus\cropCY\auxspace -filter point -resize 800\% \cropfile.jpg}
                \fi
                \if\useCroppedImages1
                    \addInset{\cropfile.jpg}{\insetCColor}
                \else
                    \trimInset{#1}{\insetCColor}{\cropCX}{\cropCY}{\cropCW}{\cropCH}%
                \fi%
            \fi%
        \end{adjustbox}
    \end{adjustbox}
}

\definecolor{mathematicaBlue}{rgb}{0.38, 0.51, 0.71}
\definecolor{mathematicaOrange}{rgb}{0.88, 0.61, 0.14}
\definecolor{mathematicaGreen}{rgb}{0.56, 0.69, 0.19}
\definecolor{mathematicaRed}{rgb}{0.92,0.39, 0.21}
\definecolor{mathematicaPurple}{rgb}{0.53, 0.47, 0.7}

\begin{document}
\title{Neural Importance Sampling}

\author{Thomas M\"uller}
\affiliation{%
  \institution{Disney Research \& ETH Z\"urich}
  }
\email{thomas94@gmx.net}

\author{Brian McWilliams}
\affiliation{%
  \institution{Disney Research}
  }
\email{bvpmcwilliams@gmail.com}

\author{Fabrice Rousselle}
\affiliation{%
  \institution{Disney Research}
  }
\email{fabrice.rousselle@gmail.com}

\author{Markus Gross}
\affiliation{%
  \institution{Disney Research \& ETH Z\"urich}
  }
\email{grossm@inf.ethz.ch}

\author{Jan Nov\'ak}
\affiliation{%
  \institution{Disney Research}
  }
\email{novakj4@gmail.com}

\renewcommand\shortauthors{M\"uller et al.}

\begin{abstract}
We propose to use deep neural networks for generating samples in Monte Carlo integration.
Our work is based on non-linear independent components estimation (NICE), which we extend in numerous ways to improve performance and enable its application to integration problems.
First, we introduce piecewise-polynomial coupling transforms that greatly increase the modeling power of individual coupling layers.
Second, we propose to preprocess the inputs of neural networks using one-blob encoding, which stimulates localization of computation and improves inference.
Third, we derive a gradient-descent-based optimization for the KL and the $\chi^2$ divergence for the specific application of Monte Carlo integration with unnormalized stochastic estimates of the target distribution.
Our approach enables fast and accurate inference and efficient sample generation independently of the dimensionality of the integration domain.
We show its benefits on generating natural images and in two applications to light-transport simulation:
first, we demonstrate learning of joint path-sampling densities in the primary sample space and importance sampling of multi-dimensional path prefixes thereof.
Second, we use our technique to extract conditional directional densities driven by the product of incident illumination and the BSDF in the rendering equation, and we leverage the densities for path guiding.
In all applications, our approach yields on-par or higher performance than competing techniques at equal sample count.
\end{abstract}


\maketitle

\section{Introduction}%
\label{sec:introduction}

Solving integrals is a fundamental problem of calculus that appears in many disciplines of science and engineering.
Since closed-form antiderivatives exist only for elementary problems, many applications resort to numerical recipes.
Monte Carlo (MC) integration is one such approach, which relies on sampling a number of points within the integration domain and averaging the integrand thereof.
The main drawback of MC methods is the relatively low convergence rate.
Many techniques have thus been developed to reduce the integration error, e.g.\ via importance sampling, control variates, Markov chains, integration on multiple accuracy levels, and use of quasi-random numbers.

In this work, we focus on the concept of importance sampling and propose to parameterize the sampling density using a collection of neural networks.
Generative neural networks have been successfully leveraged in many fields, including signal processing, variational inference, and probabilistic modeling, but their application to MC integration---in the form of sampling densities---remains to be investigated; this is what we strive for in the present paper.

Given an integral
\begin{align}
F = \int_{\Domain} f(x) \Diff{x} \,,
\end{align}
we can introduce a probability density function (PDF) $\PdfMC(x)$, which, under certain constraints, allows expressing $F$ as the expected ratio of the integrand and the PDF:\@
\begin{align}
F
= \int_{\Domain} \frac{f(x)}{\PdfMC(x)} \PdfMC(x) \Diff{x}
= \Expectation \left[ \frac{f(X)}{\PdfMC(X)} \right] \,.
\end{align}
The above expectation can be approximated using $N$ independent, randomly chosen points $\{X_1, X_2, \cdots X_N\}; X_i \in \Domain, X_i \sim \PdfMC(x)$, with the following MC estimator:
\begin{align}
F  \approx \langle F \rangle_{N} = \frac{1}{N}  \sum_{i=1}^{N}\frac{f(X_i)}{\PdfMC(X_i)} \,.
\label{eq:basic-mc-estimator}
\end{align}
The variance of the estimator, besides being inversely proportional to $N$, heavily depends on the shape of $\PdfMC$.
If $\PdfMC$ follows normalized $f$ closely, the variance is low. If the shapes of the two differ significantly, the variance tends to be high.
In the special case when samples are drawn from a PDF proportional to $f(x)$, i.e.\ $\PdfGt(x) \equiv f(x)/F$, we obtain a zero-variance estimator, $\langle F \rangle_N = F$, for any $N \geq 1$.

It is thus crucial to use \emph{expressive} sampling densities that match the shape of the integrand well.
Additionally, generating sample $X_i$ must be \emph{fast} (relative to the cost of evaluating $f$), and \emph{invertible}. That is, given a sample $X_i$, we require an efficient and exact evaluation of its corresponding probability density $\PdfMC(X_i)$---a necessity for evaluating the unbiased estimator of~\autoref{eq:basic-mc-estimator}.
Being expressive, fast to evaluate, and invertible are the key properties of good sampling densities, and all our design decisions can be traced back to these.

We focus on the general setting where little to no prior knowledge about $f$ is given, but $f$ can be observed at a sufficiently high number of points.
Our goal is to extract the sampling PDF from these observations
while handling complex distributions with possibly many modes and arbitrary frequencies.
To that end, we approximate the ground-truth $\PdfGt(x)$ using a generative probabilistic parametric model $\PdfMC(x;\Params)$
that utilizes deep neural networks.

Our work builds on approaches that are capable of compactly representing complex manifolds in high-dimensional spaces, and permit fast and exact inference, sampling, and \ADD{PDF evaluation}.
We extend the work of~\citet{dinh2014nice,dinh2016density} on learning stably invertible transformations, represented by so-called \emph{coupling layers}, that are stacked to produce highly nonlinear mappings between an observation $x$ and a latent variable $z$.
Specifically, we introduce \ADD{two types of \emph{piecewise-polynomial} coupling layers---piecewise-linear and piecewise-quadratic---}that greatly increase the expressive power of individual coupling layers, allowing us to employ fewer of those and thereby reduce the total cost of evaluation.

After reviewing related work on generative neural networks in \autoref{sec:related-work}, we detail the framework of \emph{non-linear independent components estimation} (NICE)~\cite{dinh2014nice,dinh2016density} in \autoref{sec:nice}; this forms the foundation of our approach.

In \autoref{sec:piecewise-poly}, we describe a \ADD{new} class of \emph{invertible piecewise-polynomial} coupling transforms that replace affine transforms proposed in the original work.
We also introduce the \emph{one-blob}-encoding of network inputs, which stimulates localization of computation and improves inference.
We illustrate the benefits on low-dimensional regression
problems and test the performance when learning a (high-dimensional) distribution of natural images.

In \autoref{sec:mc}, we apply NICE to Monte Carlo integration and
propose an optimization strategy for minimizing estimation variance.

Finally in \autoref{sec:path-guiding}, we present the benefits of using NICE in light-transport simulations. We use NICE with our polynomial warps to guide the construction of light paths and demonstrate that, while currently being impractical due to large computational overhead, it outperforms the state of the art at equal sample counts in \emph{path guiding} and primary-sample-space \emph{path sampling}.
\ADDTWO{We combine our path-guiding distribution with other sampling distributions using multiple importance sampling (MIS)~\cite{Veach:1995:MIS} and use a dedicated network to learn the approximately optimal selection probabilities to further improve the MIS performance.}

In summary, our contributions are
\begin{itemize}
    \item two \emph{piecewise-polynomial} coupling transforms (piecewise-linear and piecewise-quadratic) that improve expressiveness,
    \item \emph{one-blob}-encoded network inputs---a generalization of one-hot encoding---for improving learning speed and quality,
    \item stochastic gradients that can be used for optimizing the KL and $\chi^2$ divergences when only MC estimates of the unnormalized target distribution are available, and
    \item an application of NICE with the aforementioned tools to the problem of light-transport simulation with
    \item \ADDTWO{data-driven probabilities for selecting sampling techniques.}
\end{itemize}

\section{Background and Related Work}%
\label{sec:related-work}

\ADDTWO{Since our goal is to learn a probability distribution and---among other things---draw samples from it, our approach falls into the category of \emph{generative modeling}.
In the following, we review the most relevant generative neural networks focusing on the requirements for MC integration.
Specifically, we seek a model that provides a parametric PDF $\PdfMC(x; \Params)$ for approximating the ideal PDF $\PdfGt(x) \equiv f(x)/F$. We also need an optimization scheme that tolerates noisy estimates of the integrand $f(x)$.
Lastly, the trained model must permit efficient sampling and evaluation of $\PdfMC(x; \Params)$.}

\ADDTWO{Several prior generative models were built on less stringent requirements.
For example, it is often the case that only the \emph{synthesis} of samples is required without explicitly evaluating $\PdfMC(x; \Params)$~\citep{goodfellow2014generative, germain2015made, oord2016wavenet, van2016pixel}.
These models are thus not easily applicable to MC integration in the aforementioned manner.
In the following, we focus on existing techniques that show promise in satisfying our requirements.}
\ADDTWO{\paragraph{The Latent-Variable Model.}
Many existing generative models rely on auxiliary unobserved ``latent'' variables $z$ with fixed, prescribed PDF $\PdfMC(z)$, where each possible value of $z$ gives rise to a unique conditional distribution $\PdfMC(x | z; \Params)$ that is learnable via parameters $\Params$.
Since any particular value of $x$ can be caused by multiple different values of $z$, one must resort to integration to obtain $\PdfMC(x; \Params)$
\begin{align}
    \PdfMC(x; \Params) = \int \PdfMC(x | z; \Params) \PdfMC(z) \Diff{z} \,.
    \label{eq:latent}
\end{align}
In this context, $\PdfMC(x; \Params)$ is referred to as the ``marginal likelihood'' and $\PdfMC(z)$ as the ``prior distribution''.
The prior is often modeled as a simple distribution (e.g.\ Gaussian or uniform) and it is the task of a neural network to learn the parameters of $\PdfMC(x | z; \Params)$.}

\ADDTWO{Unfortunately, the above integral is often not solvable in closed form, necessitating its estimation with another MC estimator.
It may be tempting to use such estimates of $\PdfMC(x; \Params)$ in the denominator of our desired MC estimator
\begin{align}
    \langle F \rangle_{N} \approx \frac{1}{N}  \sum_{i=1}^{N}\frac{f(X_i)}{\MC{\PdfMC(X_i)}} \,, \nonumber
\end{align}
but this scheme introduces undesirable bias\footnote{This can be verified using Jensen's inequality.}.}
\ADDTWO{\paragraph{Normalizing Flows.}
To avoid the limitation of having to estimate $\PdfMC(x; \Params)$, a growing literature emerged that models $x$ as a deterministic bijective mapping of $z$, a so-called ``normalizing flow'' $x = \map(z;\Params)$.
This has the effect of assigning only a \emph{single} value of $x$ to each possible value of $z$ (and vice versa), thereby avoiding the difficult integration when computing the likelihood.
Mathematically, $\PdfMC(x | z; \Params)$ becomes a Dirac-delta function $\delta\big(x - \map(z;\Params)\big)$, resulting in
\begin{align}
    \PdfMC(x; \Params) = \int \delta\big(x - \map(z;\Params)\big) \PdfMC(z) \Diff{z} = \PdfMC(z) \left\vert \, \det\left( \frac{\partial \map(z;\Params)}{\partial z^T} \right) \, \right\vert^{-1} \,.
    \label{eq:pdf-change-of-variable}
\end{align}
Here, the inverse Jacobian determinant accounts for the change in density due to $\map$ in the infinitesimal neighborhood around $x$.}

\ADDTWO{Although \autoref{eq:pdf-change-of-variable} no longer contains a difficult integral, there exist a number of additional requirements on $\map$ to make the usage of $\PdfMC(x; \Params)$ in MC integration practical.
Since $\PdfMC(x; \Params)$ is expressed in terms of $z$, one must know the value of $z$ that corresponds to $x$.
Generally, this requires a tractable inverse $z = \map^{-1}(x)$.\footnote{MC estimators that only sample from $\PdfMC(x; \Params)$ are an exception, because they can simply use the $z$ that generated $x$. However, when combining multiple PDFs using MIS heuristics~\cite{Veach:1995:MIS} one must be able to evaluate $\PdfMC(x; \Params)$ for $x$ that were drawn from the other distributions.
}
Additionally, to ensure efficiency, the evaluation of both $\map$ and $\map^{-1}$ as well as the corresponding Jacobian determinant must be \emph{fast} relative to the cost of evaluating $f$.}
\ADDTWO{\paragraph{Prior Work on Normalizing Flows.}
In the following, we mention a number of existing techniques based on normalizing flows that satisfy some (but not necessarily all) of our requirements.}

\ADDTWO{\citet{rezende2015variational} model the posterior distribution using normalizing flows to perform variational inference more effectively.
Unfortunately, their computation graph is difficult to invert, not permitting exact evaluation of $\PdfMC(x; \Params)$.}

\ADDTWO{\citet{chen2018neural} propose a continuous analog of normalizing flows that utilizes the \emph{instantaneous} change-of-variable formula, which only requires computing the \emph{trace} of the Jacobian as opposed to the determinant.
This reduces computational cost in many situations.
Unfortunately, the evaluation of their continuous normalizing flows relies on a numeric ODE solver, which reintroduces computational cost in other places and results in approximation error when computing $\PdfMC(x; \Params)$.
This approximation causes bias in our use case of MC integration and therefore disqualifies their approach.}

\ADDTWO{A number of recent approaches investigate the usage of normalizing flows for auto-regressive density estimation~\citep{kingma2016improved, Huang2018NeuralAF, papamakarios2017masked}.
These ``autoregressive flows'' offer the desired exact evaluation of $\PdfMC(x; \Params)$.
Unfortunately, they generally only permit \emph{either} efficient sample generation \emph{or} efficient evaluation of $\PdfMC(x; \Params)$, which makes them prohibitively expensive for our application to MC integration.}

Lastly, ``non-linear independent components estimation'' (NICE) \citep{dinh2014nice,dinh2016density} is a special case of autoregressive flows that allows \emph{both} fast sampling \emph{and} density evaluation through the use of so-called ``coupling layers'', the composition of which constitutes a normalizing flow.
Because of NICE's efficient sample generation and its efficient, exact density evaluation, NICE satisfies all our postulated requirements for usage in MC integration and we therefore base our work on it.
Concurrently with us, \citet{DBLP:journals/corr/abs-1808-07840} also investigate the applicability of NICE to MC integration.

\ADDTWO{\citet{kingma18glow} extend the work of \citet{dinh2016density} with invertible $1\times1$ convolutions, achieving better results.
The $1\times1$ convolutions, however, require a deep computation graph to be effective, which is oppsite to the shallow computation graph we desire for efficiency. We therefore do not adopt the $1\times1$ convolutions.}

\ADDTWO{In the following section, we introduce NICE in detail and proceed with describing our piecewise-polynomial coupling layers that increase its modeling capacity.}

\section{Non-linear Independent Components Estimation}%
\label{sec:nice}

In this section, we detail the works of \citet{dinh2014nice,dinh2016density} that form the basis of our approach.
The authors propose to learn a mapping between the data and the latent space as an invertible compound function $\compoundMap = \map_\nlayers \circ \cdots \circ \map_2 \circ \map_1$, where each $\map_i$ is a relatively simple bijective transformation (warp).
The choice of the type of $\map$ is different in the two prior works and in our paper (details follow in \autoref{sec:piecewise-poly}), but the key design principle remains: $\map$~needs to be stably invertible with (computationally) tractable Jacobians. This enables exact and fast inference of latent variables \ADD{as well as exact and fast probability density evaluation}:
given a differentiable mapping $\map: \aSpace \rightarrow \bSpace$
of points ${x \sim \PdfGt_\aSpace(x)}$ to points $y \in \bSpace$,
we can compute the \ADD{PDF} $\PdfGt_\bSpace(y)$ of transformed points $y = \map(x)$ using the change-of-variables formula:
\begin{align}
    \PdfGt_\bSpace(y) = \PdfGt_\aSpace(x) \left\vert \, \det\left( \frac{\partial \map(x)}{\partial x^T} \right) \, \right\vert^{-1} \,,
    \label{eq:change-of-variable}
\end{align}
where $\frac{\partial \map(x)}{\partial x^T}$ is the Jacobian of $\map$ at $x$.

The cost of computing the determinant grows superlinearly with the dimensionality of the Jacobian.
If $\aSpace$ and $\bSpace$ are high-dimensional, computing $\PdfGt_\bSpace(y)$ is therefore computationally intractable.
The key proposition of \citet{dinh2014nice} is to focus on a specific class of mappings---referred to as \emph{coupling layers}---that admit Jacobian matrices where determinants reduce to the product of diagonal terms.

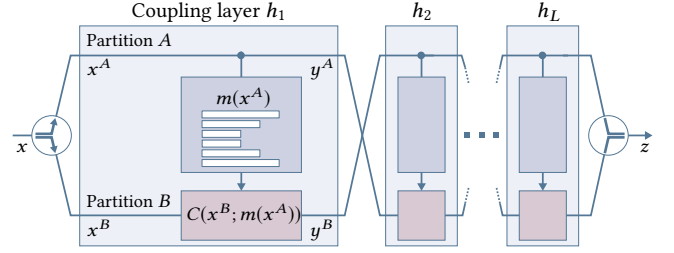
\begin{figure}[t]
    \vspace{4mm}
    \begin{overpic}[width=\columnwidth]{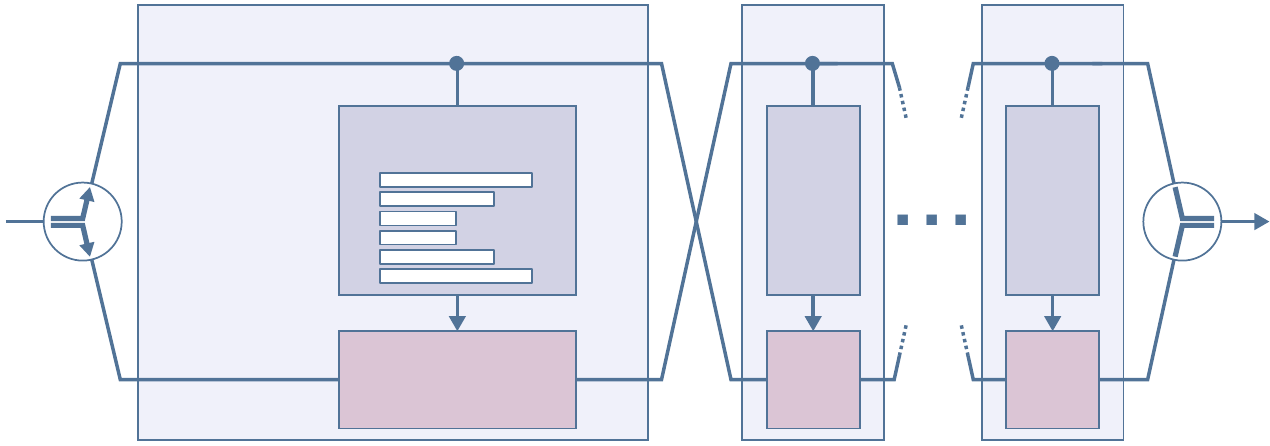}
        \put(19, 36){\small Coupling layer $\map_1$}
        \put(62.5, 36){\small $\map_2$}
        \put(82, 36){\small $\map_\nlayers$}
        \put(1, 15){\footnotesize $\layerIn$}
        \put(98, 15){\footnotesize $z$}
        \put(12, 31.5){\footnotesize Partition $\partitionA$}
        \put(12,  6.5){\footnotesize Partition $\partitionB$}
        \put(12, 27){\footnotesize $\layerIn^\partitionA$}
        \put(12, 2){\footnotesize $\layerIn^\partitionB$}
        \put(46.7, 27){\footnotesize $\layerOut^\partitionA$}
        \put(46.7, 2){\footnotesize $\layerOut^\partitionB$}
        \put(32, 22.5){\footnotesize $\nnet(\layerIn^\partitionA)$}
        \put(27.5, 4.5){\footnotesize $\cmap(\layerIn^\partitionB;\nnet(\layerIn^\partitionA))$}
    \end{overpic}
    \vspace{-4mm}%
    \caption{
        A coupling layer splits the input $\layerIn$ into two partitions $\partitionA$ and $\partitionB$. One partition is left untouched, whereas dimensions in the other partition are warped using a parametric coupling transform $\cmap$ driven by the output of a neural network $\nnet$. Multiple coupling layers may need to be compounded to achieve truly expressive transforms.%
        \label{fig:coupling-layers}
    }
    \vspace{-1mm}
\end{figure}

\subsection{Coupling Layers}
A single coupling layer takes a $D$-dimensional vector and partitions its dimensions into two groups. It leaves the first group intact and uses it to parameterize the transformation of the second group.

\begin{definition}[Coupling layer]
Let $\layerIn\in\R^D$ be an input vector, $\partitionA$ and $\partitionB$ denote disjoint partitions of $\ldb 1, D\rdb$, and $\nnet$ be a function on $\R^{\vert\partitionA\vert}$, then the output of a coupling layer $\layerOut=(\layerOut^\partitionA,\layerOut^\partitionB)=\map(\layerIn)$ is defined as
\begin{align}
\layerOut^\partitionA &= \layerIn^\partitionA \,, \\
\layerOut^\partitionB &= \cmap\big(\layerIn^\partitionB; \nnet(\layerIn^\partitionA)\big) \,,
\end{align}
where the \emph{coupling transform} $\cmap : \R^{\vert\partitionB\vert} \times \nnet(\R^{\vert\partitionA\vert}) \rightarrow \R^{\vert\partitionB\vert}$ is a separable and invertible map.
\end{definition}

The invertibility of the coupling transform, and the fact that partition~$\partitionA$ remains unchanged, enables a trivial inversion of the coupling layer $\layerIn = \map^{-1}(\layerOut)$ as:
\begin{align}
\layerIn^\partitionA &= \layerOut^\partitionA \,, \label{eq:cmap-inverse-a} \\
\layerIn^\partitionB &= \cmap^{-1}\big(\layerOut^\partitionB; \nnet(\layerIn^\partitionA)\big) = \cmap^{-1}\big(\layerOut^\partitionB; \nnet(\layerOut^\partitionA)\big) \,. \label{eq:cmap-inverse-b}
\end{align}
\ADD{If partition~$\partitionA$ was allowed to change arbitrarily, then the inversion (precisely the input to $m$ in \autoref{eq:cmap-inverse-b}\!) would be difficult to find.}
The invertibility is crucial in our setting as we require both \ADD{density evaluation} and sample generation in Monte Carlo integration.

The second important property of $\cmap$ is separability.
Separable $\cmap$ ensures that the Jacobian matrix is triangular and the determinant reduces to the product of diagonal terms; see \citet{dinh2014nice} or \autoref{app:coupling-layer-determinant} for a full treatment.
The computation of the determinant thus scales linearly with $D$ and is therefore tractable even in high-dimensional problems.

\subsection{Affine Coupling Transforms}%
\label{sec:affine-coupling-transforms}

\paragraph{Additive Coupling Transform.}
\citet{dinh2014nice} describe a very simple coupling transform that merely translates the signal in individual dimensions of $\partitionB$:
\begin{align}
    \cmap(\layerIn^\partitionB; t) &= \layerIn^\partitionB + t \,,
\end{align}
where the translation vector $t \in \R^{\vert\partitionB\vert}$ is produced by function $\nnet(\layerIn^\partitionA)$.

\paragraph{Multiply-add Coupling Transform.}
Since additive coupling layers have unit Jacobian determinants, i.e.\ they preserve volume,
\citet{dinh2016density} propose to add a multiplicative factor $e^s$:
\begin{align}
    \cmap(\layerIn^\partitionB; s,t) &= \layerIn^\partitionB \odot e^s + t \,,
\end{align}
where $\odot$ represents element-wise multiplication and vectors $t$ and $s \in \R^{\vert\partitionB\vert}$ are produced by $\nnet$: $(s,t) = \nnet(\layerIn^\partitionA)$.
The Jacobian determinant of a multiply-add coupling layer is simply $e^{\sum s_i}$.

The coupling transforms above are relatively simple. The trick that enables learning \emph{nonlinear} dependencies across partitions is the parametric function $\nnet$.
This function can be arbitrarily complex, e.g.\ a neural network, as we do not need its inverse to invert the coupling layer and its Jacobian does not affect the determinant of the coupling layer (cf.\ \autoref{app:coupling-layer-determinant}).
Using a sophisticated $\nnet$ allows extracting complex nonlinear relations between the two partitions. The coupling transform $\cmap$, however, remains simple, invertible, and permits tractable computation of determinants even in high-dimensional settings.

\subsection{Compounding Multiple Coupling Layers}
As mentioned initially, the complete transform between the data space and the latent space is obtained by chaining a number of coupling layers.
A different instance of neural network $\nnet$ is trained for each coupling layer.
To ensure that all dimensions can be modified, the output of one layer is fed into the next layer with the roles of the two partitions swapped; see \autoref{fig:coupling-layers}.
Compounding two coupling layers in this manner ensures that every dimension can be altered.

The number of coupling layers required to ensure that each dimension can influence every other dimension depends on the total number of dimensions.
For instance, in a 2D setting (where each partition contains exactly one dimension) we need only two coupling layers. 3D problems require three layers, and for any high-dimensional configuration there must be at least four coupling layers.

In practice, however, high-dimensional problems (e.g.\ generating images of faces), require significantly more coupling layers since each affine transform is fairly limited.
In the next section, we address this limitation by providing more expressive mappings that allow reducing the number of coupling layers and thereby the sample-generation and \ADD{density-evaluation} costs.
This improves the performance of Monte Carlo estimators presented in \autoref{sec:path-guiding}.

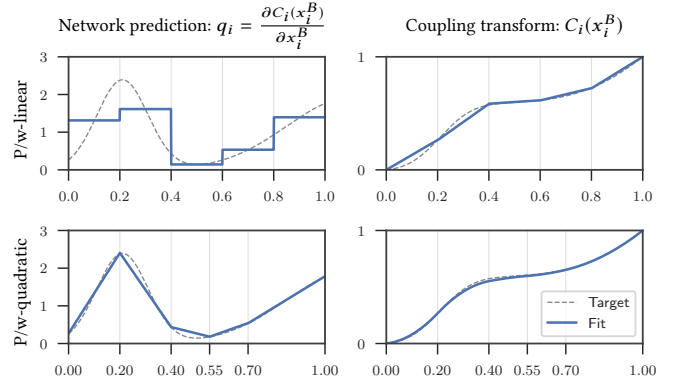
\begin{figure}
    \vspace{5mm}
    \begin{overpic}[width=\columnwidth]{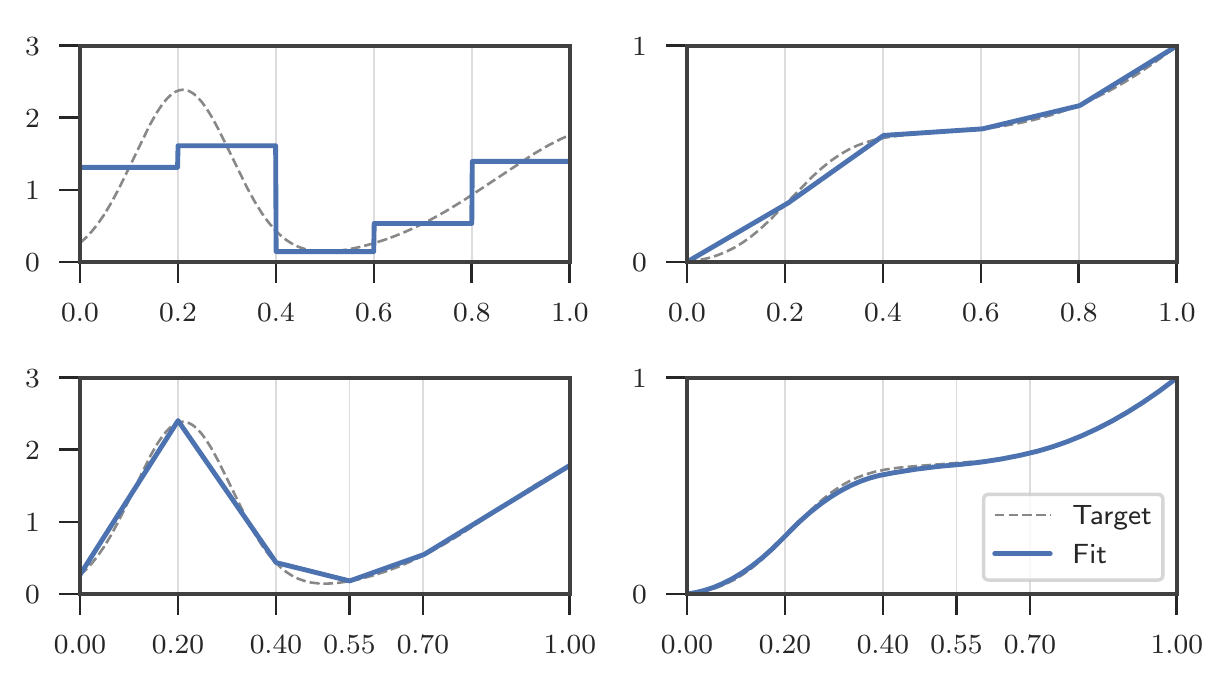}
    \put(5, 56){\footnotesize {Network prediction:} $\PdfPoly_i=\frac{\partial \cmap_i(\layerIn^\partitionB_i)}{\partial \layerIn^\partitionB_i}$}
    \put(59, 56){\footnotesize {Coupling transform:} $\cmap_i(\layerIn^\partitionB_i)$}
    \put(-2,36){\rotatebox{90}{\footnotesize {P/w-linear}}}
    \put(-2,7){\rotatebox{90}{\footnotesize {P/w-quadratic}}}
    \end{overpic}
    \hspace{-4.5mm}
    \vspace{-2mm}
    \caption{
        Predicted probability density functions (PDFs, left) and corresponding cumulative distribution functions (CDFs, right) with $\nbins = 5$ bins fitted to a target distribution (dashed).
        The top row illustrates a piecewise-linear CDF and the bottom row a piecewise-quadratic CDF.\@
        The piecewise-quadratic approximation tends to work better in practice thanks to its first-order continuity ($C^1$) and adaptive bin sizing.
        In \autoref{app:piecewise-no-bin-optimization} we show that, in contrast to piecewise-quadratic CDFs, adaptive bin sizing is difficult to achieve for piecewise-linear CDFs with gradient-based optimization methods.%
        \label{fig:piecewise}
    }
\end{figure}

\section{Piecewise-polynomial Coupling Layers}%
\label{sec:piecewise-poly}

In this section, we propose piecewise-polynomial invertible maps as coupling transforms instead of the limited affine warps reviewed previously.
\ADD{Specifically, we introduce the usage of piecewise polynomials with degrees $1$ and $2$, i.e.\ piecewise-linear and piecewise-quadratic warps.}
In contrast to \citet{dinh2014nice,dinh2016density}, who assume ${\layerIn,\layerOut\in {(-\infty,+\infty)}^D}$ and Gaussian latent variables, we choose to operate in a unit hypercube (i.e.\ ${\layerIn,\layerOut\in {[0,1]}^D}$) with uniformly distributed latent variables, as most practical problems span a finite domain.
Unbounded domains can still be handled by warping the input of $\map_1$ and the output of $\map_\nlayers$ e.g.\ using the sigmoid and logit functions.

Similarly to Dinh and colleagues, we ensure computationally tractable Jacobians via separability. \ADD{We transform each dimension independently:
\begin{align}
    \cmap\big( \layerIn^\partitionB; \nnet(\layerIn^\partitionA) \big) = {\Big(\cmap_1\big( \layerIn_1^\partitionB; \nnet(\layerIn^\partitionA) \big)\,,\, \cdots \,,\, \cmap_{\vert\partitionB\vert}\big( \layerIn_{\vert\partitionB\vert}^\partitionB; \nnet(\layerIn^\partitionA) \big) \Big)}^T \,.
\end{align}}
Operating on unit intervals allows interpreting the warping function $\cmap_\dimId$ as a cumulative distribution function (CDF).
To produce each $\cmap_\dimId$, we instrument the neural network to output the corresponding unnormalized probability density $\PdfPoly_\dimId$, and construct $\cmap_\dimId$ by integration; see \autoref{fig:piecewise} for an illustration.

In order to further improve performance, we propose to encode the inputs to the neural network using \emph{one-blob encoding}, which we discuss in \autoref{sec:one-blob-encoding}.

\begin{figure}
    \vspace{2mm}
    \begin{overpic}[width=1.07\columnwidth]{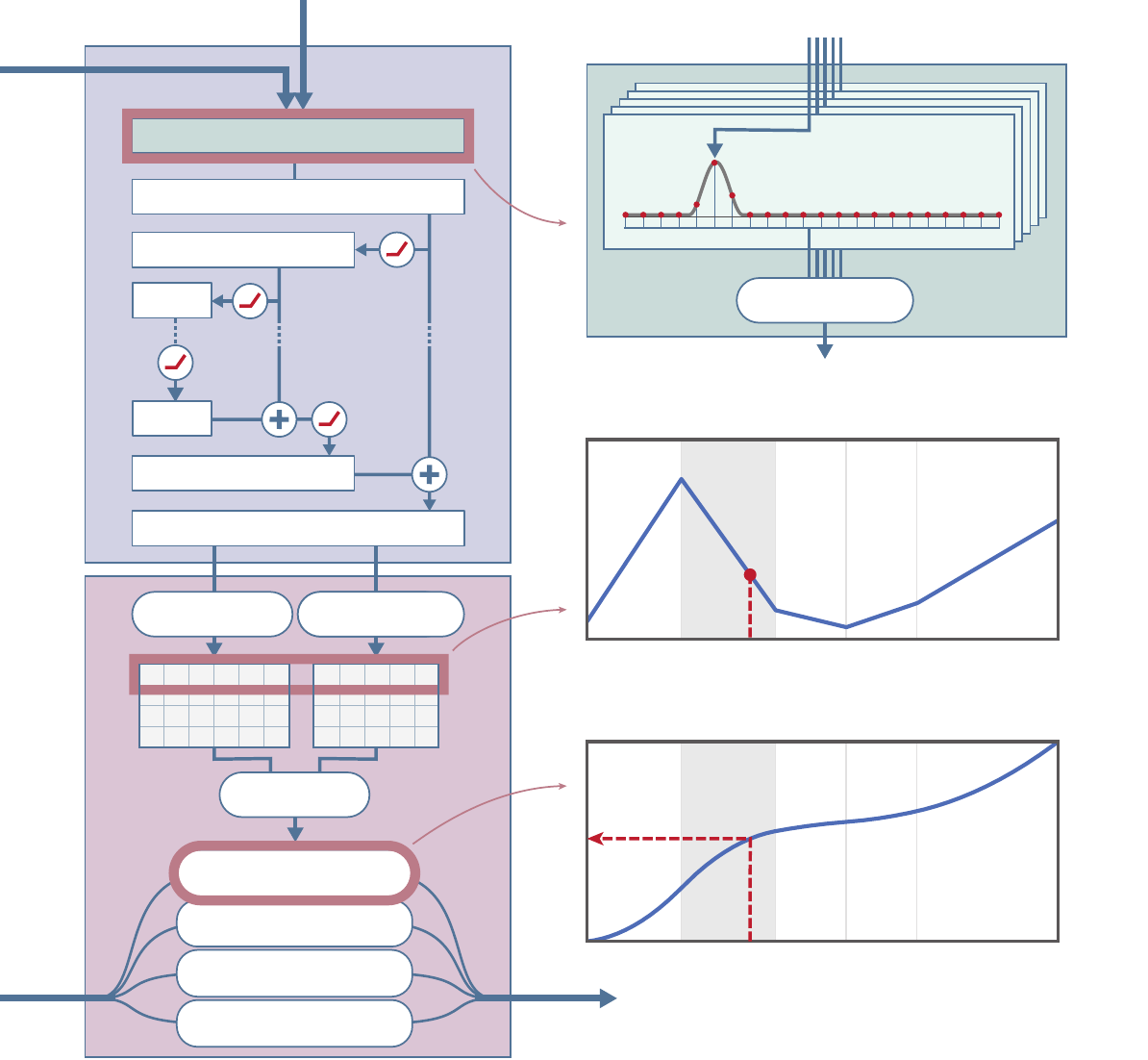}
        \put(3,48){\rotatebox{90}{\footnotesize {U-shape network $\nnet$}}}
        \put(3,9){\rotatebox{90}{\footnotesize {Coupling transforms}}}
        \put(-3,83.5){\footnotesize optional}
        \put(-1,81){\footnotesize extra}
        \put(-3,78.5){\footnotesize features}
        \put(27.2,89.5){\footnotesize $\layerIn^\partitionA$}
        \put(61.25,89.5){\footnotesize One-blob encoding}
        \put(58.5,55.5){\footnotesize Piecewise-linear PDF $\PdfPoly_1$}
        \put(64,33.4){\footnotesize $\layerIn^\partitionB_1$}
        \put(60.7,51){\footnotesize bin $\containingBinId$}
        \put(55.75,29){\footnotesize Piecewise-quadratic warp $\cmap_1$}
        \put(64,7){\footnotesize $\layerIn^\partitionB_1$}
        \put(47,19){\footnotesize $\layerOut^\partitionB_1$}
        \put(60.7,24.5){\footnotesize bin $\containingBinId$}
        \put(17.5,29.2){\footnotesize $\matV$}
        \put(31.5,29.2){\footnotesize $\matW$}
        \put(0,2){\footnotesize $\layerIn^\partitionB$}
        \put(51,2){\footnotesize $\layerOut^\partitionB$}
        \put(13,38.1){\footnotesize normalize}
        \put(27.6,38.1){\footnotesize normalize}
        \put(20.7,22.3){\footnotesize integrate}
        \put(65.3,65.5){\footnotesize concatenate}
        \put(21.3,15.5){\footnotesize $\cmap_1(\layerIn^\partitionB_1)$}
        \put(21.3,11){\footnotesize $\cmap_2(\layerIn^\partitionB_2)$}
        \put(21.3,6.9){\footnotesize $\cmap_3(\layerIn^\partitionB_3)$}
        \put(21.3,2.4){\footnotesize $\cmap_4(\layerIn^\partitionB_4)$}
        \put(32.2,67.5){\tiny ReLU}
    \end{overpic}
    \caption{
        Our coupling layer with a piecewise-quadratic transform for $|B| = 4$. Signals in partition $\partitionA$ (and additional features) are encoded using one-blob encoding and fed into a U-shape neural network $\nnet$ with fully connected layers. The outputs of $\nnet$ are normalized yielding matrices $\matV$ and $\matW$ that define warping PDFs. The PDFs are integrated analytically to obtain piecewise-quadratic coupling transforms; one for warping each dimension of~$\layerIn^\partitionB$.%
        \label{fig:architecture-detail}
    }
\end{figure}
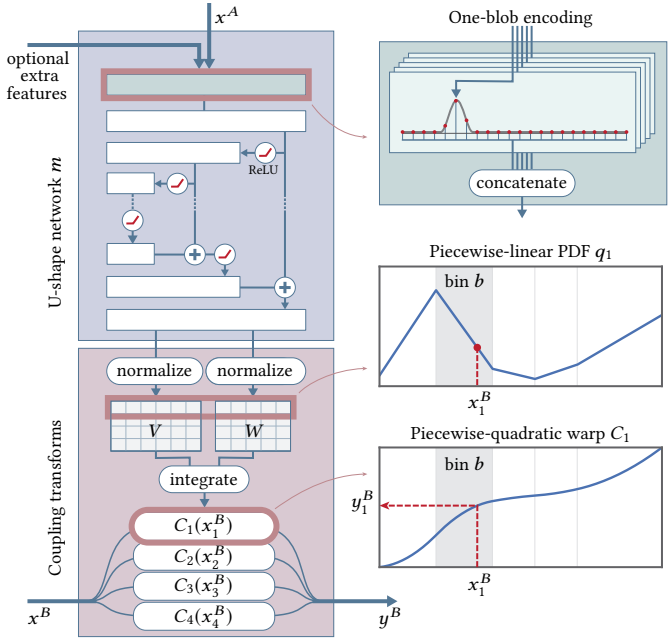

\begin{figure*}[t]
    \setlength{\tabcolsep}{1pt}%
    \renewcommand{\arraystretch}{0.25}
    \begin{tabular}{cccccccl}
        \multicolumn{3}{c}{\footnotesize \citet{dinh2016density}} & \multicolumn{2}{c}{\footnotesize Ours ($\nlayers$=2)} & \\
        \cmidrule(lr){1-3}  \cmidrule(lr){4-5}\\[-2pt]
        \footnotesize Affine ($\nlayers$=2) &
        \footnotesize Affine ($\nlayers$=4) &
        \footnotesize Affine ($\nlayers$=16) &
        \footnotesize {P/w-linear} &
        \footnotesize {P/w-quadratic} &
        \footnotesize Reference &
        \footnotesize \hspace{0.6cm} KL divergence &
        \footnotesize \hspace{1.25cm} Variance \\
        & & & & & & \multicolumn{2}{c}{\multirow{3}{*}{\includegraphics[width=0.326\linewidth]{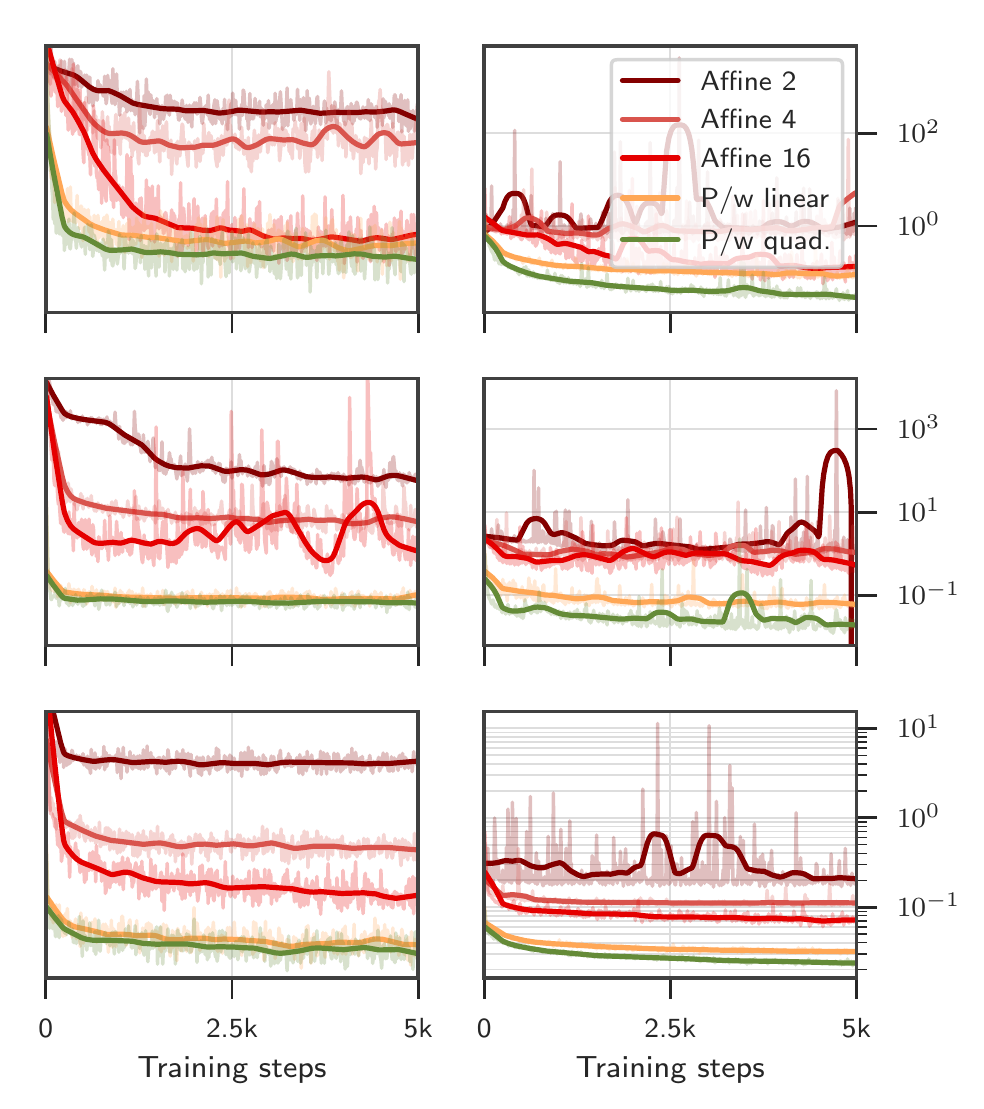}}} \\[0pt]
        \includegraphics[width=0.11\linewidth]{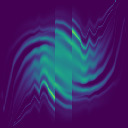} &
        \includegraphics[width=0.11\linewidth]{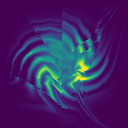} &
        \includegraphics[width=0.11\linewidth]{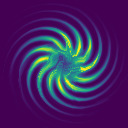} &
        \includegraphics[width=0.11\linewidth]{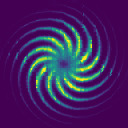} &
        \includegraphics[width=0.11\linewidth]{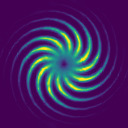} &
        \includegraphics[width=0.11\linewidth]{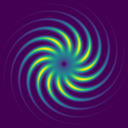} & \multicolumn{2}{c}{} \\
        \includegraphics[width=0.11\linewidth]{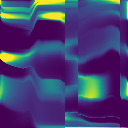} &
        \includegraphics[width=0.11\linewidth]{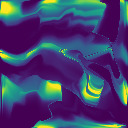} &
        \includegraphics[width=0.11\linewidth]{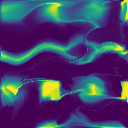} &
        \includegraphics[width=0.11\linewidth]{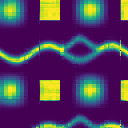} &
        \includegraphics[width=0.11\linewidth]{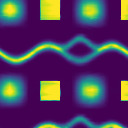} & 
        \includegraphics[width=0.11\linewidth]{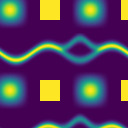} & \multicolumn{2}{c}{} \\
        \includegraphics[width=0.11\linewidth]{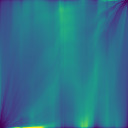} &
        \includegraphics[width=0.11\linewidth]{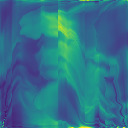} &
        \includegraphics[width=0.11\linewidth]{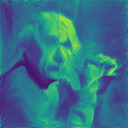} &
        \includegraphics[width=0.11\linewidth]{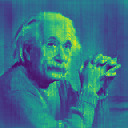} &
        \includegraphics[width=0.11\linewidth]{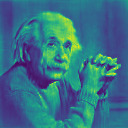} & 
        \includegraphics[width=0.11\linewidth]{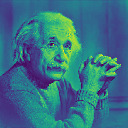} & \multicolumn{2}{c}{} \\
    \end{tabular}
    \caption{
        Our $32$-bin piecewise-linear ($4$-th column) and 32-bin piecewise-quadratic ($5$-th column) coupling layers achieve superior performance compared to affine (multiply-add) coupling layers~\citep{dinh2016density} on low-dimensional \ADD{regression} problems.
        The false-colored distributions were obtained by optimizing KL divergence with uniformly drawn \ADD{i.i.d.}\ samples (weighted by the reference value) over the 2D image domain.
        The plots on the right show logarithmically scaled training error (KL divergence) and the variance of \ADD{estimating the average image intensity when drawing samples from one of the distributions}.
    }%
    \label{fig:2d-dists}
\end{figure*}

\begin{figure*}
    \setlength{\tabcolsep}{1pt}%
    \renewcommand{\arraystretch}{0.25}
    \begin{tabular}{ccccccc}
        \multicolumn{2}{c}{\footnotesize Affine ($\nlayers$=16)} &
        \multicolumn{2}{c}{\footnotesize Piecewise-linear ($\nlayers$=2)} &
        \multicolumn{2}{c}{\footnotesize Piecewise-quadratic ($\nlayers$=2)} \\[2pt]
        \cmidrule(lr){1-2}  \cmidrule(lr){3-4} \cmidrule(lr){5-6}
        \footnotesize  scalar encoding &
        \footnotesize  one-blob encoding &
        \footnotesize  scalar encoding &
        \footnotesize  one-blob encoding &
        \footnotesize  scalar encoding &
        \footnotesize  one-blob encoding &
        \footnotesize  Reference\\[4pt]
        \includegraphics[width=0.139\linewidth]{images/2d-dists/image-affine16.jpg} &
        \includegraphics[width=0.139\linewidth]{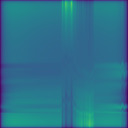} &
        \includegraphics[width=0.139\linewidth]{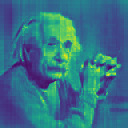} & 
        \includegraphics[width=0.139\linewidth]{images/2d-dists/image-const2.jpg} &
        \includegraphics[width=0.139\linewidth]{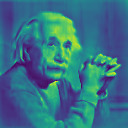} & 
        \includegraphics[width=0.139\linewidth]{images/2d-dists/image-linear2.jpg} &
        \includegraphics[width=0.139\linewidth]{images/2d-dists/image-gt.jpg} 
        \\
    \end{tabular}
    \vspace{-1mm}
    \caption{
        Comparison of results \emph{with} and \emph{without} the one-blob encoding.
        The experimental setup is the same as in \autoref{fig:2d-dists}.
        While the affine coupling transforms fail to converge with one-blob-encoded inputs, the distributions learned by the piecewise-polynomial coupling functions become sharper and more accurate.
    }%
    \label{fig:one-blob-study}
\end{figure*}

\subsection{Piecewise-Linear Coupling Transform}%
\label{sec:piecewise-linear-coupling-function}

Motivated by their simplicity, we begin by investigating the simplest continuous piecewise-polynomial coupling transforms: piecewise-linear ones.
Recall that we partition the $D$-dimensional input vector in two disjoint groups, A and B, such that $\layerIn = (\layerIn^\partitionA, \layerIn^\partitionB)$.
We divide the unit dimensions in partition $\partitionB$ into $\nbins$ bins of equal width $\binWidth=\nbins^{-1}$.
To define all $\vert\partitionB\vert$ transforms at once, we instrument the network $\nnet(\layerIn^\partitionA)$ to predict a $\vert\partitionB\vert \times \nbins$ matrix, denoted $\raw{\matQ}$.
Each $\dimId$-th row of $\raw{\matQ}$ defines the unnormalized probability mass function (PMF) of the warp in $\dimId$-th dimension in $\layerIn^\partitionB$;
we normalize the rows using the softmax function $\softmax$ and denote the normalized matrix $\matQ$; $\matQ_\dimId = \softmax(\raw{\matQ}_\dimId)$.

The PDF in $\dimId$-th dimension is then defined as $\PdfPoly_\dimId(\layerIn_\dimId^\partitionB) = \matQ_{\dimId \containingBinId} / \binWidth$, where $\containingBinId = \lfloor \nbins \layerIn_\dimId^\partitionB\rfloor$ is the bin that contains the scalar value $\layerIn_\dimId^\partitionB$.
We integrate the PDF to obtain our invertible piecewise-linear warp $\cmap_\dimId$:
\begin{align}
\cmap_\dimId(\layerIn_\dimId^\partitionB; \matQ) = \int_0^{\layerIn_\dimId^\partitionB} \PdfPoly_\dimId(t) \Diff t =
\alpha
\matQ_{\dimId \containingBinId} +
\sum_{k=1}^{\containingBinId - 1} \matQ_{\dimId k} \,,
\end{align}
where $\alpha = \nbins \layerIn_\dimId^\partitionB - \lfloor \nbins \layerIn_\dimId^\partitionB \rfloor$ represents the relative position of $\layerIn_\dimId^\partitionB$ in~$\containingBinId$.

In order to evaluate the change of density resulting from the coupling layer, we need to compute the determinant of its Jacobian matrix; see~\autoref{eq:change-of-variable}.
Since $\cmap(\layerIn^\partitionB; \matQ)$ is separable by definition, its Jacobian matrix is diagonal and the determinant is equal to the product of the diagonal terms. These can be computed using $\matQ$:
\begin{align}
\det\left( \frac{\partial \cmap\big(\layerIn^\partitionB; \matQ\big)}{\partial {(\layerIn^\partitionB)}^T} \right)
= \prod_{\dimId=1}^{|\partitionB|} \PdfPoly_\dimId(\layerIn_\dimId^\partitionB)
= \prod_{\dimId=1}^{|\partitionB|} \frac{\matQ_{\dimId \containingBinId}}{\binWidth} \,,
\end{align}
where $\containingBinId$ again denotes the bin containing the value in the $\dimId$-th dimension.
To reduce the number of bins $\nbins$ required for a good fit we would like the network to also predict bin widths.
These can unfortunately \emph{not} easily be optimized with gradient descent in the piecewise-\emph{linear} case; see \autoref{app:piecewise-no-bin-optimization}.
To address this, and to improve accuracy, we propose piecewise-\emph{quadratic} coupling transforms.

\begin{figure*}[t]
    \setlength{\tabcolsep}{2pt}%
    \begin{tabular}{ccc}
        \small Examples from the training set &
        \small Generated novel images &
        \small Manifold spanned by four images \\
        \includegraphics[height=5.02cm,trim={0 128px 0 0},clip]{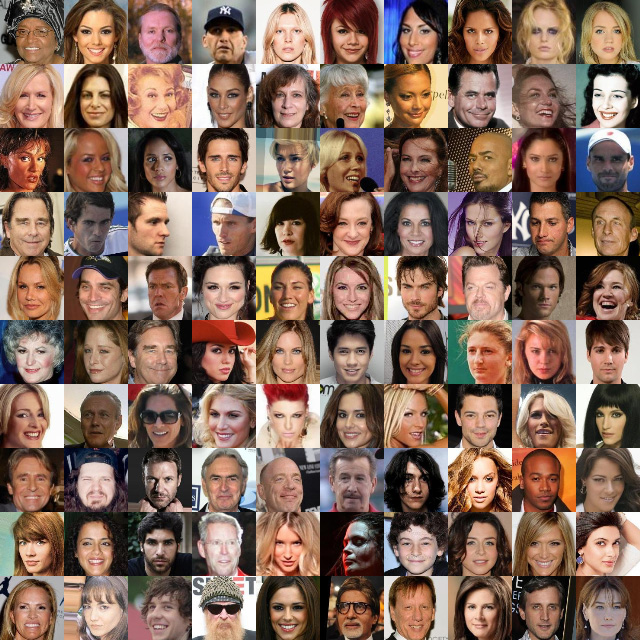} &
        \includegraphics[height=5.02cm,trim={0 128px 0 0},clip]{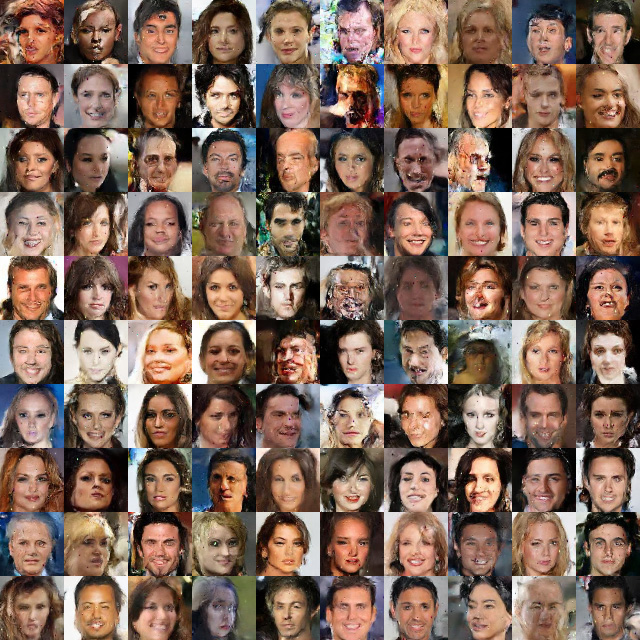} &
        \includegraphics[height=5.02cm]{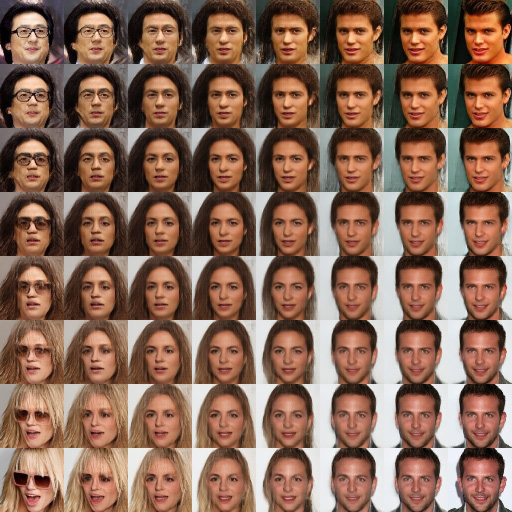}
    \end{tabular}
    \vspace{-1mm}
    \caption{
        Generative modeling of facial photographs using the architecture of \citet{dinh2016density} with our piecewise-quadratic coupling transform.
        We show training examples (left), faces generated by our trained model (middle), and a manifold of faces spanned by linear interpolation of 4 training examples in latent space (right; training examples are in the corners).
        We achieve \ADD{validation} results of slightly better quality than \citet{dinh2016density} in terms of negative log-likelihood \ADD{($2.89$ vs.\ $3.02$ bits per dimension)}, suggesting that our approach could also benefit high-dimensional problems, though one may not achieve the same magnitude of improvements as in low-dimensional settings.}%
    \label{fig:faces}
\end{figure*}

\subsection{Piecewise-Quadratic Coupling Transform}
Piecewise-quadratic coupling transforms admit a piecewise-linear PDF, which we model using $\nbins+1$ vertices; see \autoref{fig:piecewise}, bottom left.
We store their vertical coordinates (for all dimensions in $\partitionB$) in $\vert\partitionB\vert \times (\nbins+1)$ matrix $\matV$, and horizontal differences between neighboring vertices (bin widths) in $\vert\partitionB\vert \times \nbins$ matrix $\matW$.

The network $\nnet$ outputs unnormalized matrices $\raw{\matW}$ and $\raw{\matV}$.
We again normalize the matrices using the standard softmax ${\matW_\dimId = \softmax(\raw{\matW}_\dimId)}$, and an adjusted version in the case of $\matV$:
\begin{align}
\matV_{\dimId,\binId} =
\frac{\expFunc{\raw{\matV}_{\dimId,\binId}}}
{\sum_{k=1}^{K} \frac{\expFunc{\raw{\matV}_{\dimId,k}} + \expFunc{\raw{\matV}_{\dimId,k+1}}}{2} \matW_{\dimId, k}} \,,
\end{align}
where the denominator ensures that $\matV_\dimId$ represents a valid PDF.\@

The PDF in dimension $i$ is defined as
\begin{align}
\PdfPoly_\dimId(\layerIn_\dimId^\partitionB) =
\lerpFunc{\matV_{\dimId\containingBinId}}{\matV_{\dimId \containingBinId+1}}{\alpha} \,,
\label{eq:piecewise-linear-pdf}
\end{align}
where $\alpha = (\layerIn_\dimId^\partitionB - \sum_{k=1}^{\containingBinId-1} \matW_{\dimId k}) / \matW_{\dimId \containingBinId}$
represents the relative position of scalar $\layerIn_\dimId^\partitionB$ in bin $\containingBinId$ that contains it, i.e.\
${\sum_{k=1}^{\containingBinId-1} \matW_{\dimId k}
\leq \layerIn_\dimId^\partitionB <
\sum_{k=1}^{\containingBinId} \matW_{\dimId k}}$.

The invertible coupling transform is obtained by integration:
\begin{align}
\cmap_\dimId(\layerIn_\dimId^\partitionB; \matW, \matV) &=
\frac{\alpha^2}{2} (\matV_{\dimId \containingBinId+1}-\matV_{\dimId \containingBinId}) \matW_{\dimId \containingBinId}
+ \alpha \matV_{\dimId\containingBinId} \matW_{\dimId \containingBinId}
\\ &+ \sum_{k=1}^{\containingBinId - 1}
\frac{\matV_{\dimId k}+\matV_{\dimId k+1}}{2} \matW_{\dimId k} \,.
\end{align}
Note that inverting $\cmap_\dimId(\layerIn_\dimId^\partitionB; \matW, \matV)$ involves solving the root of the quadratic term, which can be done efficiently and robustly.

Computing the determinant of the Jacobian matrix follows the same logic as in the piecewise-linear case, with the only difference being that we must now interpolate the entries of $\matV$ in order to obtain the PDF value at a specific location (cf. \autoref{eq:piecewise-linear-pdf}).

\subsection{One-Blob Encoding}%
\label{sec:one-blob-encoding}

An important consideration is the encoding of the inputs to the network.
We propose to use the \emph{one-blob} encoding---a generalization of the \emph{one-hot} encoding~\cite{HARRIS201354}---where a kernel is used to activate multiple adjacent entries instead of a single one.
Assume a scalar $s \in [0,1]$ and a quantization of the unit interval into $k$ bins (we use $k=32$).
The one-blob encoding amounts to placing a kernel (we use a Gaussian with $\sigma=1/k$) at $s$ and discretizing it into the bins.
With the proposed architecture of the neural network (placement of ReLUs in particular, see \autoref{fig:architecture-detail}), the one-blob encoding effectively shuts down certain parts of the linear path of the network, allowing it to specialize the model on various sub-domains of the input.

In contrast to one-hot encoding, where the quantization causes a loss of information if applied to continuous variables, the one-blob encoding is lossless; it captures the exact position of $s$.

\subsection{Analysis}
We compare the proposed piecewise-polynomial coupling transforms to multiply-add affine transforms~\cite{dinh2016density} on a 2D \ADD{regression} problem in \autoref{fig:2d-dists}.
To produce columns 1--5, we sample the 2D domain using \ADD{\emph{uniform} i.i.d.\ samples (\num{16384} samples per training step)}, evaluate the reference function (column 6) at each sample, and optimize the neural networks that control the coupling transforms using KL divergence described in \autoref{sec:optimizing-kl}.
\ADD{We also ran the same experiment with equally weighted i.i.d.\ samples drawn \emph{proportional} to the reference function---i.e.\ in a density-estimation setting---producing near-identical results (not shown).}
Every per-layer network has a U-net (see \autoref{fig:architecture-detail}) with $8$ fully connected layers, where the outermost layers contain $256$ neurons and the number of neurons is halved at every nesting level. We use $2$ additional layers to adapt the input and output dimensionalities to and from $256$, respectively.
The networks only differ in their output layer to produce the desired parameters of their respective coupling transform ($s$ and $t$, $\raw{\matQ}$, or $\raw{\matW}$ and $\raw{\matV}$).

We use adaptive bin sizes only in the piecewise-quadratic coupling transforms because gradient descent fails to optimize them in the piecewise-linear case as demonstrated in \autoref{app:piecewise-no-bin-optimization}.

When using $\nlayers=2$ coupling layers---i.e.\ $2\times 10$ fully connected layers---the piecewise-polynomial coupling layers consistently perform better thanks to their significantly larger modeling power,
and outperform even large numbers (e.g.\ $\nlayers=16$) of multiply-add coupling layers, amounting to $16\times 10$ fully connected layers.

\autoref{fig:one-blob-study} demonstrates the benefits of the one-blob encoding when combined with our piecewise coupling transforms.
While the encoding helps our coupling transforms to learn sharp, non-linear functions more easily, it also causes the multiply-add transforms of \citet{dinh2016density} to produce excessive high frequencies that inhibit convergence.
Therefore, in the rest of the paper we use the one-blob encoding only with our piecewise-polynomial transforms; results with affine transforms do not utilize one-blob encoded inputs. 

We tested the piecewise-quadratic coupling layers also on a high-dimensional density-estimation problem: learning the manifold of a specific class of natural images. 
We used the CelebFaces Attributes dataset~\cite{Liu:2015} and reproduced the experimental setting of \citet{dinh2016density}.
Our architecture is based on the authors' publicly available implementation and differs only in the used coupling layer and the depth of the network---we use $4$ recursive subdivisions while the authors use $5$, resulting in $28$ versus $35$ coupling layers.
We chose $\nbins=4$ bins and \emph{did not} use our one-blob encoding due to GPU memory constraints.
Since our coupling layers operate on ${[0, 1]}^D$, we do not use batch normalization on the transformed data.

\autoref{fig:faces} shows a sample of the training set, a sample of generated images, and a visualization of the manifold given by four different faces.
The visual quality of our results is comparable to those obtained by Dinh and colleagues. We perform marginally better \ADD{in terms of negative log-likelihood} (lower means better): we yield $2.85$ and $2.89$ bits per dimension on training and validation data, respectively, whereas \citet{dinh2016density} reported $2.97$ and $3.02$ bits per dimension.
We tried decreasing the number of coupling layers while increasing the number of bins within each of them, but the results became overall worse.
We hypothesize that the high-dimensional problem of learning distributions of natural images benefits more from having many coupling layers rather than having fewer but expressive ones.

\section{Monte Carlo Integration with NICE}%
\label{sec:mc}

In this section, we apply the NICE framework to Monte Carlo integration. Our goal is to reduce estimation variance by extracting sampling PDFs from observations of the integrand.
Denoting $\PdfOptimized(x;\Params)$ the to-be-learned PDF for drawing samples ($\Params$ represents the trainable parameters) and $\PdfGt(x)$ the ground-truth distribution of the integrand, we can rewrite the MC estimator from \autoref{eq:basic-mc-estimator} as
\begin{align}
\langle F \rangle_{N} = \frac{1}{N}  \sum_{i=1}^{N}\frac{f(X_i)}{\PdfOptimized(X_i;\Params)}=\frac{1}{N}  \sum_{i=1}^{N}\frac{\PdfGt(X_i)\,F}{\PdfOptimized(X_i;\Params)} \,.
\label{eq:mc-estimator-with-pgt}
\end{align}
In the ideal case when $\PdfOptimized(x;\Params) = \PdfGt(x)$, the estimator returns the exact value of $F$.
Our objective here is to leverage NICE to learn $\PdfOptimized$ \emph{from data}
while optimizing the neural networks in coupling layers 
so that the distance between $\PdfGt$ and $\PdfOptimized$ is minimized.

We follow the standard approach of quantifying the distance using one of the commonly used divergence metrics.
While all divergence metrics reach their minimum if both distributions are equal, they differ in shape and therefore produce different $\PdfOptimized$ in practice.

In~\autoref{sec:optimizing-kl}, we optimize using the popular Kullback-Leibler (KL) divergence.
We further consider directly minimizing the variance of the resulting MC estimator in \autoref{sec:optimizing-var} and demonstrate that it is equivalent to minimizing the $\chi^2$ divergence.

\subsection{Minimizing Kullback-Leibler Divergence}%
\label{sec:optimizing-kl}

Most generative models based on deep neural networks do not allow evaluating the likelihood $\PdfOptimized(x;\Params)$ of data points $x$ exactly and/or efficiently.
In contrast, our work is based on bijective mappings with tractable Jacobian determinants that easily permit such evaluations.
In the following, we show that minimizing the KL divergence with gradient descent amounts to maximizing a weighted log likelihood.

The KL divergence between $\PdfGt(x)$ and the learned $\PdfOptimized(x;\Params)$ reads
\begin{align}
    \KlDiv(\PdfGt \, \| \, \PdfOptimized;\Params) 
    &= \!
    \int_\Omega \PdfGt(x) \log{\frac{\PdfGt(x)}{\PdfOptimized(x;\Params)}} \Diff{x}\nonumber \\
    &= \!
    {\int_\Omega \PdfGt(x) \log{\PdfGt(x)} \Diff{x}}
    \underbrace{- \int_\Omega \PdfGt(x) \log{\PdfOptimized(x;\Params)} \Diff{x}}_{\text{Cross entropy}} \,.
\end{align}
To minimize $\KlDiv$ with gradient descent, we need its gradient with respect to the trainable parameters $\Params$. These appear only in the cross-entropy term, hence
\begin{align}
    \nabla_\Params \KlDiv(\PdfGt \, \| \, \PdfOptimized;\Params) 
    &= -\nabla_\Params \int_\Omega \PdfGt(x) \log{\PdfOptimized(x;\Params)} \Diff{x}\label{eq:grad-kl}\\
    &= \Expectation\left[-\frac{\PdfGt(\Sample)}{\PdfOptimized(\Sample;\Params)} \nabla_\Params \log{\PdfOptimized(\Sample;\Params)}\right] \,,
    \label{eq:grad-kl-expectation}
\end{align}
where the expectation is over $\Sample \sim \PdfOptimized(x;\Params)$, i.e.\ the samples are drawn from the learned generative model%
\footnote{If samples could be drawn directly from the ground-truth distribution---as is common in computer vision problems---the stochastic gradient would simplify to that of just the log likelihood.
We discuss a generalization of log-likelihood maximization.
}.
In most integration problems, $\PdfGt(x)$ is only accessible in an unnormalized form through $f(x)$: $\PdfGt(x) = f(x) / F$. Since $F$ is unknown---this is what we are trying to estimate in the first place---the gradient can be estimated only up to the global scale factor $F$. 
This is not an issue since common gradient-descent-based optimization techniques such as Adam~\citep{KingmaB14} scale the step size by the reciprocal square root of the gradient variance, cancelling $F$.
Furthermore, if $f(x)$ can only be estimated via Monte Carlo, the gradient is still correct due to the linearity of expectations.
\autoref{eq:grad-kl-expectation} therefore shows that minimizing the KL divergence via gradient descent is equivalent to minimizing the negative log likelihood weighted by MC estimates of~$F$.

\subsection{Minimizing Variance via $\chi^2$ Divergence}%
\label{sec:optimizing-var}

The most attractive quantity to minimize in the context of (unbiased) Monte Carlo integration is the variance of the estimator.
Inspired by previous works that strive to directly minimize variance~\citep{DBLP:journals/corr/PantaleoniH17,Vevoda2018:BOR,herholz2016,Herholz:2018}, we demonstrate how this can be achieved for the MC estimator $\PdfGt(\Sample)/\PdfOptimized(\Sample;\Params)$, with $\Sample \sim \PdfOptimized(x;\Params)$, via gradient descent.
We begin with the variance definition and~simplify:
\begin{alignat}{2}
    \Variance\left[ \frac{\PdfGt(\Sample)}{\PdfOptimized(\Sample;\Params)} \right]
    &= \Expectation\left[ \frac{{\PdfGt(\Sample)}^2}{{\PdfOptimized(\Sample;\Params)}^2} \right]
    &&-{\Expectation\left[ \frac{\PdfGt(\Sample)}{\PdfOptimized(\Sample;\Params)} \right]}^2 \nonumber \\
    &= \int_\Omega \frac{{\PdfGt(x)}^2}{\PdfOptimized(x;\Params)} \Diff{x}
    &&- \underbrace{{\left( \int_\Omega \PdfGt(x) \Diff{x} \right)}^2}_1 \,.
    \label{eq:variance-loss}
\end{alignat}
The stochastic gradient of the variance for gradient descent is then
\begin{align}
    \nabla_\Params \Variance\left[ \frac{\PdfGt(\Sample)}{\PdfOptimized(\Sample;\Params)} \right]
    &= \nabla_\Params \int_\Omega \frac{{\PdfGt(x)}^2}{\PdfOptimized(x;\Params)} \Diff{x} \nonumber \\
    &= \int_\Omega {\PdfGt(x)}^2 \,\nabla_\Params \frac{1}{\PdfOptimized(x;\Params)} \Diff{x} \nonumber \\
    &= \int_\Omega -\frac{{\PdfGt(x)}^2}{\PdfOptimized(x;\Params)} \,\nabla_\Params \log{\PdfOptimized(x;\Params)} \Diff{x} \nonumber \\
    &= \Expectation \left[ -{\left(\frac{\PdfGt(\Sample)}{\PdfOptimized(\Sample;\Params)}\right)}^2 \nabla_\Params \log{\PdfOptimized(\Sample;\Params)} \right] \,.
\end{align}

\paragraph{Relation to the Pearson $\chi^2$ divergence}
Upon close inspection it turns out the variance objective (Equation~\ref{eq:variance-loss}) is equivalent to the Pearson $\chi^2$ divergence $\ChiDiv(\PdfGt \, \| \, \PdfOptimized;\Params)$:
\begin{align}
    \! \! \! \ChiDiv(\PdfGt \, \| \, \PdfOptimized;\Params)
    &= \int_\Omega \frac{{\left(\PdfGt(x) - \PdfOptimized(x;\Params)\right)}^2}{\PdfOptimized(x;\Params)} \Diff{x} \nonumber \\
    &= \int_\Omega \frac{{\PdfGt(x)}^2}{\PdfOptimized(x;\Params)} \Diff{x} 
    - \underbrace{\left(2 \int_\Omega \PdfGt(x) \Diff{x}
    - \int_\Omega \PdfOptimized(x;\Params) \Diff{x}\right)}_{1} .
\end{align}
As such, minimizing the variance of a Monte Carlo estimator amounts to minimizing the Pearson $\chi^2$ divergence between the ground-truth and the learned distributions.

\paragraph{Connection between the $\chi^2$ and KL divergences}
Notably, the gradients of the KL divergence and the $\chi^2$ divergence differ only in the weight applied to the log likelihood.
In $\nabla_\Params \KlDiv$ the log likelihood is weighted by the MC weight, whereas when optimizing $\nabla_\Params \ChiDiv$, the log likelihood is weighted by the \emph{squared} MC weight.
This illustrates the difference between the two loss functions: the $\chi^2$ divergence penalizes large discrepancies stronger, specifically, low values of $\PdfOptimized$ in regions of large density $\PdfGt$.
As such, it tends to produce more conservative $\PdfOptimized$ than $\KlDiv$, which we observe in \autoref{sec:path-guiding} as fewer outliers at the cost of slightly worse average performance.

\section{Neural Path Sampling and Path Guiding}%
\label{sec:path-guiding}

In this section, we take NICE (\autoref{sec:nice}) with piecewise-polynomial warps (\autoref{sec:piecewise-poly}) and apply it to sequential MC integration of light transport using the methodology described in \autoref{sec:mc}.
Our goal is to reduce estimation variance by ``guiding'' the construction of light paths using on-the-fly learned sampling densities.
We explore two different settings:
a global setting that leverages the path-integral formulation of light transport and employs high-dimensional sampling in the primary sample space (PSS) to build complete light-path samples (\autoref{sec:pss-path-sampling}),
and a local setting, natural to the rendering equation, where the integration spans a 2D (hemi-)spherical domain and the path is built incrementally (\autoref{sec:imcremental-path-guiding}).

\subsection{Primary-Sample-Space Path Sampling}%
\label{sec:pss-path-sampling}

In order to produce an image, a renderer must estimate the amount of light reaching the camera after taking any of the possible paths through the scene.
The transport can be formalized using the path-integral formulation~\cite{Veach:1997:Thesis}, where a radiance measurement $\Measurement$ to a sensor (e.g.\ a pixel) is given by an integral over path space~$\Paths$:
\begin{align}
    \Measurement &= \int_\Paths
    \emittedRadiance(\pos_0, \pos_1) \,
    \PathThroughput(\Path) \,
    \Importance(\pos_{k-1}, \pos_k)\,
    \Diff{\Path}\,.
    \label{eqn:path-integral}
\end{align}
The chain of positions $\Path = \pos_0 \cdots \pos_k$ represents a single light path with $k$ vertices. The path throughput $\PathThroughput(\Path)$ quantifies the ability of $\Path$ to transport radiance.
$\emittedRadiance$ represents emitted radiance and
$\Importance$ is the sensor response to one unit of incident radiance.

The measurement can be estimated as
\begin{align}
\langle \Measurement \rangle = \frac{1}{N} \sum_{j=1}^N \frac{\emittedRadiance(\pos_{j 0}, \pos_{j 1}) \,
    \PathThroughput(\Path_j) \,
    \Importance(\pos_{j k-1}, \pos_{j k})\,}
    {\PdfMC(\Path_j)}\, ,
\end{align}
where $\PdfMC(\Path)$ is the joint probability density of generating all $k$ vertices of path $\Path$.
Drawing samples from the joint distribution is challenging due to the constrained nature of vertices; e.g.\ they have to reside on surfaces.
Several approaches thus propose to operate in the primary sample space (PSS)~\cite{Kelemen:2002} represented by a unit hypercube $\PrimarySS$.
A path is then obtained by transforming a vector of random numbers $\PSSVector \in \PrimarySS$
using one of the standard path-construction techniques $\PathConstructionMap$ (e.g.\ camera tracing): $\Path = \PathConstructionMap(\PSSVector)$.

Operating in PSS has a number of compelling advantages.
The sampling routine has to be evaluated only once per path, instead of once per path vertex, and the generic nature of PSS coordinates enables treating the path construction as a black box.
Importance sampling of paths can thus be applied to any single path-tracing technique, and, with some effort, also to multiple strategies~\cite{LW1995A5TTRTVOMCRT,Veach:1994:BPT,Kelemen:2002,Hachisuka:2014:MMLT,Guo:2018,DBLP:journals/corr/abs-1808-07840}.
Lastly, the sampling routine directly benefits from existing importance-sampling techniques in the underlying path-tracing algorithm since those make the path-contribution function smoother in PSS and thus easier to learn.

\paragraph{Methodology}
Given that NICE scales well to high-dimensional problems, applying it in PSS is straightforward.
We split the dimensions of $\PrimarySS$ into two equally-sized groups $\partitionA$ and $\partitionB$, where $\partitionA$ contains the even dimensions and $\partitionB$ contains the odd dimensions. One group serves as the input of the neural network (each dimension is processed using the one-blob encoding) while the other group is being warped; their roles are swapped in the next coupling layer.
To infer the parameters $\Params$ of the networks, we minimize one of the losses from \autoref{sec:mc} against $\PdfGt(\Path) = \emittedRadiance(\pos_0, \pos_1) \PathThroughput(\Path) \Importance(\pos_{k-1}, \pos_k) \, F^{-1}$, ignoring the unknown normalization factor, i.e.\ assuming $F=1$.

In order to obtain a path sample $\Path$, we generate a random vector $\PSSVector$, warp it using the reversed inverted coupling layers, and apply the path-construction technique: $\Path = \rho\left( \map^{-1}_1\left( \cdots \map^{-1}_{\nlayers}(\PSSVector)\right) \right)$; please refer back to \autoref{eq:cmap-inverse-a} and~\eqref{eq:cmap-inverse-b} for details on the inverses.

Before we analyze the performance of primary-sample-space path sampling in \autoref{sec:results},
we discuss a slightly different approach to data-driven construction of path samples---the so-called path guiding---which applies neural importance sampling at each vertex of the path and typically yields higher performance.

\subsection{Path Guiding}%
\label{sec:imcremental-path-guiding}

A popular alternative to formalizing light transport using the path-integral formulation is to adopt a local view and focus on the radiative equilibrium of individual points in the scene.
The equilibrium radiance at a surface point $\pos$ in direction $\diro$ is given by the rendering equation~\citep{Kajiya:1986:TRE}:
\begin{align}
\outRadiance(\pos, \diro) &= \emittedRadiance(\pos, \diro) + \int_{\Omega} \inRadiance(\pos, \diri) \bsdf(\pos, \diro, \diri) |\cos \fsangle| \,\Diff\diri \,,
\label{eq:rendering-eq}
\end{align}
where $\bsdf$ is the bidirectional scattering distribution function, $\outRadiance(\pos,\diro)$, $\emittedRadiance(\pos,\diro)$, and $\inRadiance(\pos,\dir)$ are respectively the reflected, emitted, and incident radiance, ${\Omega}$ is the unit sphere, and $\fsangle$ is the angle between $\diri$ and the surface normal.

The rendering task is formulated as finding the outgoing radiance at points directly visible from the sensor.
The overall efficiency of the renderer heavily depends on the variance of estimating the amount of \emph{reflected} light:
\begin{align}
\langle \reflectedRadiance(\pos,\diro) \rangle = \frac{1}{N} \sum_{j=1}^N \frac{\inRadiance(\pos, \diri_j) \bsdf(\pos, \diro, \diri_j) |\cos \fsangle_j| }
{\PdfMC(\diri_j|\pos,\diro)} \,.
\label{eq:reflection-mc-estimator}
\end{align}
A large body of research has therefore focused on devising sampling densities $\PdfMC(\diri|\pos,\diro)$ that yield low variance.
While the density is defined over a 2D space, it is conditioned on position $\pos$ and direction $\diro$. These extra five dimensions make the goal of $\PdfMC(\diri|\pos,\diro) \propto \inRadiance(\pos, \diri) \bsdf(\pos, \diro, \diri) |\cos \fsangle|$ substantially harder.

Since the 7D domain is fairly challenging to handle using hand-crafted, spatio-directional data structures in the general case, most research has focused on the simpler 5D setting where $\PdfMC(\diri|\pos,\diro) \propto \inRadiance(\pos, \diri)$~\cite{Jensen1995,Hey:2002:ISH:584458.584476,PegoraroEGSR08SMCALAPM,PegoraroIRT08TIGIESMCA,Vorba:2014:OnlineLearningPMMinLTS,Dahm16,mueller2017practical} and only a few attempts have been made to consider the full product~\cite{LW1995A5TTRTVOMCRT,Steinhurst:06:GlossySampling,herholz2016,Herholz:2018}. These \emph{path-guiding} approaches rely on carefully chosen data structures (e.g.\ BVHs, kD-trees) in combination with relatively simple PDF models (e.g.\ histograms, quad-trees, Gaussian mixture models), which are populated in a data-driven manner either in a pre-pass or online during rendering.
Like in previous works our goal is to learn local sampling densities, but we differ in that we utilize NICE to represent and sample from $\PdfMC(\diri|\pos,\diro)$.

\paragraph{Methodology.}
We use a single instance of NICE, which is trained and sampled from in an interleaved manner: \ADD{drawn samples are immediately used for training, and training results are immediately used for further sampling.}
In the most general setting, we consider learning $\PdfMC(\diri\vert\pos,\diro)$ that is proportional to the product of \emph{all} terms in the integrand.
Since the integration domain is only 2D, partitions $\partitionA$ and $\partitionB$ in all coupling layers contain only one dimension each---one of the two cylindrical coordinates that we use to parameterize the sphere of direction.

To produce the parameters of the first piecewise-polynomial coupling function, the neural network $\nnet$ takes the cylindrical coordinate from $\partitionA$, the position $\pos$ and direction $\diro$ that condition the density, and additional information that may improve inference; we also input the normal of the intersected shape at $\pos$ to aid the network in learning distributions that correlate with the local shading frame.

We one-blob-encode all of the inputs as described in~\autoref{sec:one-blob-encoding} with $k = 32$.
In the case of $\pos$, we normalize it by the scene bounding box,
encode each coordinate independently, and concatenate the results into a single array of $3\times k$ values.
We proceed analogously with directions, which we parameterize using world-space cylindrical coordinates: we transform each coordinate to $[0,1]$ interval, encode it, and append to the array.
The improved performance enabled by our proposed one-blob encoding is reported in~\autoref{tab:one-blob-encoding}.

At any given point during rendering, a sample is generated by drawing a random pair $u\in {[0,1]}^2$, passing it through the inverted coupling layers in reverse order,
${\map^{-1}_1( \cdots \map^{-1}_{\nlayers}(u))}$, and transforming to the range of cylindrical coordinates to obtain $\diri$.

\paragraph{MIS-Aware Optimization}
\ADDTWO{In order to optimize the networks, we use Adam with one of the loss functions from \autoref{sec:mc}, but with an important, problem-specific alteration.
To sample $\diri$, most current renderers apply multiple importance sampling (MIS)~\citep{Veach:1995:MIS} to combine multiple sampling densities, \ADD{each tailored to a specific component of the integrand (direct illumination, BSDF, etc.).}
When learning the product, we take this into account by optimizing the networks with respect to the final \emph{effective} PDF $\PdfMC^\prime$ instead of the density learned using NICE.\@
If certain parts of the product are already covered well by existing densities, the networks will be optimized to handle only the remaining problematic case.}

\ADDTWO{We minimize $D(\PdfGt \, \| \, \PdfMC^\prime)$, where $D$ is either $\KlDiv$ or $\ChiDiv$ divergence,
and we employ the balance heuristic~\cite{Veach:1995:MIS} for combining the learned distribution $\PdfMC$ and the BSDF distribution $\PdfBSDF$. This yields the following effective PDF ${\PdfMC^\prime = \selectProb \PdfMC + (1-\selectProb) \PdfBSDF}$,
where $\selectProb$ is the probability of drawing samples from $\PdfMC$.
The ideal PDF $\PdfGt(\diri|\pos,\diro) = \inRadiance(\pos, \diri) \bsdf(\pos, \diro, \diri) |\cos \fsangle| F^{-1}$ is evaluated ignoring the normalization constant $F$ (as discussed in \autoref{sec:optimizing-kl}).\@}

\paragraph{Learned Selection Probabilities}
To further reduce variance we use an additional network $\hat{m}$ that learns approximately optimal selection probability $\selectProb = l\big(\hat{m}(\pos,\diro)\big)$, where $l$ is the logistic function. We optimize $\hat{m}$ jointly with the other networks; all use the same architecture except for the last layer.
\ADD{To prevent overfitting to local optima with degenerate selection probability, we use loss funtion $\beta(\tau) \, D(\PdfGt \, \| \, \PdfMC) + \big(1 - \beta(\tau)\big) \, D(\PdfGt \, \| \, \PdfMC^\prime)$ where $\tau \in [0, 1]$ is the fraction of exhausted render budget (either time or sample count) and $\beta(\tau) = \nicefrac{1}{2} \cdot {\left(\nicefrac{1}{3}\right)}^{5\tau}$.}

Since we employ the balance heuristic, which is provably optimal in the context of the MIS one-sample model~\cite{Veach:1997:Thesis}, learning the selection probabilities is the only means to further improve the performance, for instance by shutting down MIS completely in situations when it may hurt (e.g.\ on near-specular surfaces).

It is worth noting, however, that learning also the MIS weights (instead of relying on the balance heuristic) would remove the need to evaluate all relevant distributions for each sample.
This would remove one of the requirements that we stated in \autoref{sec:related-work}, namely the need for tractable inverses, thereby extending the set of admissible generative models; we leave this branch of investigations as future work.

\paragraph{BSDFs with Delta Components} BSDFs that are a mixture of Dirac-delta and smooth functions---such as \ADD{smooth} plastic---require special handling.
While our stochastic gradient in \autoref{sec:mc} is, in theory, well behaved with delta functions, they need to be treated as finite quantities in practice due to the limitations of floating-point numbers.
When the path tracer samples delta components, continuous densities need to be set to $0$ and optimization of our coupling functions disabled (by setting their loss to $0$), effectively only optimizing for selection probabilities.

\paragraph{Discussion}
Our approach to sampling the full product of the rendering equation $\inRadiance(\pos, \diri_j) \bsdf(\pos, \diro, \diri_j) |\cos \fsangle_j|$ has three distinct advantages.
First, it is agnostic to the number of dimensions that the 2D domain is conditioned on.
This allows for high-dimensional conditionals without sophisticated data structures.
One can simply input extra information into the neural networks and let them learn which dimensions are useful in which situations.
While we only pass in the surface normal, the networks could be supplied with additional information---e.g.\ textured BSDF parameters---to further improve the performance in cases where the product correlates with such information.
In that sense, our approach is more automatic than previous works.

The second advantage is that our method does not require any precomputation, such as fitting of (scene-dependent) materials into a mixture of Gaussians~\citep{herholz2016, Herholz:2018}.
While a user still needs to specify the hyperparameters as is also required by most other approaches, we found our configuration of hyperparameters to work well across all tested scenes.
\ADD{Note, however, that the lack of explicit factorization in our approach can be detrimental in situations where the individual factors are simpler to learn than their product, and the product can be easily importance sampled.}

Lastly, our approach offers trivial persistence across renders.
A set of networks trained on one camera view can be reused from a different view or within a slightly modified scene; see \autoref{sec:results}.
Unlike previous approaches, where the learned data structure requires explicit support of adaptation to new scenes, neural networks can be adapted by the same optimization procedure that was used in initial training.
\ADD{Applying our approach to animations could thus yield sub-linear training cost by amortizing it over multiple frames.}

\begin{figure*}
  \setlength{\fboxrule}{10pt}%
\setlength{\insetvsep}{20pt}%
\setlength{\tabcolsep}{-1pt}%
\renewcommand{\arraystretch}{1}%
\small%
\hspace*{-2mm}%
\begin{tabular}{ccccccccc}
  & & & & Affine, $\nlayers=16$ & \multicolumn{3}{c}{Ours, piecewise-quadratic, $\nlayers=4$} \\
  \cmidrule(lr){6-8}
  & & PT-Unidir & PSSPS---4D & NPS---4D & NPS---2D & NPS---4D & NPS---6D & Reference \\
  \setInset{A}{red}{55}{140}{51}{38}%
  \setInset{B}{orange}{440}{120}{51}{38}%
  \rotatebox{90}{\hspace{-1.75cm}\Bathroom{}}\hspace{0.14cm} & 
  \addBeautyCrop{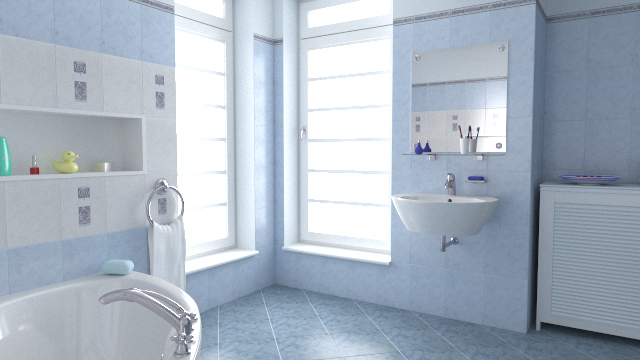}{0.27}{640}{360}{0}{0}{640}{360} &
  \addInsets{images/fig-pss-guiding/glossy-bathroom-PT-unidir.jpg} & 
  \addInsets{images/fig-pss-guiding/glossy-bathroom-PSSPS.jpg} & 
  \addInsets{images/fig-pss-guiding/glossy-bathroom-Affine-4d.jpg} & 
  \addInsets{images/fig-pss-guiding/glossy-bathroom-Ours-2d.jpg} & 
  \addInsets{images/fig-pss-guiding/glossy-bathroom-Ours-4d.jpg} & 
  \addInsets{images/fig-pss-guiding/glossy-bathroom-Ours-6d.jpg} & 
  \addInsets{images/references/glossy-bathroom-Reference.jpg} \\
  & \multicolumn{1}{r}{MAPE:} & 0.1467 & \textbf{0.1426} & 0.1529 & 0.1474 & 0.1450 & 0.1460 \\[2pt]
  \setInset{A}{red}{185}{76}{51}{38}%
  \setInset{B}{orange}{140}{230}{51}{38}%
  \rotatebox{90}{\hspace{-1.70cm}\Spaceship{}}\hspace{0.14cm} & 
  \addBeautyCrop{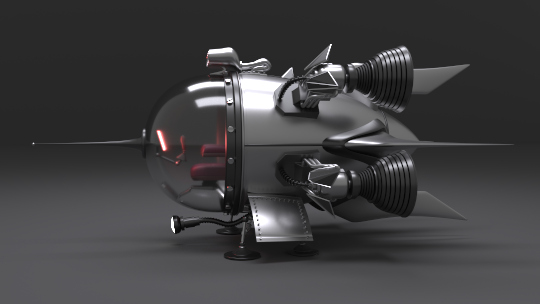}{0.27}{540}{304}{0}{0}{540}{304} &
  \addInsets{images/fig-pss-guiding/spaceship-PT-unidir.jpg} & 
  \addInsets{images/fig-pss-guiding/spaceship-PSSPS.jpg} & 
  \addInsets{images/fig-pss-guiding/spaceship-Affine-4d.jpg} & 
  \addInsets{images/fig-pss-guiding/spaceship-Ours-2d.jpg} & 
  \addInsets{images/fig-pss-guiding/spaceship-Ours-4d.jpg} & 
  \addInsets{images/fig-pss-guiding/spaceship-Ours-6d.jpg} & 
  \addInsets{images/references/spaceship-Reference.jpg} \\
  & \multicolumn{1}{r}{MAPE:} & 0.0422 & 0.0331 & 0.0343 & \textbf{0.0259} & 0.0259 & 0.0261 \\[2pt]
  \setInset{A}{red}{450}{110}{51}{38}%
  \setInset{B}{orange}{160}{245}{51}{38}%
  \rotatebox{90}{\hspace{-2.20cm}\CopperHairball{}}\hspace{0.14cm} & 
  \addBeautyCrop{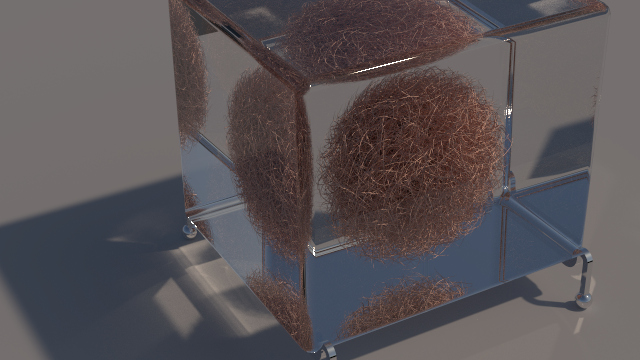}{0.27}{640}{360}{0}{0}{640}{360} &
  \addInsets{images/fig-pss-guiding/hairball-PT-unidir.jpg} & 
  \addInsets{images/fig-pss-guiding/hairball-PSSPS.jpg} & 
  \addInsets{images/fig-pss-guiding/hairball-Affine-4d.jpg} & 
  \addInsets{images/fig-pss-guiding/hairball-Ours-2d.jpg} & 
  \addInsets{images/fig-pss-guiding/hairball-Ours-4d.jpg} & 
  \addInsets{images/fig-pss-guiding/hairball-Ours-6d.jpg} & 
  \addInsets{images/references/hairball-Reference.jpg} \\
  & \multicolumn{1}{r}{MAPE:} & 0.4685 & 0.3139 & 0.2400 & 0.2236 & 0.1987 & \textbf{0.1956} \\[2pt]
  \setInset{A}{red}{540}{230}{51}{38}%
  \setInset{B}{orange}{430}{320}{51}{38}%
  \rotatebox{90}{\hspace{-2.20cm}\CountryKitchen{}}\hspace{0.14cm} & 
  \addBeautyCrop{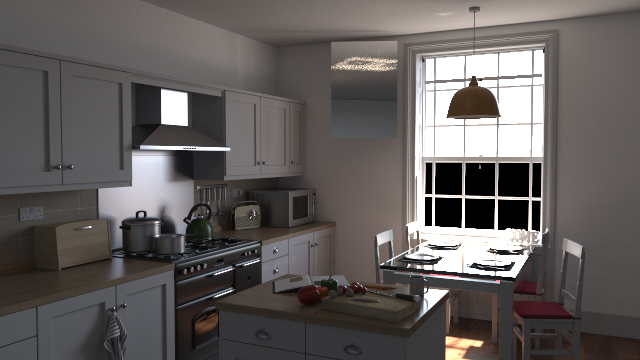}{0.27}{640}{360}{0}{0}{640}{360} &
  \addInsets{images/fig-pss-guiding/kitchen-PT-unidir.jpg} & 
  \addInsets{images/fig-pss-guiding/kitchen-PSSPS.jpg} & 
  \addInsets{images/fig-pss-guiding/kitchen-Affine-4d.jpg} & 
  \addInsets{images/fig-pss-guiding/kitchen-Ours-2d.jpg} & 
  \addInsets{images/fig-pss-guiding/kitchen-Ours-4d.jpg} & 
  \addInsets{images/fig-pss-guiding/kitchen-Ours-6d.jpg} & 
  \addInsets{images/references/kitchen-Reference.jpg} \\
  & \multicolumn{1}{r}{MAPE:} & 0.6879 & 0.6476 & 0.6607 & 0.6598 & 0.6033 & \textbf{0.5633} \\[2pt]
  \setInset{A}{red}{55}{220}{51}{38}%
  \setInset{B}{orange}{430}{90}{51}{38}%
  \rotatebox{90}{\hspace{-1.75cm}\Bookshelf{}}\hspace{0.14cm} & 
  \addBeautyCrop{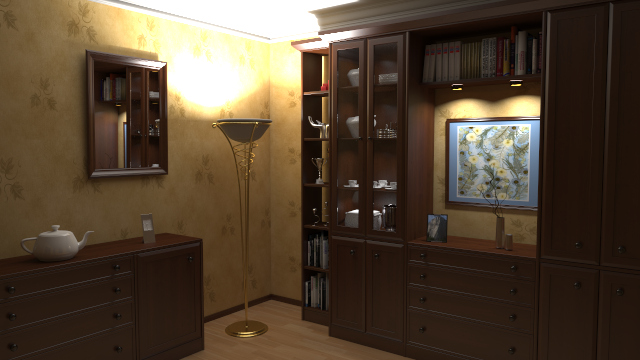}{0.27}{640}{360}{0}{0}{640}{360} &
  \addInsets{images/fig-pss-guiding/bookshelf-PT-unidir.jpg} & 
  \addInsets{images/fig-pss-guiding/bookshelf-PSSPS.jpg} & 
  \addInsets{images/fig-pss-guiding/bookshelf-Affine-4d.jpg} & 
  \addInsets{images/fig-pss-guiding/bookshelf-Ours-2d.jpg} & 
  \addInsets{images/fig-pss-guiding/bookshelf-Ours-4d.jpg} & 
  \addInsets{images/fig-pss-guiding/bookshelf-Ours-6d.jpg} & 
  \addInsets{images/references/bookshelf-Reference.jpg} \\
  & \multicolumn{1}{r}{MAPE:} & 0.7955 & 0.7553 & 0.6712 & 0.6915 & \textbf{0.6702} & 0.6810 \\[2pt]
  \setInset{A}{red}{280}{110}{51}{38}%
  \setInset{B}{orange}{480}{150}{51}{38}%
  \rotatebox{90}{\hspace{-1.65cm}\Necklace{}}\hspace{0.14cm} & 
  \addBeautyCrop{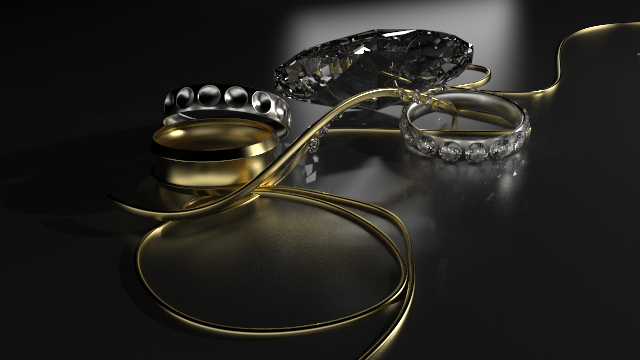}{0.27}{640}{360}{0}{0}{640}{360} &
  \addInsets{images/fig-pss-guiding/necklace-PT-unidir.jpg} & 
  \addInsets{images/fig-pss-guiding/necklace-PSSPS.jpg} & 
  \addInsets{images/fig-pss-guiding/necklace-Affine-4d.jpg} & 
  \addInsets{images/fig-pss-guiding/necklace-Ours-2d.jpg} & 
  \addInsets{images/fig-pss-guiding/necklace-Ours-4d.jpg} & 
  \addInsets{images/fig-pss-guiding/necklace-Ours-6d.jpg} & 
  \addInsets{images/references/necklace-Reference.jpg} \\
  & \multicolumn{1}{r}{MAPE:} & 0.3436 & 0.3334 & 0.3349 & \textbf{0.2210} & 0.2594 & 0.2680 \\[-5pt]
\end{tabular}
\unsetInset{A}%
\unsetInset{B}%

  \caption{
    Neural path sampling in primary sample space.
    We compare a standard uni-directional path tracer (PT-Unidir), the method by \citet{Guo:2018} (PSSPS), neural path sampling using $\nlayers=16$ multiply-add coupling layers~\cite{dinh2016density}, and $\nlayers=4$ of our proposed piecewise-quadratic coupling layers, both optimized using the KL divergence.
    We experimented with sampling the 1, 2, or 3 first non-specular bounces (NPS--2D, NPS--4D and NPS--6D).
    Overall, our technique performs best in terms of \emph{mean absolute percentage error} (MAPE) in this experiment, but only offers improvement beyond the 4D case if paths stay coherent, e.g.\ in the top crop of the \Spaceship{} scene.
    More results and error visualizations can be found in the supplemented image viewer.
  }%
  \label{fig:pss_pg}
\end{figure*}

\begin{figure*}
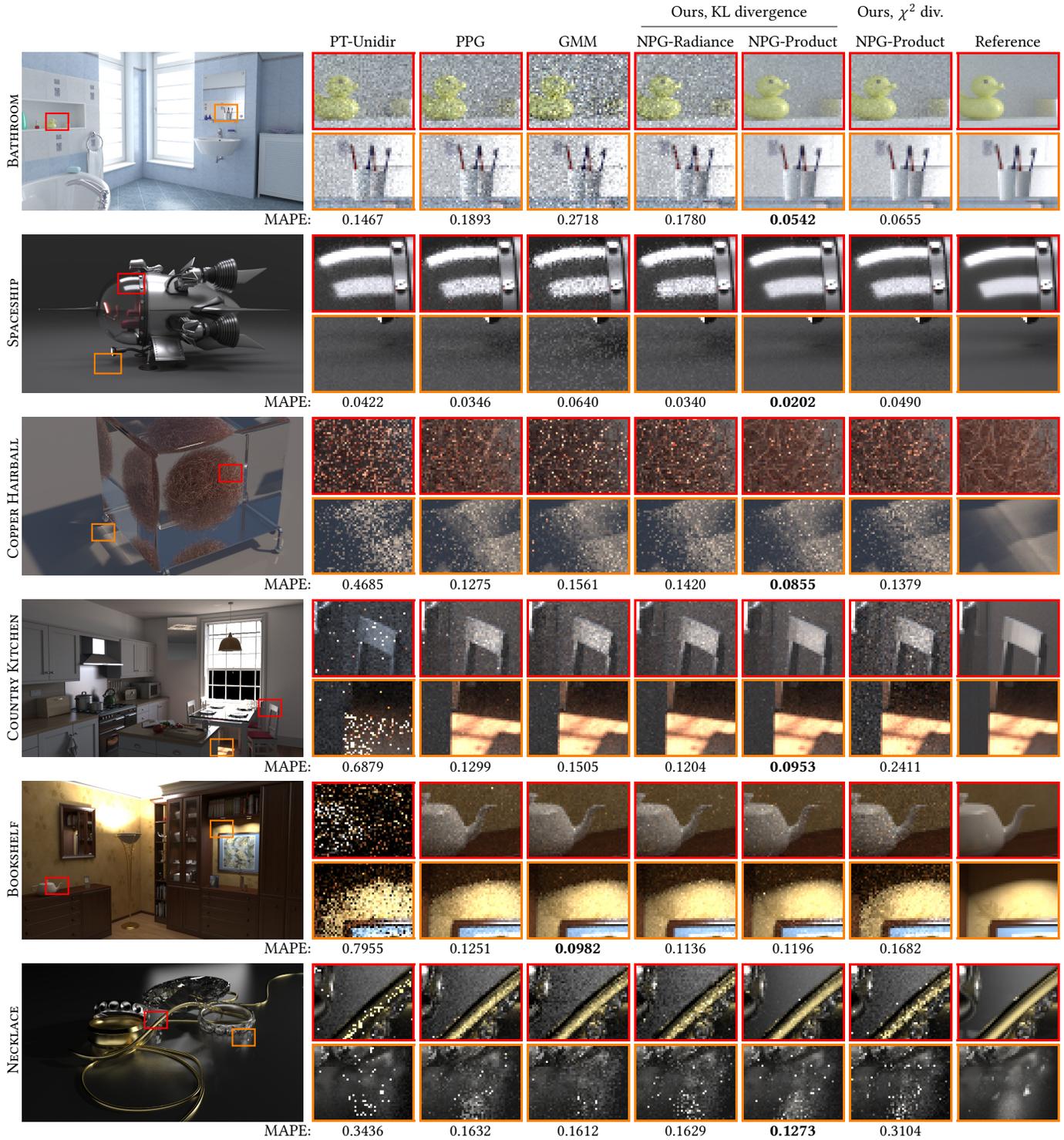

  \setlength{\fboxrule}{10pt}%
\setlength{\insetvsep}{20pt}%
\setlength{\tabcolsep}{-1pt}%
\renewcommand{\arraystretch}{1}%
\small%
\hspace*{-2mm}%
\begin{tabular}{ccccccccc}
  & & & & & \multicolumn{2}{c}{Ours, KL divergence} & Ours, $\chi^2$ div. \\
  \cmidrule(lr){6-7}
  & & PT-Unidir & PPG & GMM & NPG-Radiance & NPG-Product & NPG-Product & Reference \\
    \setInset{A}{red}{55}{140}{51}{38}%
    \setInset{B}{orange}{440}{120}{51}{38}%
    \rotatebox{90}{\hspace{-1.75cm}\Bathroom{}}\hspace{0.14cm} & 
    \addBeautyCrop{images/references/glossy-bathroom-Reference.jpg}{0.27}{640}{360}{0}{0}{640}{360} &
    \addInsets{images/fig-path-guiding/glossy-bathroom-PT-unidir.jpg} &
    \addInsets{images/fig-path-guiding/glossy-bathroom-PPG.jpg} &
    \addInsets{images/fig-path-guiding/glossy-bathroom-Vorba.jpg} &
    \addInsets{images/fig-path-guiding/glossy-bathroom-Ours-IRS.jpg} &
    \addInsets{images/fig-path-guiding/glossy-bathroom-Ours-FPS.jpg} &
    \addInsets{images/fig-path-guiding/glossy-bathroom-Ours-FPS-chi.jpg} &
    \addInsets{images/references/glossy-bathroom-Reference.jpg} \\
  & \multicolumn{1}{r}{MAPE:} & 0.1467 & 0.1893 & 0.2718 & 0.1780 & \textbf{0.0542} & 0.0655 \\[2pt]
    \setInset{A}{red}{185}{76}{51}{38}%
    \setInset{B}{orange}{140}{230}{51}{38}%
    \rotatebox{90}{\hspace{-1.70cm}\Spaceship{}}\hspace{0.14cm} & 
    \addBeautyCrop{images/references/spaceship-Reference.jpg}{0.27}{540}{304}{0}{0}{540}{304} &
    \addInsets{images/fig-path-guiding/spaceship-PT-unidir.jpg} &
    \addInsets{images/fig-path-guiding/spaceship-PPG.jpg} &
    \addInsets{images/fig-path-guiding/spaceship-Vorba.jpg} &
    \addInsets{images/fig-path-guiding/spaceship-Ours-IRS.jpg} &
    \addInsets{images/fig-path-guiding/spaceship-Ours-FPS.jpg} &
    \addInsets{images/fig-path-guiding/spaceship-Ours-FPS-chi.jpg} &
    \addInsets{images/references/spaceship-Reference.jpg} \\
  & \multicolumn{1}{r}{MAPE:} & 0.0422 & 0.0346 & 0.0640 & 0.0340 & \textbf{0.0202} & 0.0490 \\[2pt]
    \setInset{A}{red}{450}{110}{51}{38}%
    \setInset{B}{orange}{160}{245}{51}{38}%
    \rotatebox{90}{\hspace{-2.20cm}\CopperHairball{}}\hspace{0.14cm} & 
    \addBeautyCrop{images/references/hairball-Reference.jpg}{0.27}{640}{360}{0}{0}{640}{360} &
    \addInsets{images/fig-path-guiding/hairball-PT-unidir.jpg} &
    \addInsets{images/fig-path-guiding/hairball-PPG.jpg} &
    \addInsets{images/fig-path-guiding/hairball-Vorba.jpg} &
    \addInsets{images/fig-path-guiding/hairball-Ours-IRS.jpg} &
    \addInsets{images/fig-path-guiding/hairball-Ours-FPS.jpg} &
    \addInsets{images/fig-path-guiding/hairball-Ours-FPS-chi.jpg} &
    \addInsets{images/references/hairball-Reference.jpg} \\
  & \multicolumn{1}{r}{MAPE:} & 0.4685 & 0.1275 & 0.1561 & 0.1420 & \textbf{0.0855} & 0.1379 \\[2pt]
    \setInset{A}{red}{540}{230}{51}{38}%
    \setInset{B}{orange}{430}{320}{51}{38}%
    \rotatebox{90}{\hspace{-2.20cm}\CountryKitchen{}}\hspace{0.14cm} & 
    \addBeautyCrop{images/references/kitchen-Reference.jpg}{0.27}{640}{360}{0}{0}{640}{360} &
    \addInsets{images/fig-path-guiding/kitchen-PT-unidir.jpg} &
    \addInsets{images/fig-path-guiding/kitchen-PPG.jpg} &
    \addInsets{images/fig-path-guiding/kitchen-Vorba.jpg} &
    \addInsets{images/fig-path-guiding/kitchen-Ours-IRS.jpg} &
    \addInsets{images/fig-path-guiding/kitchen-Ours-FPS.jpg} &
    \addInsets{images/fig-path-guiding/kitchen-Ours-FPS-chi.jpg} &
    \addInsets{images/references/kitchen-Reference.jpg} \\
  & \multicolumn{1}{r}{MAPE:} & 0.6879 & 0.1299 & 0.1505 & 0.1204 & \textbf{0.0953} & 0.2411 \\[2pt]
    \setInset{A}{red}{55}{220}{51}{38}%
    \setInset{B}{orange}{430}{90}{51}{38}%
    \rotatebox{90}{\hspace{-1.75cm}\Bookshelf{}}\hspace{0.14cm} & 
    \addBeautyCrop{images/references/bookshelf-Reference.jpg}{0.27}{640}{360}{0}{0}{640}{360} &
    \addInsets{images/fig-path-guiding/bookshelf-PT-unidir.jpg} &
    \addInsets{images/fig-path-guiding/bookshelf-PPG.jpg} &
    \addInsets{images/fig-path-guiding/bookshelf-Vorba.jpg} &
    \addInsets{images/fig-path-guiding/bookshelf-Ours-IRS.jpg} &
    \addInsets{images/fig-path-guiding/bookshelf-Ours-FPS.jpg} &
    \addInsets{images/fig-path-guiding/bookshelf-Ours-FPS-chi.jpg} &
    \addInsets{images/references/bookshelf-Reference.jpg} \\
  & \multicolumn{1}{r}{MAPE:} & 0.7955 & 0.1251 & \textbf{0.0982} & 0.1136 & 0.1196 & 0.1682 \\[2pt]
    \setInset{A}{red}{280}{110}{51}{38}%
    \setInset{B}{orange}{480}{150}{51}{38}%
    \rotatebox{90}{\hspace{-1.65cm}\Necklace{}}\hspace{0.14cm} & 
    \addBeautyCrop{images/references/necklace-Reference.jpg}{0.27}{640}{360}{0}{0}{640}{360} &
    \addInsets{images/fig-path-guiding/necklace-PT-unidir.jpg} &
    \addInsets{images/fig-path-guiding/necklace-PPG.jpg} &
    \addInsets{images/fig-path-guiding/necklace-Vorba.jpg} &
    \addInsets{images/fig-path-guiding/necklace-Ours-IRS.jpg} &
    \addInsets{images/fig-path-guiding/necklace-Ours-FPS.jpg} &
    \addInsets{images/fig-path-guiding/necklace-Ours-FPS-chi.jpg} &
    \addInsets{images/references/necklace-Reference.jpg} \\
  & \multicolumn{1}{r}{MAPE:} & 0.3436 & 0.1632 & 0.1612 & 0.1629 & \textbf{0.1273} & 0.3104 \\[-5pt]
\end{tabular}
\unsetInset{A}
\unsetInset{B}
  \caption{
    Neural path guiding.
    We compare a uni-directional path tracer (PT-Unidir), the practical path-guiding (PPG) algorithm of~\citet{mueller2017practical}, the Gaussian mixture model (GMM) of \citet{Vorba:2014:OnlineLearningPMMinLTS}, and variants of our framework with $L=4$ coupling layers sampling the incident radiance alone~(NPG-Radiance) or the whole integrand (NPG-Product), when optimizing either the KL and $\chi^2$ divergences.
    Overall, sampling the whole integrand with the KL divergence yields the most robust results.
    More results and error visualizations can be found in the supplemented image viewer.
  }%
  \label{fig:path_guiding}
\end{figure*}

\begin{table*}[t!]
  \caption{
    Mean average percentage error (MAPE) and render times of various importance-sampling approaches.
    At equal sampling rates---we report the number of samples in each scene as mega samples (MS)---our technique performs on par or better than the practical path guiding (PPG) algorithm of~\citet{mueller2017practical} and the bidirectionally trained Gaussian mixture model (GMM) of \citet{Vorba:2014:OnlineLearningPMMinLTS} in all scenes but the \Bookshelf{}, but incurs a large computational overhead.
    Since the GMMs are trained in a pre-pass, we report both their training and rendering times.
    Please note, that the provided implementation of the GMM training does not scale well beyond $8$ threads.
    Furthermore, we do not report GMM results for the \GlossyKitchen{} and \VeachDoor{} scenes due to crashes and bias, respectively.
    Our neural path sampling (NPS) likewise compares favorably against the method by \citet{Guo:2018} (PSSPS).
    Using one-blob encoding significantly improves the quality of our results; see~\autoref{fig:metrics-bar-chart} for a histogram visualization of these metrics.
    We also evaluated SMAPE, L1, MRSE, L2, and SSIM, which all can be inspected as false-color maps and aggregates in the supplemented image viewer.
  }%
  \vspace{-3mm}%
  \label{tab:one-blob-encoding}%
  \small
\rowcolors{2}{white}{gray!8}
\setlength{\tabcolsep}{6.33pt}
\begin{tabular*}{\textwidth}{rrccccccccc}
  \toprule
  {} & {} & {} & {} & {} & {} & \multicolumn{4}{c}{Ours, KL divergence} & \makebox[0.45cm]{Ours, $\chi^2$ div.} \\
  \cmidrule(lr){7-10}
  {} & {} & \makebox[0.40cm]{PT-Unidir} & PPG & GMM & PSSPS & NPS & \makebox[0.45cm]{NPG-Rad.} & \multicolumn{2}{c}{NPG-Product} & \makebox[0.45cm]{NPG-Product} \\
  \cmidrule(lr){9-10}
  \rowcolor{white}
  {} & {} & {} & {} & {} & {} & one-blob & one-blob & scalar & one-blob & one-blob \\
  \rowcolor{white}
  {} & {} & {} & {} & \cellcolor[rgb]{0.92,0.95,0.82}{} & \cellcolor[rgb]{0.89,0.93,0.73}{} & \cellcolor[rgb]{0.72, 0.86, 0.73}{} & \cellcolor[rgb]{0.44, 0.72, 0.71}{} & \cellcolor[rgb]{0.25, 0.59, 0.69}{} & \cellcolor[rgb]{0.20, 0.42, 0.62}{} & \cellcolor[rgb]{0.2, 0.26, 0.53}{}\\
  \midrule
  \small{\Bathroom{}} & \small{\makebox[0.7cm][r]{\textcolor{gray}{236 MS}}}\hspace{-1.7mm} & \small{\makebox[0.46cm]{{0.147}}\,\,\,\,\,\,\,\makebox[0.27cm]{\textcolor{gray}{88s}}} & \small{\makebox[0.46cm]{{0.189}}\,\,\,\,\,\,\,\makebox[0.27cm]{\textcolor{gray}{2.3m}}} & \small{\makebox[0.46cm]{{0.272}}\,\,\,\,\,\,\makebox[0.40cm]{\textcolor{gray}{9.1m}} \,\textcolor{gray}{+}\,\, \makebox[0.27cm]{\textcolor{gray}{48s}}} & \small{\makebox[0.46cm]{{0.143}}\,\,\,\,\,\,\,\makebox[0.27cm]{\textcolor{gray}{89s}}} & \small{\makebox[0.46cm]{{0.146}}\,\,\,\,\,\,\,\makebox[0.27cm]{\textcolor{gray}{3.5m}}} & \small{\makebox[0.46cm]{{0.178}}\,\,\,\,\,\,\,\makebox[0.27cm]{\textcolor{gray}{9.3m}}} & \small{\makebox[0.46cm]{{0.071}}\,\,\,\,\,\,\,\makebox[0.27cm]{\textcolor{gray}{11m}}} & \small{\makebox[0.46cm]{\textbf{{0.054}}}\,\,\,\,\,\,\,\makebox[0.27cm]{\textcolor{gray}{12m}}} & \small{\makebox[0.46cm]{{0.066}}\,\,\,\,\,\,\,\makebox[0.27cm]{\textcolor{gray}{15m}}} \\
  \small{\Bedroom{}} & \small{\makebox[0.7cm][r]{\textcolor{gray}{236 MS}}}\hspace{-1.7mm} & \small{\makebox[0.46cm]{{0.078}}\,\,\,\,\,\,\,\makebox[0.27cm]{\textcolor{gray}{75s}}} & \small{\makebox[0.46cm]{{0.053}}\,\,\,\,\,\,\,\makebox[0.27cm]{\textcolor{gray}{1.8m}}} & \small{\makebox[0.46cm]{{0.063}}\,\,\,\,\,\,\makebox[0.40cm]{\textcolor{gray}{34m}} \,\textcolor{gray}{+}\,\, \makebox[0.27cm]{\textcolor{gray}{60s}}} & \small{\makebox[0.46cm]{{0.068}}\,\,\,\,\,\,\,\makebox[0.27cm]{\textcolor{gray}{77s}}} & \small{\makebox[0.46cm]{{0.068}}\,\,\,\,\,\,\,\makebox[0.27cm]{\textcolor{gray}{3.5m}}} & \small{\makebox[0.46cm]{{0.045}}\,\,\,\,\,\,\,\makebox[0.27cm]{\textcolor{gray}{6.4m}}} & \small{\makebox[0.46cm]{{0.037}}\,\,\,\,\,\,\,\makebox[0.27cm]{\textcolor{gray}{7.2m}}} & \small{\makebox[0.46cm]{\textbf{{0.032}}}\,\,\,\,\,\,\,\makebox[0.27cm]{\textcolor{gray}{7.7m}}} & \small{\makebox[0.46cm]{{0.042}}\,\,\,\,\,\,\,\makebox[0.27cm]{\textcolor{gray}{10m}}} \\
  \small{\Bookshelf{}} & \small{\makebox[0.7cm][r]{\textcolor{gray}{236 MS}}}\hspace{-1.7mm} & \small{\makebox[0.46cm]{{0.796}}\,\,\,\,\,\,\,\makebox[0.27cm]{\textcolor{gray}{74s}}} & \small{\makebox[0.46cm]{{0.125}}\,\,\,\,\,\,\,\makebox[0.27cm]{\textcolor{gray}{2.5m}}} & \small{\makebox[0.46cm]{\textbf{{0.098}}}\,\,\,\,\,\,\makebox[0.40cm]{\textcolor{gray}{16m}} \,\textcolor{gray}{+}\,\, \makebox[0.27cm]{\textcolor{gray}{66s}}} & \small{\makebox[0.46cm]{{0.755}}\,\,\,\,\,\,\,\makebox[0.27cm]{\textcolor{gray}{78s}}} & \small{\makebox[0.46cm]{{0.681}}\,\,\,\,\,\,\,\makebox[0.27cm]{\textcolor{gray}{3.5m}}} & \small{\makebox[0.46cm]{{0.114}}\,\,\,\,\,\,\,\makebox[0.27cm]{\textcolor{gray}{8.1m}}} & \small{\makebox[0.46cm]{{0.250}}\,\,\,\,\,\,\,\makebox[0.27cm]{\textcolor{gray}{10m}}} & \small{\makebox[0.46cm]{{0.120}}\,\,\,\,\,\,\,\makebox[0.27cm]{\textcolor{gray}{10m}}} & \small{\makebox[0.46cm]{{0.168}}\,\,\,\,\,\,\,\makebox[0.27cm]{\textcolor{gray}{12m}}} \\
  \small{\CopperHairball} & \small{\makebox[0.7cm][r]{\textcolor{gray}{472 MS}}}\hspace{-1.7mm} & \small{\makebox[0.46cm]{{0.468}}\,\,\,\,\,\,\,\makebox[0.27cm]{\textcolor{gray}{2.0m}}} & \small{\makebox[0.46cm]{{0.128}}\,\,\,\,\,\,\,\makebox[0.27cm]{\textcolor{gray}{1.8m}}} & \small{\makebox[0.46cm]{{0.156}}\,\,\,\,\,\,\makebox[0.40cm]{\textcolor{gray}{11m}} \,\textcolor{gray}{+}\,\, \makebox[0.27cm]{\textcolor{gray}{2.0m}}} & \small{\makebox[0.46cm]{{0.314}}\,\,\,\,\,\,\,\makebox[0.27cm]{\textcolor{gray}{2.0m}}} & \small{\makebox[0.46cm]{{0.196}}\,\,\,\,\,\,\,\makebox[0.27cm]{\textcolor{gray}{6.9m}}} & \small{\makebox[0.46cm]{{0.142}}\,\,\,\,\,\,\,\makebox[0.27cm]{\textcolor{gray}{4.9m}}} & \small{\makebox[0.46cm]{{0.092}}\,\,\,\,\,\,\,\makebox[0.27cm]{\textcolor{gray}{15m}}} & \small{\makebox[0.46cm]{\textbf{{0.086}}}\,\,\,\,\,\,\,\makebox[0.27cm]{\textcolor{gray}{16m}}} & \small{\makebox[0.46cm]{{0.138}}\,\,\,\,\,\,\,\makebox[0.27cm]{\textcolor{gray}{17m}}} \\
  \small{\CornellBox{}} & \small{\makebox[0.7cm][r]{\textcolor{gray}{268 MS}}}\hspace{-1.7mm} & \small{\makebox[0.46cm]{{0.185}}\,\,\,\,\,\,\,\makebox[0.27cm]{\textcolor{gray}{23s}}} & \small{\makebox[0.46cm]{{0.044}}\,\,\,\,\,\,\,\makebox[0.27cm]{\textcolor{gray}{82s}}} & \small{\makebox[0.46cm]{{0.049}}\,\,\,\,\,\,\makebox[0.40cm]{\textcolor{gray}{6.2m}} \,\textcolor{gray}{+}\,\, \makebox[0.27cm]{\textcolor{gray}{24s}}} & \small{\makebox[0.46cm]{{0.130}}\,\,\,\,\,\,\,\makebox[0.27cm]{\textcolor{gray}{26s}}} & \small{\makebox[0.46cm]{{0.109}}\,\,\,\,\,\,\,\makebox[0.27cm]{\textcolor{gray}{4.0m}}} & \small{\makebox[0.46cm]{{0.035}}\,\,\,\,\,\,\,\makebox[0.27cm]{\textcolor{gray}{6.6m}}} & \small{\makebox[0.46cm]{{0.027}}\,\,\,\,\,\,\,\makebox[0.27cm]{\textcolor{gray}{8.5m}}} & \small{\makebox[0.46cm]{\textbf{{0.021}}}\,\,\,\,\,\,\,\makebox[0.27cm]{\textcolor{gray}{10m}}} & \small{\makebox[0.46cm]{{0.027}}\,\,\,\,\,\,\,\makebox[0.27cm]{\textcolor{gray}{9.1m}}} \\
  \small{\CountryKitchen{}} & \small{\makebox[0.7cm][r]{\textcolor{gray}{236 MS}}}\hspace{-1.7mm} & \small{\makebox[0.46cm]{{0.688}}\,\,\,\,\,\,\,\makebox[0.27cm]{\textcolor{gray}{48s}}} & \small{\makebox[0.46cm]{{0.130}}\,\,\,\,\,\,\,\makebox[0.27cm]{\textcolor{gray}{81s}}} & \small{\makebox[0.46cm]{{0.151}}\,\,\,\,\,\,\makebox[0.40cm]{\textcolor{gray}{13m}} \,\textcolor{gray}{+}\,\, \makebox[0.27cm]{\textcolor{gray}{33s}}} & \small{\makebox[0.46cm]{{0.648}}\,\,\,\,\,\,\,\makebox[0.27cm]{\textcolor{gray}{49s}}} & \small{\makebox[0.46cm]{{0.563}}\,\,\,\,\,\,\,\makebox[0.27cm]{\textcolor{gray}{3.5m}}} & \small{\makebox[0.46cm]{{0.120}}\,\,\,\,\,\,\,\makebox[0.27cm]{\textcolor{gray}{5.4m}}} & \small{\makebox[0.46cm]{{0.123}}\,\,\,\,\,\,\,\makebox[0.27cm]{\textcolor{gray}{6.9m}}} & \small{\makebox[0.46cm]{\textbf{{0.095}}}\,\,\,\,\,\,\,\makebox[0.27cm]{\textcolor{gray}{7.8m}}} & \small{\makebox[0.46cm]{{0.241}}\,\,\,\,\,\,\,\makebox[0.27cm]{\textcolor{gray}{8.1m}}} \\
  \small{\GlossyKitchen{}} & \small{\makebox[0.7cm][r]{\textcolor{gray}{236 MS}}}\hspace{-1.7mm} & \small{\makebox[0.46cm]{{1.476}}\,\,\,\,\,\,\,\makebox[0.27cm]{\textcolor{gray}{78s}}} & \small{\makebox[0.46cm]{{0.308}}\,\,\,\,\,\,\,\makebox[0.27cm]{\textcolor{gray}{87s}}} & \small{---} & \small{\makebox[0.46cm]{{1.452}}\,\,\,\,\,\,\,\makebox[0.27cm]{\textcolor{gray}{77s}}} & \small{\makebox[0.46cm]{{1.491}}\,\,\,\,\,\,\,\makebox[0.27cm]{\textcolor{gray}{3.5m}}} & \small{\makebox[0.46cm]{{0.243}}\,\,\,\,\,\,\,\makebox[0.27cm]{\textcolor{gray}{5.6m}}} & \small{\makebox[0.46cm]{{0.810}}\,\,\,\,\,\,\,\makebox[0.27cm]{\textcolor{gray}{11m}}} & \small{\makebox[0.46cm]{\textbf{{0.136}}}\,\,\,\,\,\,\,\makebox[0.27cm]{\textcolor{gray}{11m}}} & \small{\makebox[0.46cm]{{0.251}}\,\,\,\,\,\,\,\makebox[0.27cm]{\textcolor{gray}{13m}}} \\
  \small{\Necklace{}} & \small{\makebox[0.7cm][r]{\textcolor{gray}{236 MS}}}\hspace{-1.7mm} & \small{\makebox[0.46cm]{{0.344}}\,\,\,\,\,\,\,\makebox[0.27cm]{\textcolor{gray}{29s}}} & \small{\makebox[0.46cm]{{0.163}}\,\,\,\,\,\,\,\makebox[0.27cm]{\textcolor{gray}{42s}}} & \small{\makebox[0.46cm]{{0.161}}\,\,\,\,\,\,\makebox[0.40cm]{\textcolor{gray}{3.2m}} \,\textcolor{gray}{+}\,\, \makebox[0.27cm]{\textcolor{gray}{15s}}} & \small{\makebox[0.46cm]{{0.333}}\,\,\,\,\,\,\,\makebox[0.27cm]{\textcolor{gray}{31s}}} & \small{\makebox[0.46cm]{{0.268}}\,\,\,\,\,\,\,\makebox[0.27cm]{\textcolor{gray}{3.5m}}} & \small{\makebox[0.46cm]{{0.163}}\,\,\,\,\,\,\,\makebox[0.27cm]{\textcolor{gray}{2.6m}}} & \small{\makebox[0.46cm]{{0.249}}\,\,\,\,\,\,\,\makebox[0.27cm]{\textcolor{gray}{10m}}} & \small{\makebox[0.46cm]{\textbf{{0.127}}}\,\,\,\,\,\,\,\makebox[0.27cm]{\textcolor{gray}{11m}}} & \small{\makebox[0.46cm]{{0.310}}\,\,\,\,\,\,\,\makebox[0.27cm]{\textcolor{gray}{10m}}} \\
  \small{\SalleDeBain{}} & \small{\makebox[0.7cm][r]{\textcolor{gray}{236 MS}}}\hspace{-1.7mm} & \small{\makebox[0.46cm]{{0.185}}\,\,\,\,\,\,\,\makebox[0.27cm]{\textcolor{gray}{51s}}} & \small{\makebox[0.46cm]{{0.071}}\,\,\,\,\,\,\,\makebox[0.27cm]{\textcolor{gray}{89s}}} & \small{\makebox[0.46cm]{{0.080}}\,\,\,\,\,\,\makebox[0.40cm]{\textcolor{gray}{11m}} \,\textcolor{gray}{+}\,\, \makebox[0.27cm]{\textcolor{gray}{37s}}} & \small{\makebox[0.46cm]{{0.161}}\,\,\,\,\,\,\,\makebox[0.27cm]{\textcolor{gray}{54s}}} & \small{\makebox[0.46cm]{{0.158}}\,\,\,\,\,\,\,\makebox[0.27cm]{\textcolor{gray}{3.5m}}} & \small{\makebox[0.46cm]{{0.057}}\,\,\,\,\,\,\,\makebox[0.27cm]{\textcolor{gray}{5.4m}}} & \small{\makebox[0.46cm]{{0.052}}\,\,\,\,\,\,\,\makebox[0.27cm]{\textcolor{gray}{5.7m}}} & \small{\makebox[0.46cm]{\textbf{{0.042}}}\,\,\,\,\,\,\,\makebox[0.27cm]{\textcolor{gray}{6.1m}}} & \small{\makebox[0.46cm]{{0.062}}\,\,\,\,\,\,\,\makebox[0.27cm]{\textcolor{gray}{7.9m}}} \\
  \small{\Spaceship{}} & \small{\makebox[0.7cm][r]{\textcolor{gray}{236 MS}}}\hspace{-1.7mm} & \small{\makebox[0.46cm]{{0.042}}\,\,\,\,\,\,\,\makebox[0.27cm]{\textcolor{gray}{27s}}} & \small{\makebox[0.46cm]{{0.035}}\,\,\,\,\,\,\,\makebox[0.27cm]{\textcolor{gray}{56s}}} & \small{\makebox[0.46cm]{{0.064}}\,\,\,\,\,\,\makebox[0.40cm]{\textcolor{gray}{4.4m}} \,\textcolor{gray}{+}\,\, \makebox[0.27cm]{\textcolor{gray}{4.6m}}} & \small{\makebox[0.46cm]{{0.033}}\,\,\,\,\,\,\,\makebox[0.27cm]{\textcolor{gray}{28s}}} & \small{\makebox[0.46cm]{{0.026}}\,\,\,\,\,\,\,\makebox[0.27cm]{\textcolor{gray}{3.5m}}} & \small{\makebox[0.46cm]{{0.034}}\,\,\,\,\,\,\,\makebox[0.27cm]{\textcolor{gray}{2.8m}}} & \small{\makebox[0.46cm]{{0.027}}\,\,\,\,\,\,\,\makebox[0.27cm]{\textcolor{gray}{3.5m}}} & \small{\makebox[0.46cm]{\textbf{{0.020}}}\,\,\,\,\,\,\,\makebox[0.27cm]{\textcolor{gray}{3.9m}}} & \small{\makebox[0.46cm]{{0.049}}\,\,\,\,\,\,\,\makebox[0.27cm]{\textcolor{gray}{4.1m}}} \\
  \small{\Sponza{}} & \small{\makebox[0.7cm][r]{\textcolor{gray}{236 MS}}}\hspace{-1.7mm} & \small{\makebox[0.46cm]{{1.709}}\,\,\,\,\,\,\,\makebox[0.27cm]{\textcolor{gray}{84s}}} & \small{\makebox[0.46cm]{{0.353}}\,\,\,\,\,\,\,\makebox[0.27cm]{\textcolor{gray}{91s}}} & \small{\makebox[0.46cm]{{0.115}}\,\,\,\,\,\,\makebox[0.40cm]{\textcolor{gray}{11m}} \,\textcolor{gray}{+}\,\, \makebox[0.27cm]{\textcolor{gray}{53s}}} & \small{\makebox[0.46cm]{{1.692}}\,\,\,\,\,\,\,\makebox[0.27cm]{\textcolor{gray}{81s}}} & \small{\makebox[0.46cm]{{1.616}}\,\,\,\,\,\,\,\makebox[0.27cm]{\textcolor{gray}{3.5m}}} & \small{\makebox[0.46cm]{\textbf{{0.109}}}\,\,\,\,\,\,\,\makebox[0.27cm]{\textcolor{gray}{7.4m}}} & \small{\makebox[0.46cm]{{0.213}}\,\,\,\,\,\,\,\makebox[0.27cm]{\textcolor{gray}{9.0m}}} & \small{\makebox[0.46cm]{{0.110}}\,\,\,\,\,\,\,\makebox[0.27cm]{\textcolor{gray}{11m}}} & \small{\makebox[0.46cm]{{0.247}}\,\,\,\,\,\,\,\makebox[0.27cm]{\textcolor{gray}{11m}}} \\
  \small{\WoodenStaircase{}} & \small{\makebox[0.7cm][r]{\textcolor{gray}{236 MS}}}\hspace{-1.7mm} & \small{\makebox[0.46cm]{{0.163}}\,\,\,\,\,\,\,\makebox[0.27cm]{\textcolor{gray}{57s}}} & \small{\makebox[0.46cm]{{0.057}}\,\,\,\,\,\,\,\makebox[0.27cm]{\textcolor{gray}{81s}}} & \small{\makebox[0.46cm]{{0.065}}\,\,\,\,\,\,\makebox[0.40cm]{\textcolor{gray}{14m}} \,\textcolor{gray}{+}\,\, \makebox[0.27cm]{\textcolor{gray}{41s}}} & \small{\makebox[0.46cm]{{0.138}}\,\,\,\,\,\,\,\makebox[0.27cm]{\textcolor{gray}{58s}}} & \small{\makebox[0.46cm]{{0.130}}\,\,\,\,\,\,\,\makebox[0.27cm]{\textcolor{gray}{3.5m}}} & \small{\makebox[0.46cm]{{0.044}}\,\,\,\,\,\,\,\makebox[0.27cm]{\textcolor{gray}{5.5m}}} & \small{\makebox[0.46cm]{{0.038}}\,\,\,\,\,\,\,\makebox[0.27cm]{\textcolor{gray}{5.7m}}} & \small{\makebox[0.46cm]{\textbf{{0.033}}}\,\,\,\,\,\,\,\makebox[0.27cm]{\textcolor{gray}{6.3m}}} & \small{\makebox[0.46cm]{{0.043}}\,\,\,\,\,\,\,\makebox[0.27cm]{\textcolor{gray}{7.7m}}} \\
  \small{\SwimmingPool{}} & \small{\makebox[0.7cm][r]{\textcolor{gray}{236 MS}}}\hspace{-1.7mm} & \small{\makebox[0.46cm]{{0.688}}\,\,\,\,\,\,\,\makebox[0.27cm]{\textcolor{gray}{40s}}} & \small{\makebox[0.46cm]{{0.082}}\,\,\,\,\,\,\,\makebox[0.27cm]{\textcolor{gray}{63s}}} & \small{\makebox[0.46cm]{{0.077}}\,\,\,\,\,\,\makebox[0.40cm]{\textcolor{gray}{29m}} \,\textcolor{gray}{+}\,\, \makebox[0.27cm]{\textcolor{gray}{22s}}} & \small{\makebox[0.46cm]{{0.494}}\,\,\,\,\,\,\,\makebox[0.27cm]{\textcolor{gray}{41s}}} & \small{\makebox[0.46cm]{{0.198}}\,\,\,\,\,\,\,\makebox[0.27cm]{\textcolor{gray}{3.5m}}} & \small{\makebox[0.46cm]{{0.077}}\,\,\,\,\,\,\,\makebox[0.27cm]{\textcolor{gray}{3.0m}}} & \small{\makebox[0.46cm]{{0.073}}\,\,\,\,\,\,\,\makebox[0.27cm]{\textcolor{gray}{4.5m}}} & \small{\makebox[0.46cm]{\textbf{{0.066}}}\,\,\,\,\,\,\,\makebox[0.27cm]{\textcolor{gray}{5.1m}}} & \small{\makebox[0.46cm]{{0.087}}\,\,\,\,\,\,\,\makebox[0.27cm]{\textcolor{gray}{5.1m}}} \\
  \small{\VeachDoor{}} & \small{\makebox[0.7cm][r]{\textcolor{gray}{236 MS}}}\hspace{-1.7mm} & \small{\makebox[0.46cm]{{0.910}}\,\,\,\,\,\,\,\makebox[0.27cm]{\textcolor{gray}{38s}}} & \small{\makebox[0.46cm]{{0.227}}\,\,\,\,\,\,\,\makebox[0.27cm]{\textcolor{gray}{70s}}} & \small{---} & \small{\makebox[0.46cm]{{0.903}}\,\,\,\,\,\,\,\makebox[0.27cm]{\textcolor{gray}{41s}}} & \small{\makebox[0.46cm]{{0.716}}\,\,\,\,\,\,\,\makebox[0.27cm]{\textcolor{gray}{3.5m}}} & \small{\makebox[0.46cm]{{0.135}}\,\,\,\,\,\,\,\makebox[0.27cm]{\textcolor{gray}{9.0m}}} & \small{\makebox[0.46cm]{{0.135}}\,\,\,\,\,\,\,\makebox[0.27cm]{\textcolor{gray}{12m}}} & \small{\makebox[0.46cm]{\textbf{{0.099}}}\,\,\,\,\,\,\,\makebox[0.27cm]{\textcolor{gray}{13m}}} & \small{\makebox[0.46cm]{{0.137}}\,\,\,\,\,\,\,\makebox[0.27cm]{\textcolor{gray}{12m}}} \\
  \small{\WhiteRoom{}} & \small{\makebox[0.7cm][r]{\textcolor{gray}{236 MS}}}\hspace{-1.7mm} & \small{\makebox[0.46cm]{{0.102}}\,\,\,\,\,\,\,\makebox[0.27cm]{\textcolor{gray}{79s}}} & \small{\makebox[0.46cm]{{0.066}}\,\,\,\,\,\,\,\makebox[0.27cm]{\textcolor{gray}{1.9m}}} & \small{\makebox[0.46cm]{{0.076}}\,\,\,\,\,\,\makebox[0.40cm]{\textcolor{gray}{24m}} \,\textcolor{gray}{+}\,\, \makebox[0.27cm]{\textcolor{gray}{55s}}} & \small{\makebox[0.46cm]{{0.090}}\,\,\,\,\,\,\,\makebox[0.27cm]{\textcolor{gray}{79s}}} & \small{\makebox[0.46cm]{{0.093}}\,\,\,\,\,\,\,\makebox[0.27cm]{\textcolor{gray}{3.5m}}} & \small{\makebox[0.46cm]{{0.058}}\,\,\,\,\,\,\,\makebox[0.27cm]{\textcolor{gray}{7.0m}}} & \small{\makebox[0.46cm]{{0.046}}\,\,\,\,\,\,\,\makebox[0.27cm]{\textcolor{gray}{7.8m}}} & \small{\makebox[0.46cm]{\textbf{{0.040}}}\,\,\,\,\,\,\,\makebox[0.27cm]{\textcolor{gray}{8.3m}}} & \small{\makebox[0.46cm]{{0.045}}\,\,\,\,\,\,\,\makebox[0.27cm]{\textcolor{gray}{11m}}} \\
  \small{\GlossyCornellBox{}} & \small{\makebox[0.7cm][r]{\textcolor{gray}{1073 MS}}}\hspace{-1.7mm} & \small{\makebox[0.46cm]{{1.228}}\,\,\,\,\,\,\,\makebox[0.27cm]{\textcolor{gray}{1.9m}}} & \small{\makebox[0.46cm]{{0.141}}\,\,\,\,\,\,\,\makebox[0.27cm]{\textcolor{gray}{4.7m}}} & \small{\makebox[0.46cm]{{0.079}}\,\,\,\,\,\,\makebox[0.40cm]{\textcolor{gray}{6.0m}} \,\textcolor{gray}{+}\,\, \makebox[0.27cm]{\textcolor{gray}{1.7m}}} & \small{\makebox[0.46cm]{{1.200}}\,\,\,\,\,\,\,\makebox[0.27cm]{\textcolor{gray}{2.1m}}} & \small{\makebox[0.46cm]{{1.108}}\,\,\,\,\,\,\,\makebox[0.27cm]{\textcolor{gray}{15m}}} & \small{\makebox[0.46cm]{\textbf{{0.059}}}\,\,\,\,\,\,\,\makebox[0.27cm]{\textcolor{gray}{23m}}} & \small{\makebox[0.46cm]{{0.222}}\,\,\,\,\,\,\,\makebox[0.27cm]{\textcolor{gray}{30m}}} & \small{\makebox[0.46cm]{{0.065}}\,\,\,\,\,\,\,\makebox[0.27cm]{\textcolor{gray}{36m}}} & \small{\makebox[0.46cm]{{0.110}}\,\,\,\,\,\,\,\makebox[0.27cm]{\textcolor{gray}{33m}}} \\
  \bottomrule
\end{tabular*}
\end{table*}

\begin{figure*}[t]
    \vspace{-3mm}%
    \includegraphics[width=1\linewidth]{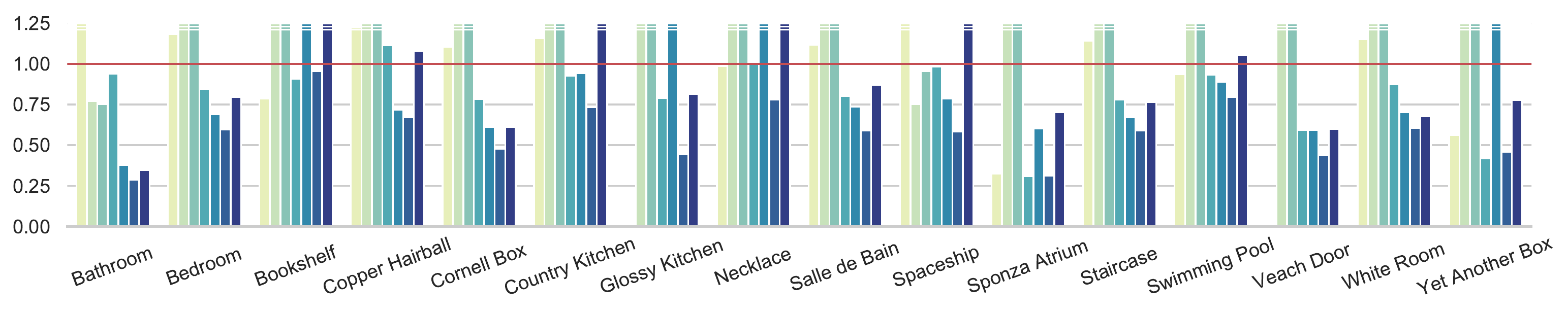}
    \vspace{-6mm}%
    \caption{
      MAPE achieved by the bidirectionally trained Gaussian mixture model by \citet{Vorba:2014:OnlineLearningPMMinLTS} (the \GlossyKitchen{} and \VeachDoor{} are omitted because of limitations of their implementation), the primary-sample-space method by \citet{Guo:2018},
      and our neural importance-sampling approaches
      on different scenes (the order and colors of bars follows \autoref{tab:one-blob-encoding}). The bars are normalized with respect to practical path guiding~\cite{mueller2017practical}; a height below $1$ signifies better performance. Some bars exceed outside of the displayed range; \autoref{tab:one-blob-encoding} provides the actual numbers.
      Primary-sample-space techniques generally perform worse than path-guiding approaches. The product-driven neural path guiding usually performs the best.
    }%
    \vspace{-1mm}%
    \label{fig:metrics-bar-chart}
\end{figure*}

\begin{figure*}[t]
  \vspace{-2mm}
  \hspace*{-2mm}\includegraphics[width=1.025\linewidth]{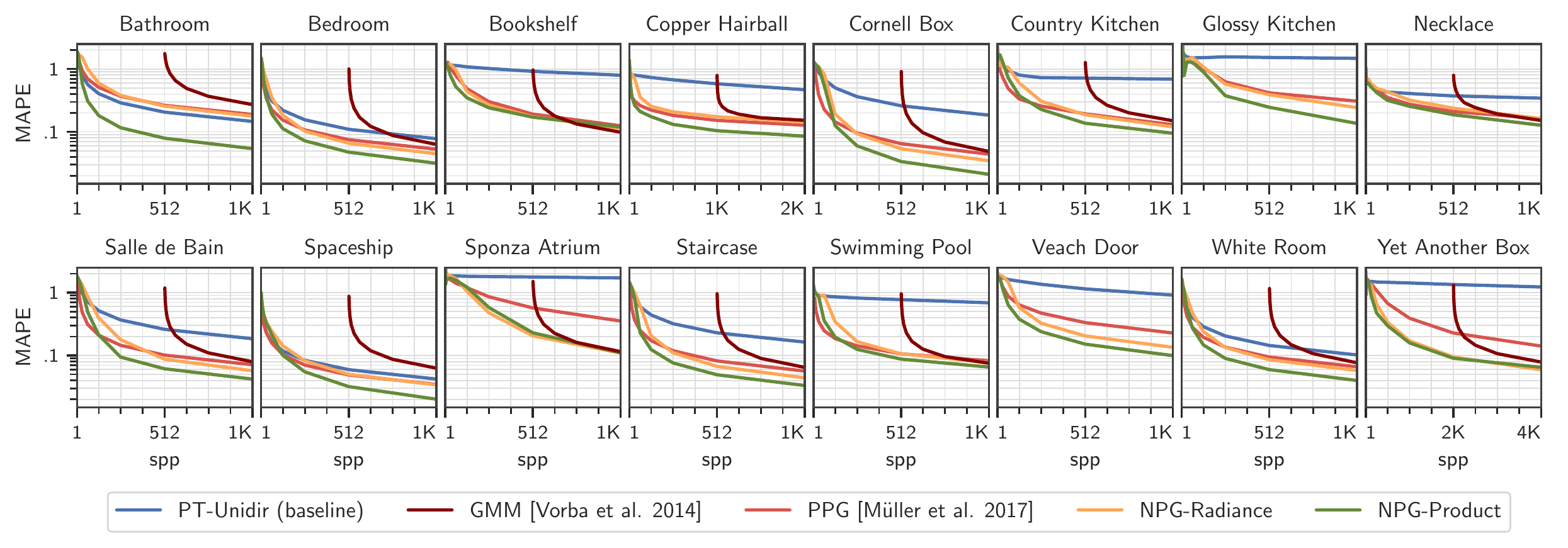}
  \vspace{-7mm}%
  \caption{
    \ADD{Convergence plots of unidirectional path tracing (PT-Unidir), practical path guiding (PPG)~\citep{mueller2017practical}, the algorithm of \citet{Vorba:2014:OnlineLearningPMMinLTS} (GMM), and our radiance- and product-based neural path guiding (NPG-Radiance and NPG-Product, respectively).
    We plot MAPE as a function of samples per pixel (spp) on a logarithmic scale.
    All guiding methods perform slightly worse than na\"{\i}ve path tracing initially, but overtake it rapidly on most scenes as they learn to importance sample.
    PPG tends to learn slightly faster than our NPG, but falls behind due to learning a worse distribution.
    The algorithm of \citet{Vorba:2014:OnlineLearningPMMinLTS} starts producing images only half-way through the sample budget as the first half of samples is used for offline pretraining.}
  }%
  \label{fig:convergence-plots}
  \vspace{-2mm}
\end{figure*}

\subsection{Experimental Setup}%
\label{sec:experimental-setup}

We implemented our technique in Tensorflow~\cite{tensorflow2015-whitepaper} and integrated it with the Mitsuba renderer~\cite{Mitsuba}.
Before we start rendering, we initialize the trainable parameters of our networks using Xavier initialization~\cite{Glorot:2010}.
While rendering the image, we optimize them using Adam~\cite{KingmaB14}.
Our rendering procedure is implemented as a hybrid CPU/GPU algorithm, tracing rays in large batches on the CPU while two GPUs perform all neural-network-related tasks.
One GPU is responsible for optimizing the MIS selection probabilities and evaluating and sampling from $\PdfOptimized$, while the other trains the networks using Monte Carlo estimates from completed paths.
Both GPUs use minibatch sizes of \num{100000} samples.
\ADD{Communication between the CPU and GPUs happens via asynchronous buffers to aid parallelization.
Computation of selection probabilities and $\PdfOptimized$-evaluation and $\PdfOptimized$-sampling communicate via asynchronous \emph{queues} that are processed as fast as possible.
Our training buffer is configured to always contain the latest \num{2000000} samples of which minibatches are \emph{randomly} selected for optimization.
This procedure decorrelates samples that are nearby in the image plane.}

In order to obtain the final image with $N$ samples, we perform $M = \lfloor \log_2 (N+1) \rfloor$ iterations with power-of-two sample counts $2^i; i \in \{0, \ldots, M\}$.
This approach was initially proposed by \citet{mueller2017practical}
to limit the impact of initial high-variance estimates on the final image.
In contrast to their work, we do not reset the learned distributions at every power-of-two iteration and keep training the same set of networks from start to finish.
Furthermore, instead of discarding the pixel estimates of earlier iterations, we weight the images produced within each iteration by their reciprocal mean pixel variance, which we estimate on-the-fly.
While this approach introduces bias, it is imperceptibly small in practice due to averaging across all pixels.
Furthermore, the bias vanishes as the quality of the variance estimate increases, making this approach consistent.
We apply the same weighting scheme to our implementation of the method by \citet{mueller2017practical} to ensure a fair comparison.

All results were produced on a workstation with two Intel Xeon E5-2680v3 CPUs (24 cores; 48 threads) and two NVIDIA Titan Xp GPUs.
Due to the combined usage of both the CPU and the GPU, runtimes of different techniques depend strongly on the particular setup. We therefore \ADD{focus} on comparing the performance using \emph{equal-sample-count} metrics that are \ADD{independent of} hardware.
Absolute timings \ADD{and an equal-time comparison of a subset of the scenes and methods} are provided for completeness.

We quantify the error using the \emph{mean absolute percentage error} (MAPE), which is defined as $\frac{1}{N} \sum_{i=1}^N |v_i - \hat{v}_i| / (\hat{v}_i + \epsilon)$, where $\hat{v}_i$ is the value of the $i$-th pixel in the ground-truth image, $v_i$ is the value of the $i$-th rendered pixel, and $\epsilon = 0.01$ serves the dual objective of avoiding the singularity at $\hat{v}_i = 0$ and down-weighting close-to-black pixels.
We use a relative metric to avoid putting overly much emphasis on bright image regions.
We also evaluated L1, L2, \emph{mean relative squared error} (MRSE)~\cite{rousselle11gem}, \emph{symmetric MAPE} (SMAPE), and SSIM, which all can be inspected as false-color maps and aggregates in the supplemented image viewer.

\subsection{Results}%
\label{sec:results}

In order to best illustrate the benefits of different neural-importance-sampling approaches, we compare their performance when used on top of a unidirectional path tracer that uses BSDF sampling only. While none of the algorithms utilized next-event estimation (including prior works) to emphasize the performance of individual path sampling/guiding approaches, we recommend using it in practice for best performance.
In the following, all results with our piecewise-polynomial coupling functions utilize $\nlayers = 4$ coupling layers.
We use $1023$ samples per pixel (spp) on all scenes except for the \CopperHairball{} ($2047$ spp) and \GlossyCornellBox{} ($4095$ spp).
\ADD{The images were rendered at resolutions $640\times 360$ and $512\times 512$.
It is worth noting that since the quality of learned distribution depends primarily on the \emph{total} number of drawn samples (reported as ``mega samples'' (MS)) rendering at higher resolutions yields high-quality distributions ``faster'', i.e.\ with fewer spp.}

\paragraph{Path Sampling}
In \autoref{fig:pss_pg}, we study \emph{primary-sample-space path sampling} (PSSPS) using our implementation of the technique by \citet{Guo:2018} and our \emph{neural path sampling} (NPS) with piecewise-polynomial and affine coupling transforms.
We apply the sampling to only a limited number of non-specular interactions in the beginning of each path and sample all other interactions using uniform random numbers.
We experimented with three different prefix dimensionalities: 2D, 4D, and 6D, which correspond to importance sampling path prefixes of 1, 2, and 3 non-specular vertices, respectively.
As shown in the figure, going beyond 4D brings typically little improvement in tested scenes, except for the highlights in the \Spaceship{}, where even longer paths are correlated thanks to highly-glossy interactions with the glass of the cockpit\footnote{Due to faster training of lower-dimensional distributions, the 2D case still has the least overall noise in the \Spaceship{} scene.}.
This confirms the observation of \citet{Guo:2018} that cases where more than two bounces are needed to connect to the light source offer minor to no improvement.
We speculate that the poor performance in higher dimensions is due to the relatively weak correlation between path geometries and PSS coordinates, i.e.\ paths with similar PSS coordinates may have drastically different vertex positions.
The correlation tends to weaken at each additional bounce (e.g.\ in the diffuse \CornellBox{}) unless the underlying path importance-sampling technique preserves path coherence.

\begin{figure*}
  \vspace{-1mm}
  \setlength{\fboxrule}{1pt}%
\setlength{\insetvsep}{20pt}%
\setlength{\tabcolsep}{0.75pt}%
\renewcommand{\arraystretch}{0.75}%
\small%
\begin{tabular}{cccccccccccc}
  \multicolumn{4}{c}{\CornellBox{}}&\multicolumn{4}{c}{\CountryKitchen{}}&\multicolumn{4}{c}{\SwimmingPool{}}\\
  \multicolumn{4}{c}{\begin{overpic}[width=0.33\textwidth]{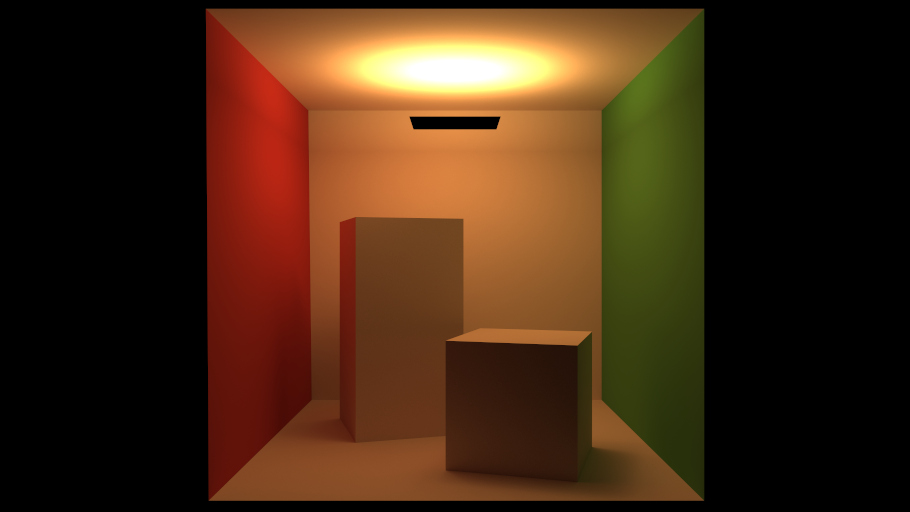}\put(47.5,47){ \tikz\draw[red,fill=red] (0,0) circle (0.5ex);}\put(44,22){ \tikz\draw[orange,fill=orange] (0,0) circle (0.5ex);}\end{overpic}}&\multicolumn{4}{c}{\begin{overpic}[width=0.33\textwidth]{images/references/kitchen-Reference.jpg}\put(25,38){ \tikz\draw[red,fill=red] (0,0) circle (0.5ex);}\put(88,17){ \tikz\draw[orange,fill=orange] (0,0) circle (0.5ex);}\end{overpic}}&\multicolumn{4}{c}{\begin{overpic}[width=0.33\textwidth]{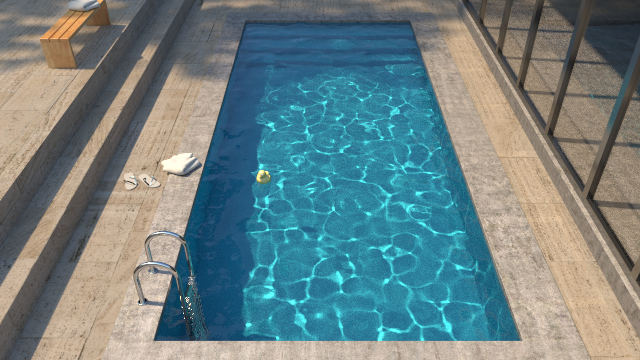}\put(82.5,44){ \tikz\draw[red,fill=red] (0,0) circle (0.5ex);}\put(60,25){ \tikz\draw[orange,fill=orange] (0,0) circle (0.5ex);}\end{overpic}}\\
  \color{red}\fbox{\begin{overpic}[width=0.07666666666666667\textwidth]{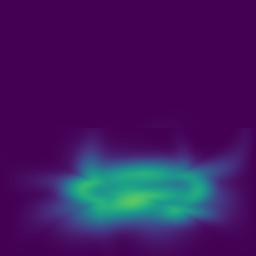} \put(2,81) { \footnotesize MAPE: 3.17 } \end{overpic}}&\color{red}\fbox{\begin{overpic}[width=0.07666666666666667\textwidth]{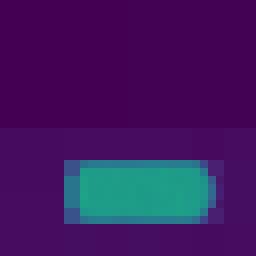} \put(60,81) { \footnotesize  2.89 } \end{overpic}}&\color{red}\fbox{\begin{overpic}[width=0.07666666666666667\textwidth]{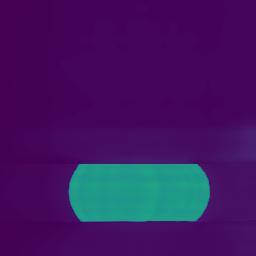} \put(60,81) { \footnotesize  0.32 } \end{overpic}}&\color{red}\fbox{\begin{overpic}[width=0.07666666666666667\textwidth]{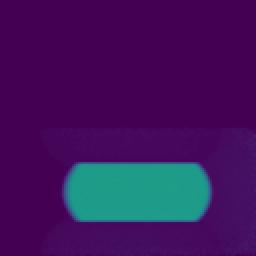}  \end{overpic}}&\color{red}\fbox{\begin{overpic}[width=0.07666666666666667\textwidth]{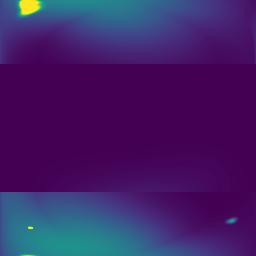} \put(60,81) { \footnotesize  1.42 } \end{overpic}}&\color{red}\fbox{\begin{overpic}[width=0.07666666666666667\textwidth]{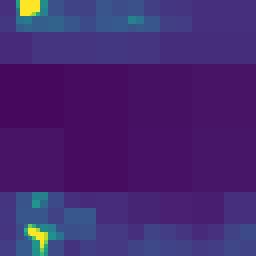} \put(60,81) { \footnotesize  1.36 } \end{overpic}}&\color{red}\fbox{\begin{overpic}[width=0.07666666666666667\textwidth]{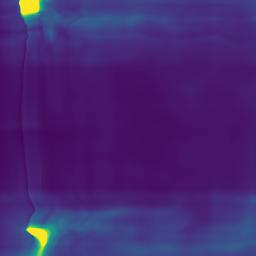} \put(60,81) { \footnotesize  1.95 } \end{overpic}}&\color{red}\fbox{\begin{overpic}[width=0.07666666666666667\textwidth]{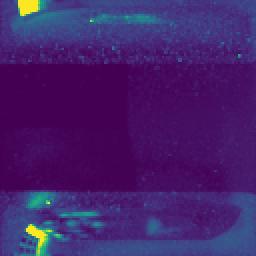}  \end{overpic}}&\color{red}\fbox{\begin{overpic}[width=0.07666666666666667\textwidth]{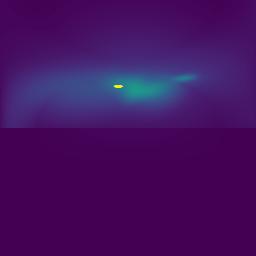} \put(60,81) { \footnotesize  1.53 } \end{overpic}}&\color{red}\fbox{\begin{overpic}[width=0.07666666666666667\textwidth]{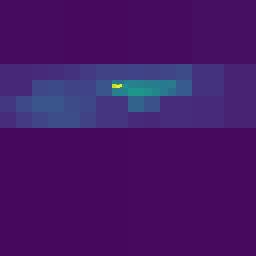} \put(60,81) { \footnotesize  2.15 } \end{overpic}}&\color{red}\fbox{\begin{overpic}[width=0.07666666666666667\textwidth]{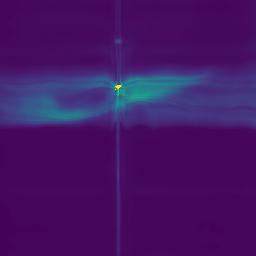} \put(60,81) { \footnotesize  0.89 } \end{overpic}}&\color{red}\fbox{\begin{overpic}[width=0.07666666666666667\textwidth]{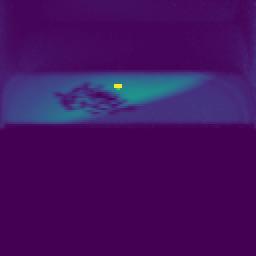}  \end{overpic}}\\
  \color{orange}\fbox{\begin{overpic}[width=0.07666666666666667\textwidth]{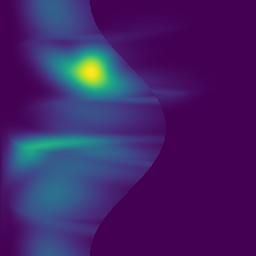} \put(2,5) { \footnotesize MAPE: 4.68 } \end{overpic}}&\color{orange}\fbox{\begin{overpic}[width=0.07666666666666667\textwidth]{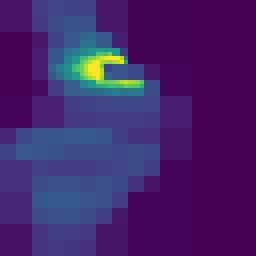} \put(60,5) { \footnotesize  5.00 } \end{overpic}}&\color{orange}\fbox{\begin{overpic}[width=0.07666666666666667\textwidth]{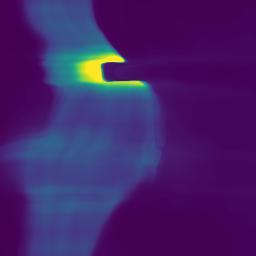} \put(60,5) { \footnotesize  1.30 } \end{overpic}}&\color{orange}\fbox{\begin{overpic}[width=0.07666666666666667\textwidth]{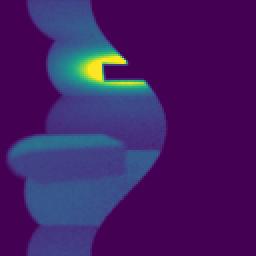}  \end{overpic}}&\color{orange}\fbox{\begin{overpic}[width=0.07666666666666667\textwidth]{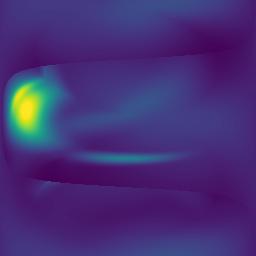} \put(60,5) { \footnotesize  2.87 } \end{overpic}}&\color{orange}\fbox{\begin{overpic}[width=0.07666666666666667\textwidth]{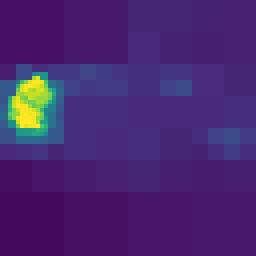} \put(60,5) { \footnotesize  0.81 } \end{overpic}}&\color{orange}\fbox{\begin{overpic}[width=0.07666666666666667\textwidth]{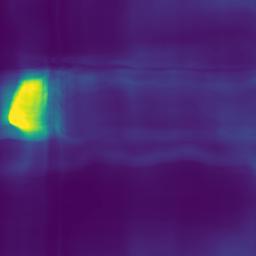} \put(60,5) { \footnotesize  0.72 } \end{overpic}}&\color{orange}\fbox{\begin{overpic}[width=0.07666666666666667\textwidth]{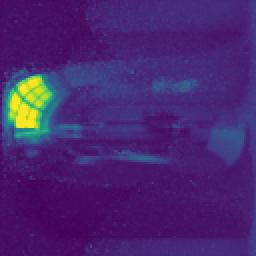}  \end{overpic}}&\color{orange}\fbox{\begin{overpic}[width=0.07666666666666667\textwidth]{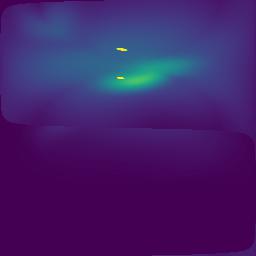} \put(60,5) { \footnotesize  1.32 } \end{overpic}}&\color{orange}\fbox{\begin{overpic}[width=0.07666666666666667\textwidth]{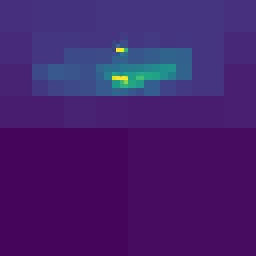} \put(60,5) { \footnotesize  1.03 } \end{overpic}}&\color{orange}\fbox{\begin{overpic}[width=0.07666666666666667\textwidth]{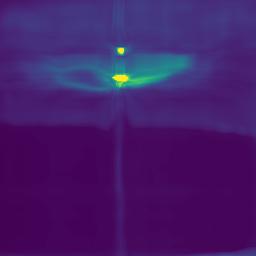} \put(60,5) { \footnotesize  0.69 } \end{overpic}}&\color{orange}\fbox{\begin{overpic}[width=0.07666666666666667\textwidth]{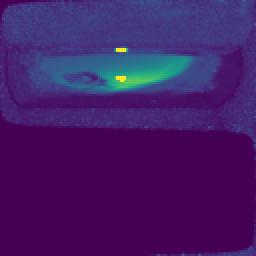}  \end{overpic}}\\
  GMM&PPG&NPG-Rad.&Reference&GMM&PPG&NPG-Rad.&Reference&GMM&PPG&NPG-Rad.&Reference\\
\end{tabular}
\unsetInset{A}%
\unsetInset{B}%

  \vspace{-1mm}
  \caption{
    Directional radiance distributions.
    From left to right: we visualize the distributions learned by a Gaussian mixture model (GMM) \citep{Vorba:2014:OnlineLearningPMMinLTS}, an SD-tree (PPG) \citep{mueller2017practical}, our neural path-guiding approach trained on radiance (NPG-Rad.), and a spatial binary tree with a directional regular $128\times128$ grid (Reference).
    The first three approaches were trained with an equal sample count and require roughly equal amounts of memory in the above scenes (around 10 MB).
    We used $2^{16}$ samples per pixel to generate the reference distributions, which require roughly \SI{5}{\giga\byte} per scene.
    \ADD{Despite its large computational cost, the reference solution is still slightly blurred (see e.g.\ \CornellBox, red inset).
    Our approach produces the most accurate distributions in the majority of cases, measured here using
    the mean average percentage error (MAPE). Unlike the competing techniques, however, we learn a continuous function in both the spatial and the directional domain, allowing for a smaller amount of blur in some cases, e.g.\ in the \CornellBox.}
  }%
  \label{fig:distributions}
\end{figure*}
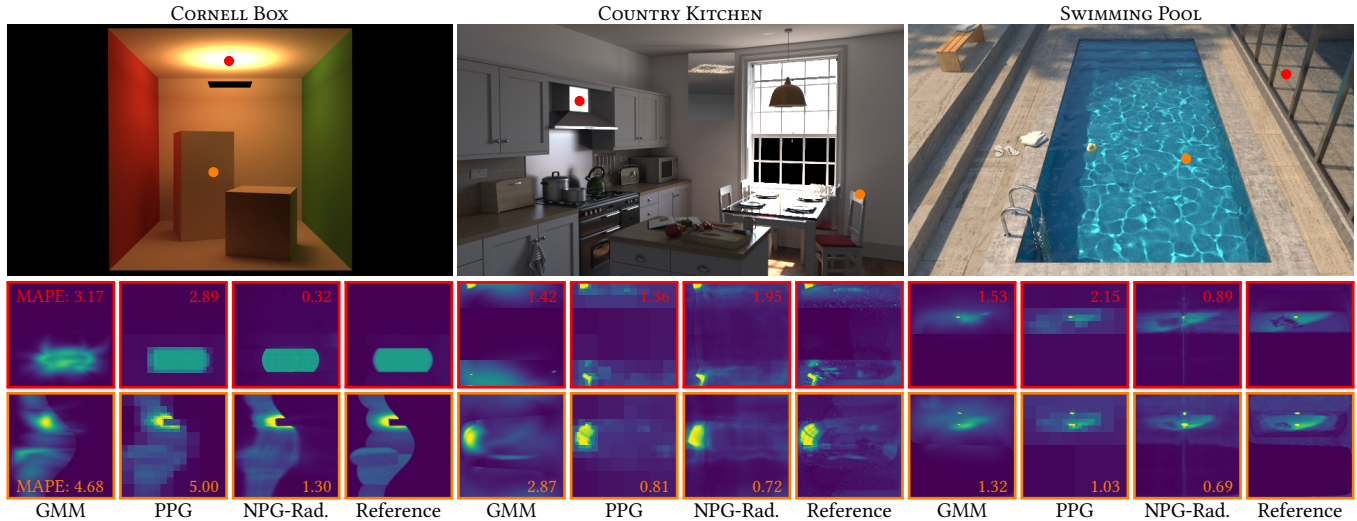

\paragraph{Path Guiding}
In \autoref{fig:path_guiding}, we analyze the performance of different path-guiding approaches, referring to ours as \emph{neural path guiding} (NPG). We compare our work to the respective authors' implementations of \emph{practical path guiding} (PPG) by \citet{mueller2017practical} and the bidirectionally trained \emph{Gaussian mixture model} (GMM) by \citet{Vorba:2014:OnlineLearningPMMinLTS}, which are both learning sampling distributions that are, in contrast to ours, proportional to incident radiance only.
We extended the GMM implementation to (oriented) spherical domains.

To isolate the benefits of using NICE with piecewise-quadratic coupling layers, 
we created a variant of our approach, NPG-Radiance, that learns PDFs proportional to incident radiance only and without the MIS-aware optimization.
The \emph{radiance-driven} neural path guiding outperforms PPG and GMM in \ADD{$13$} out of $16$ scenes and follows closely in the others (\Bookshelf, \CopperHairball, \Necklace), making it the most robust method out of the three \ADD{radiance-only approaches}.

The performance of our neural approach is further increased by learning and sampling proportional to the full product and incorporating MIS into the optimization---this technique yields the best results in nearly all scenes.
As seen on the \CopperHairball{}, our technique can learn the product even under high-frequency spatial variation by passing the surface normal as an additional input to the networks.
We trained all techniques with the same number of samples as used for rendering.
The SD-tree of PPG and our neural networks used between \SI{5}{\mega\byte} and \SI{10}{\mega\byte}, the Gaussian mixture model used between \SI{5}{\mega\byte} and \SI{118}{\mega\byte}.

\autoref{tab:one-blob-encoding} reports the MAPE metric and absolute timings of $9$ methods on a set of $16$ tested scenes.
We also visualize the results of all methods using bar charts in \autoref{fig:metrics-bar-chart}; the height is normalized with respect to PPG.\@
\ADD{We exclude GMM results for the \GlossyKitchen{} and \VeachDoor{} as we could not obtain representative results on these scenes.}
Path sampling in PSS typically yields significantly worse results than all path-guiding approaches.
Neural path guiding always benefits (sometimes significantly) from encoding the inputs with one-blob encoding as opposed to inputing raw (scalar) values. \ADD{This version performs the best except for two scenes where radiance-only NPG scored better.}

\paragraph{Empirical Convergence Analysis}
\ADD{Convergence plots in \autoref{fig:convergence-plots} provide further insight into the differences between unidirectional path tracing, the GMMs by \citet{Vorba:2014:OnlineLearningPMMinLTS}, PPG by \citet{mueller2017practical}, and our radiance- and product-based neural-path-guiding algorithms.
In most cases, the online path-guiding algorithms quickly learn a superior sampling density compared to the baseline path tracer.
The GMM algorithm---being trained offline---is inferior in the beginning of rendering, but produces competitive results after the total sample budget is exhausted.
Our neural approaches produce the best results most of the time, with our product-based approach usually being superior to our radiance-based approach.}

\paragraph{Optimizing KL vs.\ $\chi^2$ Divergence}
We compare variants of product-driven neural path guiding optimized using the Kullback-Leibler (KL) and $\chi^2$ divergences during training.
The squared Monte Carlo weight in the $\chi^2$ gradient causes a large variance, making it difficult to optimize with.
We remedy this problem by clipping the minibatch gradient norm to a maximum of $50$.
While the $\chi^2$ divergence in theory minimizes the estimator variance most directly (see~\autoref{sec:optimizing-var}), it performs worse in practice according to \ADD{all tested metrics on all test scenes (see~\autoref{tab:one-blob-encoding} and the supplemented image viewer)}.
A notable aspect of optimizing the $\chi^2$ divergence is that it tends to produce results with higher variance overall, but fewer and less-extreme outliers.

\paragraph{Accuracy of Learned Distributions}
We visualize learned radiance distributions in \autoref{fig:distributions}, comparing our path-guiding neural distributions against the SD-tree, the GMM, and a reference solution.
\ADD{In most cases, NPG learns more accurate directional distributions than the competing methods in terms of the MAPE metric.}
Additionally, NPG produces a spatially \emph{and} directionally continuous function; we illustrate the spatial continuity in the supplementary video.

\begin{figure}
  \setlength{\fboxrule}{6pt}%
  \setlength{\insetvsep}{20pt}%
  \setlength{\tabcolsep}{1.5pt}%
  \renewcommand{\arraystretch}{1.4}%
  \small%
  \hspace*{-0.5mm}%
  \begin{tabular}{cc}
    \small Fixed selection probabilities & \small Learned selection probabilities \\ 
    \includegraphics[trim={190px 140px 320px 120px}, clip, width=0.495\columnwidth]{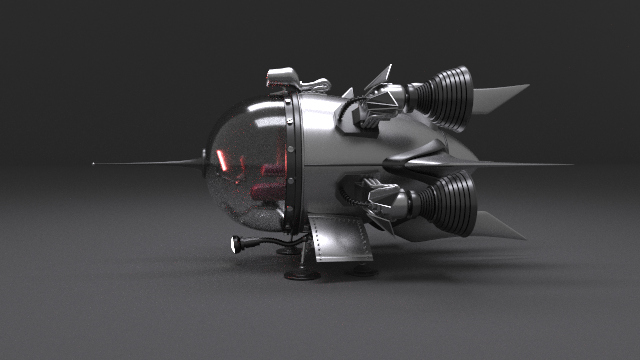} &
    \includegraphics[trim={190px 140px 320px 120px}, clip, width=0.495\columnwidth]{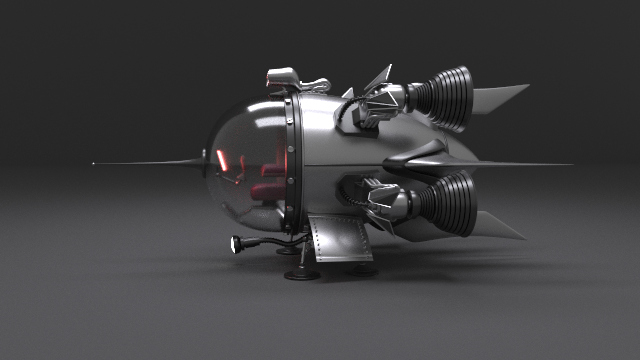} \\[-5pt]
  \end{tabular}
  \vspace{-1mm}
  \caption{%
    Learning MIS selection probabilities---even with product-driven path guiding---leads to significantly better results on the \Spaceship{} cockpit, where BSDF sampling is near optimal.
  }%
  \vspace{-2mm}%
  \label{fig:mis-weights}
\end{figure}

\paragraph{Learned Selection Probabilities and MIS-Aware Optimization}
In~\autoref{fig:mis-weights}, we demonstrate the increased robustness of neural path guiding offered by learning optimal selection probabilities.
The impact is particularly noticeable on the cockpit of the spaceship seen through specular interactions, which are handled nearly optimally by sampling the material BSDF.\@
In this region, a standard path tracer outperforms the learned sampling PDFs.
\ADD{With MIS-aware optimization---including the learning of selection probabilities---}the system downweighs the contribution of the learned PDF on the cockpit, but increases it in regions where it is more accurate, resulting in significantly improved results overall.

\begin{figure}
  \vspace{-3.5mm}%
  \setlength{\fboxrule}{6pt}%
\setlength{\insetvsep}{20pt}%
\setlength{\tabcolsep}{-1pt}%
\renewcommand{\arraystretch}{1.4}%
\small%
\hspace*{-2mm}%
  \begin{tabular}{cccc}
  & & From scratch & Reused \\[-1.2mm]
    \setInset{A}{red}{260}{105}{53}{38}%
    \setInset{B}{orange}{280}{260}{53}{38}%
    \rotatebox{90}{\hspace{-1.70cm}\Spaceship{}}\hspace{0.16cm} & 
    \addBeautyCrop{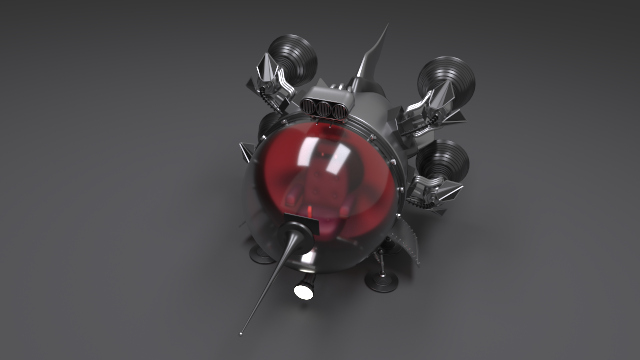}{0.27}{640}{360}{0}{0}{640}{360} &
    \addInsets{images/fig-reuse/spaceship-Ours-FPS.jpg} &
    \addInsets{images/fig-reuse/spaceship-Ours-FPS-reuse.jpg} \\
    \setInset{A}{red}{450}{90}{53}{38}%
    \setInset{B}{orange}{20}{295}{53}{38}%
    \rotatebox{90}{\hspace{-2.05cm}\SwimmingPool{}}\hspace{0.16cm} & 
    \addBeautyCrop{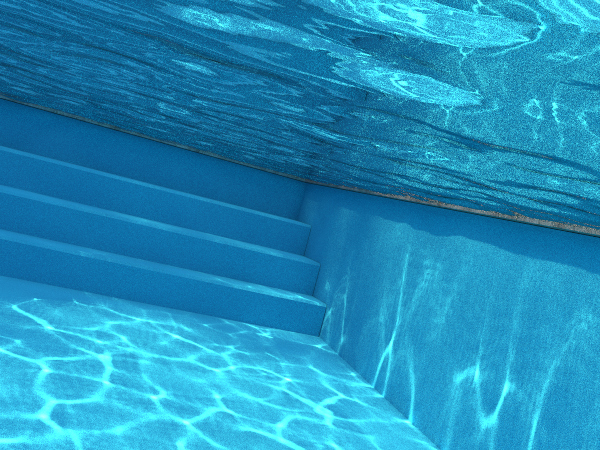}{0.27}{600}{450}{0}{0}{600}{338} &
    \addInsets{images/fig-reuse/pool-Ours-FPS.jpg} &
    \addInsets{images/fig-reuse/pool-Ours-FPS-reuse.jpg} \\
    \setInset{A}{red}{140}{110}{53}{38}%
    \setInset{B}{orange}{460}{300}{53}{38}%
    \rotatebox{90}{\hspace{-2.20cm}\CountryKitchen{}}\hspace{0.16cm} & 
    \addBeautyCrop{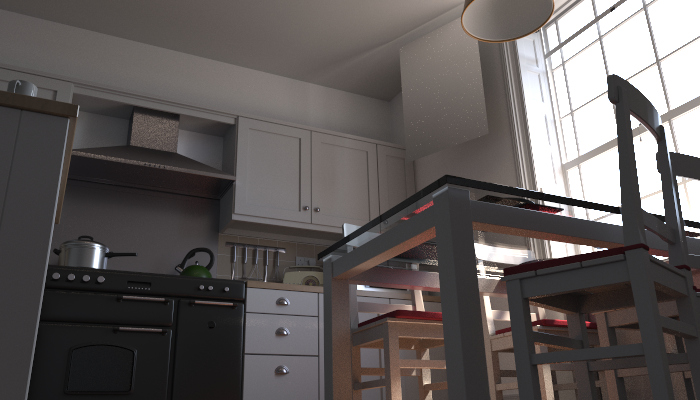}{0.27}{700}{400}{0}{0}{700}{394} &
    \addInsets{images/fig-reuse/kitchen-Ours-FPS.jpg} &
    \addInsets{images/fig-reuse/kitchen-Ours-FPS-reuse.jpg} \\
  \end{tabular}
  \unsetInset{A}
  \unsetInset{B}
  \vspace{-2mm}%
  \caption{%
    Learned distributions can be reused for novel camera views. The right column shows results where the network weights were initialized with weights learned for camera views in \autoref{fig:path_guiding}.
  }%
  \label{fig:reuse}
  \vspace{-1mm}%
\end{figure}

\paragraph{Weight Reuse Across Camera Views}
\autoref{fig:reuse} demonstrates the benefits of reusing network weights, optimized for a particular camera view, in a novel view of the scene.
We took network weights that resulted from generating images for \autoref{fig:path_guiding} as the \emph{initial} weights for rendering images in the right column of \autoref{fig:reuse}.
Similarly to training from scratch, we keep optimizing the networks.
If the initial distributions are already a good fit, our weighting by the reciprocal mean pixel variance automatically keeps initial pixel estimates rather than discarding them.

\begin{figure}
  \vspace{-2mm}%
  \setlength{\fboxrule}{6pt}%
\setlength{\insetvsep}{20pt}%
\setlength{\tabcolsep}{-1pt}%
\renewcommand{\arraystretch}{1.0}%
\small%
  \hspace*{-2mm}%
  \begin{tabular}{cccc}
    & & w/o features & w/ features \\
    \setInset{A}{red}{400}{110}{53}{38}%
    \setInset{B}{orange}{340}{290}{53}{38}%
    \rotatebox{90}{\hspace{-2.15cm}\CopperHairball{}}\hspace{0.16cm} & 
    \addBeautyCrop{images/references/hairball-Reference.jpg}{0.27}{640}{360}{0}{0}{640}{360} &
    \addInsets{images/fig-product/hairball-Ours-FPS.jpg} &
    \addInsets{images/fig-product/hairball-Ours-FPS-Feat.jpg} \\
    & \multicolumn{1}{r}{MAPE:} & 0.0994 & \textbf{0.0795} \\[2pt]
    \setInset{A}{red}{400}{110}{53}{38}%
    \setInset{B}{orange}{340}{290}{53}{38}%
    \rotatebox{90}{\hspace{-2.20cm}\AluminumSphere{}}\hspace{0.16cm} & 
    \addBeautyCrop{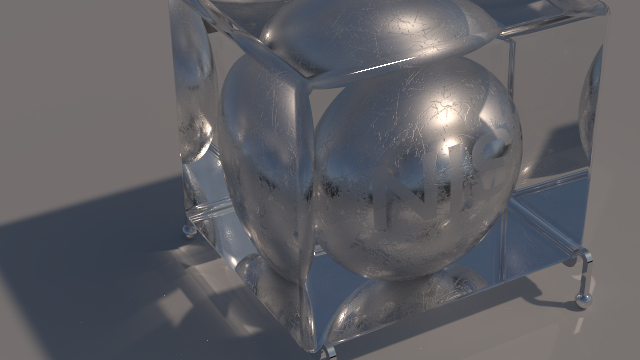}{0.27}{640}{360}{0}{0}{640}{360} &
    \addInsets{images/fig-product/aluminum-sphere-Ours-FPS.jpg} &
    \addInsets{images/fig-product/aluminum-sphere-Ours-FPS-Feat.jpg} \\
    & \multicolumn{1}{r}{MAPE:} & 0.0735 & \textbf{0.0722} \\[-5pt]
  \end{tabular}
  \unsetInset{A}
  \unsetInset{B}

  \vspace{-2mm}%
  \caption{%
    \ADDTWO{Product-driven neural path guiding by itself has difficulties capturing high-frequency material properties (left).
    Passing material properties as additional input features enables the neural networks to learn parts of the appearance in the potentially lower-frequency material-parameter space (right), leading to a slightly better fit and thereby slightly reduced noise.}
  }%
  \vspace{-1mm}%
  \label{fig:product}
\end{figure}

\paragraph{High-Frequency Material Parameters}
\ADDTWO{In \autoref{fig:product}, we show the benefits of passing spatially high-frequency material parameters (e.g.\ the surface normal and roughness) as additional network inputs.
When material parameters are \emph{not} passed, the network must learn the product distribution purely as a function of spatio-directional coordinates, which is difficult.
However, when the network receives material parameters as input, it can learn a potentially lower-frequency appearance model that exists partially in the material parameters' space, similar to the explicit factorization present in other product-guiding methods~\citep{herholz2016,Herholz:2018}.
In contrast to our method, Herholz et al.'s techniques explicitly factorize incident radiance and the BSDF which avoids this problem entirely, potentially achieving better fits than shown in the figure.}

\begin{figure*}
  \vspace{-1mm}%
  \setlength{\fboxrule}{6pt}%
\setlength{\insetvsep}{20pt}%
\setlength{\tabcolsep}{-1pt}%
\renewcommand{\arraystretch}{1}%
\small%
\begin{minipage}{.498\textwidth}%
  \hspace*{-2mm}%
  \begin{tabular}{ccccc}
    & & PT-Unidir & PPG & NPG-Product \\
    \setInset{A}{red}{55}{140}{40}{38}%
    \setInset{B}{orange}{440}{120}{40}{38}%
    \rotatebox{90}{\hspace{-1.75cm}\Bathroom{}}\hspace{0.16cm} & 
    \addBeautyCrop{images/references/glossy-bathroom-Reference.jpg}{0.457}{640}{360}{20}{0}{520}{360} &
    \addInsets{images/fig-equal-time/glossy-bathroom-PT-unidir.jpg} &
    \addInsets{images/fig-equal-time/glossy-bathroom-PPG.jpg} &
    \addInsets{images/fig-equal-time/glossy-bathroom-Ours-FPS.jpg} \\
    & \multicolumn{1}{r}{12 minutes \hspace{20mm} MAPE:} & \textbf{0.0512} & 0.1018 & 0.0542 \\[-2pt]
    & \multicolumn{1}{r}{Mega samples:} & 2164 & 860 & 236 \\[2pt]
    \setInset{A}{red}{150}{220}{40}{38}%
    \setInset{B}{orange}{460}{220}{40}{38}%
    \rotatebox{90}{\hspace{-2.10cm}\SalleDeBain{}}\hspace{0.16cm} & 
    \addBeautyCrop{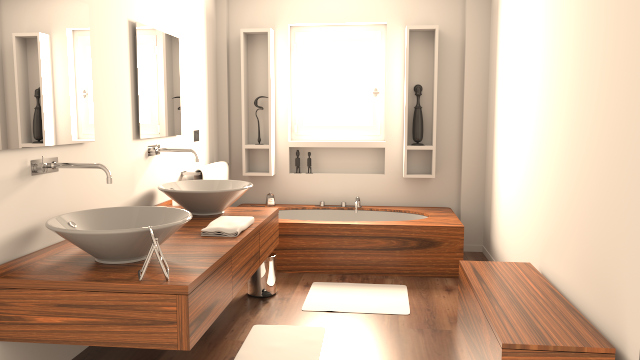}{0.457}{640}{360}{60}{0}{520}{360} &
    \addInsets{images/fig-equal-time/bathroom2-PT-unidir.jpg} &
    \addInsets{images/fig-equal-time/bathroom2-PPG.jpg} &
    \addInsets{images/fig-equal-time/bathroom2-Ours-FPS.jpg} \\
    & \multicolumn{1}{r}{6.1 minutes \hspace{20mm} MAPE:} & 0.0672 & 0.0436 & \textbf{0.0421} \\[-2pt]
    & \multicolumn{1}{r}{Mega samples:} & 1875 & 661 & 236 \\[2pt]
    \setInset{A}{red}{55}{220}{40}{38}%
    \setInset{B}{orange}{430}{90}{40}{38}%
    \rotatebox{90}{\hspace{-1.75cm}\Bookshelf{}}\hspace{0.16cm} & 
    \addBeautyCrop{images/references/bookshelf-Reference.jpg}{0.457}{640}{360}{0}{0}{520}{360} &
    \addInsets{images/fig-equal-time/bookshelf-PT-unidir.jpg} &
    \addInsets{images/fig-equal-time/bookshelf-PPG.jpg} &
    \addInsets{images/fig-equal-time/bookshelf-Ours-FPS.jpg} \\
    & \multicolumn{1}{r}{10 minutes \hspace{20mm} MAPE:} & 0.3765 & \textbf{0.0677} & 0.1196 \\[-2pt]
    & \multicolumn{1}{r}{Mega samples:} & 2157 & 664 & 236 \\[-5pt]
  \end{tabular}
  \unsetInset{A}
  \unsetInset{B}
\end{minipage}
\begin{minipage}{.498\textwidth}%
  \hspace*{0mm}%
  \begin{tabular}{ccccc}
    & & PT-Unidir & PPG & NPG-Product \\
    \setInset{A}{red}{130}{140}{40}{38}%
    \setInset{B}{orange}{320}{235}{40}{38}%
    \rotatebox{90}{\hspace{-1.95cm}\WhiteRoom{}}\hspace{0.16cm} & 
    \addBeautyCrop{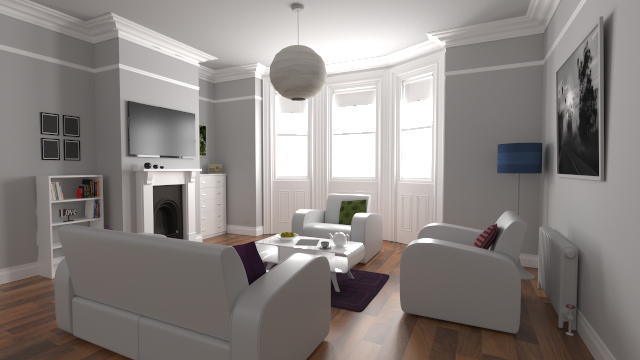}{0.457}{640}{360}{60}{0}{520}{360} &
    \addInsets{images/fig-equal-time/living-room-2-PT-unidir.jpg} &
    \addInsets{images/fig-equal-time/living-room-2-PPG.jpg} &
    \addInsets{images/fig-equal-time/living-room-2-Ours-FPS.jpg} \\
    & \multicolumn{1}{r}{8.3 minutes \hspace{20mm} MAPE:} & \textbf{0.0396} & 0.0396 & 0.0400 \\[-2pt]
    & \multicolumn{1}{r}{Mega samples:} & 1604 & 668 & 236 \\[2pt]
    \setInset{A}{red}{190}{190}{40}{38}%
    \setInset{B}{orange}{540}{310}{40}{38}%
    \rotatebox{90}{\hspace{-1.90cm}\VeachDoor{}}\hspace{0.16cm} & 
    \addBeautyCrop{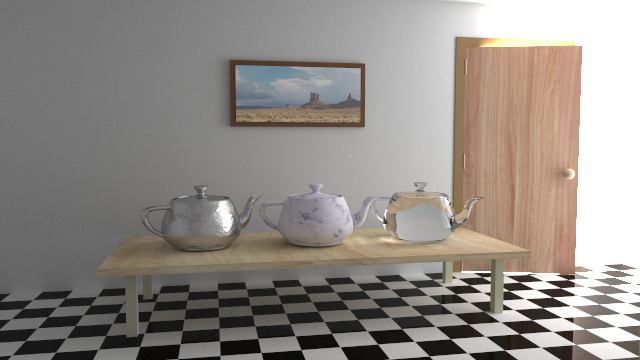}{0.457}{640}{360}{80}{0}{520}{360} &
    \addInsets{images/fig-equal-time/veach-door-PT-unidir.jpg} &
    \addInsets{images/fig-equal-time/veach-door-PPG.jpg} &
    \addInsets{images/fig-equal-time/veach-door-Ours-FPS.jpg} \\
    & \multicolumn{1}{r}{13 minutes \hspace{20mm} MAPE:} & 0.2317 & \textbf{0.0857} & 0.0995 \\[-2pt]
    & \multicolumn{1}{r}{Mega samples:} & 5413 & 1540 & 236 \\[2pt]
    \setInset{A}{red}{160}{245}{40}{38}%
    \setInset{B}{orange}{450}{225}{40}{38}%
    \rotatebox{90}{\hspace{-2.10cm}\GlossyKitchen{}}\hspace{0.16cm} & 
    \addBeautyCrop{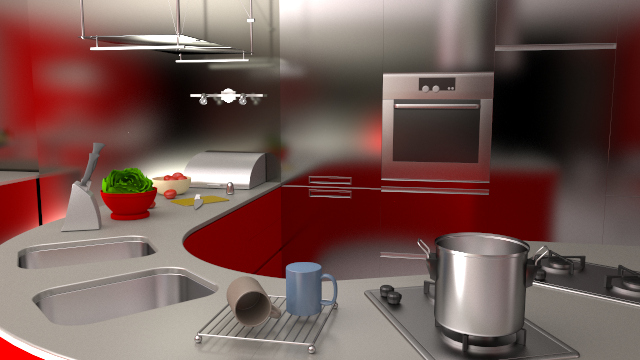}{0.457}{640}{360}{60}{0}{520}{360} &
    \addInsets{images/fig-equal-time/glossy-kitchen-PT-unidir.jpg} &
    \addInsets{images/fig-equal-time/glossy-kitchen-PPG.jpg} &
    \addInsets{images/fig-equal-time/glossy-kitchen-Ours-FPS.jpg} \\
    & \multicolumn{1}{r}{11 minutes \hspace{20mm} MAPE:} & 1.3179 & \textbf{0.1343} & 0.1363 \\[-2pt]
    & \multicolumn{1}{r}{Mega samples:} & 2222 & 1081 & 236 \\[-5pt]
  \end{tabular}
  \unsetInset{A}
  \unsetInset{B}
\end{minipage}%

  \caption{%
    \ADD{Equal-time comparison of unidirectional path tracing (PT-Unidir), practical path guiding (PPG)~\citep{mueller2017practical}, and our product-driven neural path guiding (NPG-Product).
    Despite the large computation cost of NPG, it performs competitively with PPG and outperforms unidirectional path tracing in scenes with difficult light transport (bottom two rows, sorted by difficulty in ascending order).
    The radiance-driven PPG algorithm tends to perform best because of its low computational cost, except when incident radiance is a poor approximation of the product (top row).}
  }%
  \label{fig:equal-time}
\end{figure*}

\paragraph{Equal-Time Comparison}
Lastly, we analyze the computational overhead of neural importance sampling in an equal-time comparison of unidirectional path tracing, PPG, and our product-driven NPG;\@ see~\autoref{fig:equal-time}.
All techniques utilize a CPU for tracing paths.
In addition, PPG uses the CPU for building and sampling the SD-tree, while our NPG also leverages two GPUs.

The radiance-driven PPG often performs best due to its small computational overhead, except when light transport is simple and/or the radiance-driven distribution is a poor fit to the product (e.g.\ the scenes in the top row).
Despite utilizing two extra GPUs, the product-driven NPG is comparatively slow, on average constructing only about a quarter of the number of samples that PPG constructs.\@
However, since these samples are of ``higher quality'', the technique manages to close most of the performance gap to PPG and unidirectional path tracing, in some cases producing the best results.

\section{Discussion and Future Work}

\paragraph{Runtime Cost}
An important property of practical sampling strategies is a low computational cost of generating samples and evaluating their PDF, relative to the cost of evaluating the integrand.
In our path-guiding applications, the cost is dominated by the evaluation of coupling layers:
\ADD{roughly $10\%$ of the time is spent on one-blob encoding, $60\%$ on fully connected layers, and $30\%$ on the piecewise-polynomial warp.}
This makes the overhead of our implementation prohibitive in simple scenes.
While we focused on the theoretical challenge of applying neural networks to the problem of importance sampling in this work, accelerating the computation to make our approach more practical is an important and interesting future work.
We believe specialized hardware (e.g. NVIDIA's TensorCores) and additional computation graph optimization (e.g. NVIDIA's TensorRT) are promising next steps, which alone might be enough to bring our approach into the realm of practicality.

\paragraph{Optimizing for Multiple Integrals}
In \autoref{sec:optimizing-kl}, we briefly discussed that the ground-truth density may be available only in unnormalized form. We argued that this is not a problem since the ignored factor $F$ scales all gradients uniformly; it thus does not impact the optimization.
These arguments pertain to handling a \emph{single} integration problem.
In \autoref{sec:path-guiding}, we demonstrated applications to path sampling and path guiding, where the learned density is conditioned on additional dimensions and we are thus solving many different integrals at once.
Since the normalizing $F$ varies between them, our arguments do not extend to this particular problem.
Because neglecting the normalization factors is potentially negatively influencing the optimization, we experimented with tabulating $F$, but we did not experience noticeable improvements.
This currently stands as a limitation of applying our work to path guiding/sampling and it would be worth addressing in future work.

\paragraph{Convergence of Optimization}
\ADD{Although our optimization based on stochastic gradients has many advantages, it also brings certain disadvantages.
Techniques based on stochastic gradient descent do not converge to local optima, but oscillate around them.
This can be observed in the 2D examples in our supplementary video and also happens during neural path guiding.
The problem is well known in machine learning literature and is usually addressed by decaying the learning rate over time.
We opted not to decay our learning rate for simplicity, because finding an optimal decay schedule is a difficult problem.
Solving this issue in the future would likely improve our results further, perhaps significantly.}

\paragraph{Scene Scale}
We studied the performance of neural path guiding when all positions that are input to it are relatively close compared to the scene bounding box.
We artificially scaled the positional inputs by $10^{-5}$ in the \CountryKitchen{} scene, observing a roughly $2\times$ larger error.
While the method still outperforms path tracing by a big margin, alleviating this limitation is important future work.

\paragraph{Alternative Variance Reduction Techniques}
In this paper, we studied the application of neural networks to importance sampling. Other variance-reduction techniques, such as control variates, could enjoy analogous benefits.
We believe similar derivations to \autoref{sec:mc} can be made, leading to an interconnected gradient-descent-based optimization of multiple variance reduction techniques.

\paragraph{Alternative Training Schemes}
\citet{DBLP:journals/corr/abs-1712-06115} learn near-optimal light selection probabilities for next event estimation by minimizing an approximation (via Q-learning) of the total-variation divergence.
This optimization strategy and variations thereof are an interesting alternative to our KL and $\chi^2$ divergence loss functions.

\ADD{Another attractive goal is a unified optimization across multiple different scenes, rather than training from scratch for each one.
A potentially fruitful extension of our approach would be to apply a higher-level optimization strategy in the spirit of ``learning to learn''~\cite{l2l16,pmlr-v70-chen17e}.}

\paragraph{Failure Cases}
In the pathologically difficult \GlossyCornellBox{} scene, the theoretically inferior radiance-based NPG produces slightly better results than product-based NPG.\@
We suspect that this is caused by the product distribution being much more complicated and therefore more difficult to learn than the radiance distribution.
Furthermore, in the \Bookshelf{} scene, our approaches perform worse than the GMM algorithm by \citet{Vorba:2014:OnlineLearningPMMinLTS}.
Although our method exhibits fewer of such failure cases than PPG and the GMMs, an investigation into their causes is still to be carried out and could offer interesting insights; per-scene results with discussions of rendering challenges are provided in the supplementary material.

\section{Conclusion}
We introduced a technique for importance sampling with neural networks. 
Our approach builds partly on prior works and partly on three novel extensions:
we proposed piecewise-polynomial coupling transforms that increase the modeling power of coupling layers,
we introduced the one-blob encoding that helps the network to specialize its parts to different input configurations,
and, finally, we derived an optimization strategy that aims at reducing the variance of Monte Carlo estimators that employ trainable probabilistic models.
We demonstrated the benefits of our online-learning approach in a number of settings, ranging from canonical examples to production-oriented ones: learning the distribution of natural images and path sampling and path guiding for simulation of light transport.
\ADD{In the vast majority of cases, our technique performed favorably in equal sample count comparisons against prior art.}

This paper brings together techniques from machine learning, developed initially for density estimation, and applications to Monte Carlo integration, with examples from the field of rendering.
We hope that our work will stimulate further applications of deep neural networks to importance sampling and integration problems.

\section{Acknowledgments}

We thank Thijs Vogels for bringing RealNVP to our attention and Sebastian Herholz, Yining Karl Li, and Jacob Munkberg for valuable feedback. We are grateful to \citet{Vorba:2014:OnlineLearningPMMinLTS} and \citet{dinh2016density} for releasing the source code of their work.
We also thank the following people for providing scenes and models that appear in our figures:
Benedikt \citet{resources16},
Ond\v{r}ej Karl\'{\i}k (\SwimmingPool),
Johannes Hanika (\Necklace),
Samuli Laine and Olesya Jakob (\CopperHairball),
Jay-Artist (\CountryKitchen, \WhiteRoom),
Marios Papas and Maurizio Nitti,
Marko Dabrovi\'c (\Sponza),
Miika Aittala, Samuli Laine, and Jaakko Lehtinen (\VeachDoor),
Nacimus (\SalleDeBain), 
SlykDrako (\Bedroom),
thecali (\Spaceship),
Tiziano Portenier (\Bathroom, \Bookshelf), and
Wig42 (\WoodenStaircase).
Production baby: V\'{\i}t Nov\'ak.

\appendix

\section{Determinant of Coupling Layers}%
\label{app:coupling-layer-determinant}

Here we include the derivation of the Jacobian determinant akin to~\citet{dinh2016density}. The Jacobian of a single coupling layer, where $\partitionA = \ldb 1,d\rdb$ and $\partitionB = \ldb d+1, D \rdb$, is a block matrix:
\begin{align}
\frac{\partial y}{\partial \layerIn^T} = 
\begin{bmatrix}
I_d & 
0 \\ 
\frac{\partial \cmap\left(\layerIn^\partitionB; \nnet(\layerIn^\partitionA)\right)}{{\partial (\layerIn^\partitionA)}^T} & 
\frac{\partial \cmap\left(\layerIn^\partitionB; \nnet(\layerIn^\partitionA)\right)}{{\partial (\layerIn^\partitionB)}^T} 
\end{bmatrix},
\end{align}
where $I_d$ is a $d\times d$ identity matrix.
The determinant of the Jacobian matrix reduces to the determinant of the lower right block. Note that the Jacobian $\frac{\partial \cmap\left(\layerIn^\partitionB; \nnet(\layerIn^\partitionA)\right)}{{\partial (\layerIn^\partitionA)}^T}$ (lower left block) does not appear in the determinant, hence $\nnet$ can be arbitrarily complex.

For the multiply-add coupling transform~\cite{dinh2016density} we~get
\begin{align}
\frac{\partial \cmap\big(\layerIn^\partitionB; \nnet(\layerIn^\partitionA)\big)}{{\partial (\layerIn^\partitionB)}^T} = 
  \begin{bmatrix}
    e^{s_1} & & 0 \\
    & \ddots & \\
    0 & & e^{s_{D-d}}
  \end{bmatrix}.
\end{align}
The diagonal nature stems from the separability of the coupling transform.
The determinant of the coupling layer in the forward and the inverse pass therefore reduce to $e^{\sum s_i}$ and $e^{-\sum s_i}$, respectively.

\ADD{In our piecewise-polynomial coupling transforms, we maintain separability to preserve the diagonal Jacobian, i.e.\
\begin{align}
  \cmap\big( \layerIn^\partitionB; \nnet(\layerIn^\partitionA) \big) = {\Big(\cmap_1\big( \layerIn_1^\partitionB; \nnet(\layerIn^\partitionA) \big)\,,\, \cdots \,,\, \cmap_{D-d}\big( \layerIn_{D-d}^\partitionB; \nnet(\layerIn^\partitionA) \big) \Big)}^T \,, \nonumber
\end{align}
and therefore, using $\frac{\partial \cmap_i\big( \layerIn_i^\partitionB; \nnet(\layerIn^\partitionA) \big)}{\partial \layerIn_i^\partitionB} = \PdfPoly_i( \layerIn_i^\partitionB )$, we get
\begin{align}
  \frac{\partial \cmap\big(\layerIn^\partitionB; \nnet(\layerIn^\partitionA)\big)}{{\partial (\layerIn^\partitionB)}^T} &= 
  \begin{bmatrix}
    \PdfPoly_1( \layerIn_1^\partitionB ) & & 0 \\
    & \ddots & \\
    0 & & \PdfPoly_{D-d}( \layerIn_{D-d}^\partitionB )
  \end{bmatrix}.
\end{align}
The determinant thus is the product of the marginal PDFs defining the piecewise-polynomial warp along each dimension $\prod_{\dimId=1}^{D-d} \PdfPoly_\dimId(\layerIn_\dimId^\partitionB)$.}

\section{Adaptive Bin Sizes in Piecewise-Linear Coupling Functions}%
\label{app:piecewise-no-bin-optimization}

Without loss of generality, we investigate the simplified scenario of a one-dimensional input $\partitionA = \emptyset$ and $\partitionB = \{ 1 \}$, a single coupling layer $\nlayers=1$ and the KL-divergence loss function.
Further, let the coupling layer admit a piecewise-linear coupling transform---i.e.\ it predicts a piecewise-constant PDF---with $\nbins = 2$ bins.
Let the width $\matW$ of the $2$ bins be controlled by traininable parameter $\theta \in \mathbb{R}$ such that $\matW_1 = \theta$ and $\matW_2 = 1 - \theta$ and $S = \matQ_1 \theta + \matQ_2 (1 - \theta)$, then
\begin{align}
  \PdfOptimized(x;\theta) =
  \begin{cases}
    \matQ_1 / S & \text{ if } x < \theta \\
    \matQ_2 / S & \text{ otherwise.}
  \end{cases}
\end{align}
Using \autoref{eq:grad-kl}, the gradient of the KL divergence w.r.t.\ $\theta$ is
\begin{align}
  \nabla_\theta \KlDiv(p \, \| \, q; \theta) &= \nabla_\theta \int_0^1
  \begin{cases}
    \PdfGt(x) \log( \matQ_1 / S) & \text{ if } x < \theta \\
    \PdfGt(x) \log( \matQ_2 / S) & \text{ otherwise}
  \end{cases} \Diff{x}
  \, ,
\end{align}
where---in contrast to our piecewise-quadratic coupling function---the gradient can \emph{not} be moved into the integral (see \autoref{eq:grad-kl-expectation}) due to the discontinuity of $\PdfOptimized$ at $\theta$.
This prevents us from expressing the stochastic gradient of Monte Carlo samples with respect to $\theta$ in closed form and therefore optimizing with it.

We further investigate ignoring this limitation and performing the simplification of \autoref{eq:grad-kl-expectation} regardlessly, resulting in
\begin{align}
  \nabla_\theta  \KlDiv(p \, \| \, q; \theta) \approx \Expectation \left[
  \begin{cases}
    \PdfGt(\Sample) \left( 1 - \frac{\matQ_2}{\matQ_1} \right) & \text{ if } \Sample < \theta \\
    \PdfGt(\Sample) \left( \frac{\matQ_1}{\matQ_2} - 1 \right) & \text{ otherwise}
  \end{cases}
  \right] \, ,
\end{align}
which has the same sign \emph{regardless of the value of $\theta$}, resulting in divergent behavior.
A similarly undesirable (albeit different) behavior emerges when normalizing $\PdfOptimized$ in a slightly different way by interpreting $\matQ$ as probability masses rather than unnormalized densities:
\begin{align}
  \PdfOptimized(x;\theta) =
  \begin{cases}
    \matQ_1 / \theta & \text{ if } x < \theta \\
    \matQ_2 / (1 - \theta) & \text{ otherwise.}
  \end{cases}
\end{align}
The KL divergence gradient is then 
\begin{align}
  \nabla_\theta  \KlDiv(p \, \| \, q; \theta) &\approx \int_0^1
  \begin{cases}
    \PdfGt(x) / \theta & \text{ if } x < \theta \\
    \PdfGt(x) / (\theta - 1) & \text{ otherwise,}
  \end{cases}
  \Diff{x} \nonumber \\
  &= \frac{1}{\theta} \int_0^\theta \PdfGt(x) \Diff{x} - \frac{1}{1 - \theta} \int_\theta^1 \PdfGt(x) \Diff{x}
\,.
\end{align}
To illustrate the flawed nature of this gradient, consider the simple scenario of $\PdfGt(x) = 1$, in which the RHS \emph{always} equals to zero, suggesting \emph{any} $\theta$ being a local minimum.
However, $\theta$ clearly influences $\KlDiv(p \, \| \, q; \theta)$ in this example, and therefore can not be optimal everywhere.
Empirical investigations with other shapes of $\PdfGt$, e.g.\ the examples from \autoref{fig:2d-dists}, also suffer from a broken optimization and do not converge to a meaningful result.

While we only discuss a simplified setting here, the simplification in \autoref{eq:grad-kl-expectation} is also invalid in the \emph{general} case of piecewise-linear coupling functions, likewise leading to a broken optimization.

\bibliography{bibliography}
\bibliographystyle{ACM-Reference-Format}

\end{document}